\documentclass[lettersize,journal]{IEEEtran}
\IEEEoverridecommandlockouts
\usepackage{cite}
\usepackage{amsthm}
\usepackage{amsmath,amssymb,amsfonts}
\usepackage{algorithmic}
\usepackage{graphicx}
\usepackage{textcomp}
\usepackage{algorithm}
\usepackage{graphicx}
\usepackage{amssymb}
\usepackage{caption}
\usepackage{epsfig}
\usepackage{subfigure}
\usepackage{amsmath}

\usepackage{array}
\usepackage{color}
\usepackage{amsmath}
\usepackage{longtable}
\usepackage{algorithmic}
\usepackage{algorithm}
\usepackage{cases}
\usepackage{bm}
\usepackage{mathrsfs}
\usepackage{psfrag}
\usepackage{url}
\usepackage{float}
\usepackage{multicol}
\usepackage{soul}
\usepackage[dvipsnames, svgnames, x11names]{xcolor} 
\usepackage{footnote}
\usepackage{setspace}   
\usepackage{etoolbox}

\soulregister\cite7
\soulregister\ref7
\soulregister\eqref7
\soulregister\bf7

\makesavenoteenv{algorithm} 

\makeatletter
\patchcmd{\@makecaption}
  {\scshape}
  {}
  {}
  {}
\makeatletter
\patchcmd{\@makecaption}
  {\\}
  {.\ }
  {}
  {}
\makeatother

\IEEEoverridecommandlockouts

\def\BibTeX{{\rm B\kern-.05em{\sc i\kern-.025em b}\kern-.08em
    T\kern-.1667em\lower.7ex\hbox{E}\kern-.125emX}}

\allowdisplaybreaks[4]  

\newtheorem{Thm}{Theorem}
\newtheorem{Lem}{Lemma}

\newtheorem{Asump}{Assumption}
\newtheorem{Prob}{Problem}

\newtheorem{Rmk}{Remark}

\newcommand{\setNbar}{\bar{\mathcal N}}

\newcommand{\clrtmp}{\textcolor{black}}

\begin{document}

\title{GQFedWAvg: Optimization-Based Quantized Federated Learning in General Edge Computing Systems}

\author{Yangchen Li, Ying Cui, and Vincent Lau
\thanks{
Yangchen Li is with the School of EIEE, Shanghai Jiao Tong University, Shanghai 200240, China and IoT Thrust, the HKUST (GZ), Guangzhou 511400, Guangdong, China.
Ying Cui is with IoT Thrust, the HKUST (GZ), Guangzhou 511400, Guangdong, China and the Department of ECE, the HKUST, Hong Kong SAR, China.
Vincent Lau is with the Department of ECE, the HKUST, Hong Kong SAR, China.
This paper was presented in part at IEEE ICC 2023 \cite{ICC2023}.}
}
\maketitle

\begin{abstract}
The optimal implementation of federated learning (FL) in practical edge computing systems has been an outstanding problem.
In this paper, we propose an optimization-based quantized FL algorithm, which can appropriately fit a general edge computing system with uniform or nonuniform computing and communication resources at the workers.
Specifically, we first present a new random quantization scheme and analyze its properties.
Then, we propose a general quantized FL algorithm, namely GQFedWAvg.
Specifically, GQFedWAvg applies the proposed quantization scheme to quantize wisely chosen model update-related vectors and adopts a generalized mini-batch stochastic gradient descent (SGD) method with the weighted average local model updates in global model aggregation.
Besides, GQFedWAvg has several adjustable algorithm parameters to flexibly adapt to the computing and communication resources at the server and workers.
We also analyze the convergence of GQFedWAvg.
Next, we optimize the algorithm parameters of GQFedWAvg to minimize the convergence error under the time and energy constraints.
We successfully tackle the challenging non-convex problem using general inner approximation (GIA) and multiple delicate tricks.
Finally, we interpret GQFedWAvg's function principle and show its considerable gains over existing FL algorithms using numerical results.
\end{abstract}


\setcounter{page}{1}
\section{Introduction}\label{sec:Introduction}
Recent years have witnessed a growing interest in federated learning (FL) in general edge computing systems, where the computing and communication resources at the workers may be uniform or distinct~\cite{EdgeIntelligence,DeepLearning-EdgeComputing,FL-MobileEdgeNetworks}. FL enables the server and workers to collaboratively train a global model with local data maintained at the workers, thereby successfully protecting data privacy.
Existing FL algorithms, such as Parallel Mini-batch stochastic gradient descent (PM-SGD) \cite{PMSGD}, Parallel Restarted SGD (PR-SGD) \cite{YuHao}, and Federated Averaging (FedAvg) \cite{FedAvg}, mainly rely on mini-batch SGD methods.
Specifically, in each global iteration, the server updates the global model by aggregating and averaging the latest local models, and each worker updates its local model based on the latest global model via executing multiple local iterations of mini-batch SGD on its local data and then sends it to the server.
It is worth noting that in \cite{PMSGD,YuHao,FedAvg}, exact model updates specified by a huge number of information bits are sent, which may yield a long transmission time and a high energy consumption;
the algorithm parameters are chosen experimentally only to improve convergence, which may, in turn, impose huge computing and communication costs.
Existing works attempt to address these issues from the following two aspects.

Firstly, the authors in \cite{SIGNSGD,QSGD,Wyner-Ziv,Q-CompressiveSampling,UVeQFed,
High-Dimensional-SGQ,FedPAQ,CongShen,GenQSGD,Qsparse-local-SGD,FastFL,
AdaptCtrl-LocUpdt-ModCompr,AdaptCommunStrat,SparseBinaryCompression,SparseCommun-DistributedGD,
eSGD} propose to compress the model updates before sending them to reduce the communication cost in each global iteration.
Specifically, in \cite{SIGNSGD,QSGD,Wyner-Ziv,Q-CompressiveSampling}, each element of a model update-related vector is quantized using a deterministic bi-level scalar quantizer \cite{SIGNSGD} or a random multi-level scalar quantizer \cite{QSGD,Wyner-Ziv,Q-CompressiveSampling}, respectively;
in \cite{UVeQFed,High-Dimensional-SGQ}, a model update-related vector is partitioned into subvectors, and each subvector is quantized using a vector quantizer;
in \cite{FedPAQ,CongShen,GenQSGD}, a model update-related vector is quantized using an unspecified quantizer with a given precision;
and in \cite{SparseBinaryCompression,eSGD,SparseCommun-DistributedGD,FastFL,
AdaptCtrl-LocUpdt-ModCompr,Qsparse-local-SGD,AdaptCommunStrat}, a model update-related vector is compressed using sparsification.
In addition, \cite{FedPAQ,UVeQFed,QSGD,SIGNSGD,High-Dimensional-SGQ,Wyner-Ziv,
Q-CompressiveSampling,CongShen,GenQSGD,Qsparse-local-SGD,FastFL,
AdaptCtrl-LocUpdt-ModCompr,AdaptCommunStrat} analyze the convergences of the corresponding FL algorithms,
mainly considering three assumptions on gradients, i.e., bounded gradient variance \cite{PMSGD,SIGNSGD,QSGD,CongShen,AdaptCommunStrat}, bounded gradient moment \cite{QSGD,Q-CompressiveSampling,UVeQFed,CongShen}, and bounded individual gradient \cite{Wyner-Ziv,High-Dimensional-SGQ,Qsparse-local-SGD}, in order from weak to strong.
\cite{High-Dimensional-SGQ,AdaptCtrl-LocUpdt-ModCompr} further optimize the compression parameters to reduce the convergence error.
However, in \cite{FedPAQ,CongShen,Qsparse-local-SGD,Wyner-Ziv,UVeQFed,QSGD,
Q-CompressiveSampling,GenQSGD}, the exact norms of model update-related vectors are assumed to be represented with finite bits and sent without quantization, which is not reasonable;
in \cite{FedPAQ,UVeQFed,High-Dimensional-SGQ,Wyner-Ziv,Q-CompressiveSampling,
SparseBinaryCompression,eSGD,Qsparse-local-SGD,CongShen,FastFL,
AdaptCtrl-LocUpdt-ModCompr,AdaptCommunStrat}, the workers send quantized local model update-related vectors, but the server still sends exact global model update-related vectors, which does not effectively reduce the communication cost;
in \cite{SparseBinaryCompression,eSGD,SparseCommun-DistributedGD}, the authors do not study the convergence of the corresponding FL algorithms;
in \cite{SIGNSGD,Q-CompressiveSampling,High-Dimensional-SGQ}, the convergence results hold only for FL algorithms where each worker executes one local iteration in each global iteration, which are not communication-efficient;
in \cite{UVeQFed,CongShen}, the convergence results hold only for strongly-convex ML problems.

Secondly, some existing works propose to optimize the parameters of the underlying mini-batch SGD algorithms, such as the numbers of local iterations \cite{AdaptCommunStrat,AdaptFL,ResAlls2-1,ResAlls3-2,ResAlls4-1,FastFL,AdaptCtrl-LocUpdt-ModCompr,GenQSGD}, mini-batch size \cite{GenQSGD}, step size sequence \cite{GenQSGD}, and weights in the weighted average local model updates \cite{CongShen}, to minimize the convergence error \cite{CongShen,AdaptCommunStrat,AdaptFL,FastFL}, time consumption \cite{ResAlls3-2,ResAlls4-1,AdaptCtrl-LocUpdt-ModCompr}, and energy consumption \cite{GenQSGD,ResAlls2-1,ResAlls4-1}.
Nevertheless, the numbers of local iterations at all workers \cite{AdaptCommunStrat,FastFL,AdaptFL} and weights in the weighted average local model updates \cite{UVeQFed,High-Dimensional-SGQ,Wyner-Ziv,Q-CompressiveSampling,eSGD,Qsparse-local-SGD,FastFL,AdaptCommunStrat,ResAlls2-1,ResAlls3-2,ResAlls4-1,GenQSGD} are set to be identical,
which may degrade the performance of FL in general edge computing systems with possibly nonuniform computing and communication resources at the workers.

In summary, in existing literature \cite{SIGNSGD,QSGD,FedPAQ,UVeQFed,Wyner-Ziv,Q-CompressiveSampling,
High-Dimensional-SGQ,CongShen,GenQSGD,Qsparse-local-SGD,FastFL,
AdaptCtrl-LocUpdt-ModCompr,SparseBinaryCompression,SparseCommun-DistributedGD,
eSGD,AdaptCommunStrat,AdaptFL,ResAlls3-2,ResAlls4-1,ResAlls2-1}, the quantization schemes for FL are not practical or efficient, and the choices of the FL algorithm parameters are not optimal for general systems with arbitrary computing and communication resources at the workers.
This paper will shed some light on these issues, aiming to implement FL effectively and efficiently in a general edge computing system with uniform or nonuniform computing and communication resources at the server and workers.
The main contributions are summarized as follows.
\begin{itemize}
\item[$\bullet$]\textbf{Quantization Design:} We present a new random quantization scheme that quantizes the amplitude of a vector and each element of the normalized vector separately using scalar random quantizers.
    Each scalar quantizer is parameterized by the number of quantization levels and input range, which can flexibly adapt to communication resources.
    Note that the proposed quantization scheme is more practical than those in \cite{FedPAQ,CongShen,Qsparse-local-SGD,Wyner-Ziv,UVeQFed,QSGD,
    Q-CompressiveSampling,GenQSGD} since it quantizes the magnitude of a vector besides each element of the normalized vector.
    We also analyze the properties of the proposed quantization scheme.
    It is worth noting that the analysis is more challenging than those in \cite{SIGNSGD,QSGD,UVeQFed,Wyner-Ziv,Q-CompressiveSampling} due to the quantization of the magnitude of the vector.
\item[$\bullet$]\textbf{FL Algorithm Design:} We propose a general quantized FL algorithm, namely GQFedWAvg, for the server and workers to collaboratively solve a machine learning (ML) problem.
    GQFedWAvg applies the proposed quantization scheme to quantize wisely chosen scaled model update-related vectors and adopts a generalized mini-batch SGD method with the weighted average local model updates in global model aggregation.
    Besides, GQFedWAvg has several adjustable algorithm parameters including the parameters of the quantization scheme, i.e., the numbers of quantization levels and input ranges at the server and each worker, and the parameters of the underlying mini-batch SGD algorithm, i.e., the numbers of global and local iterations, mini-batch size, step size sequence, and weights in the weighted average local model updates, to flexibly adapt to the computing and communication resources at the server and workers.
    Compared to \cite{UVeQFed,High-Dimensional-SGQ,Wyner-Ziv,Q-CompressiveSampling,AdaptFL,eSGD,Qsparse-local-SGD,FastFL,AdaptCommunStrat,ResAlls2-1,ResAlls3-2,ResAlls4-1,GenQSGD}, GQFedWAvg can effectively utilize the computing and communication resources at the workers to improve the convergence performance.
    We also analyze the convergence of GQFedWAvg.
    Note that the convergence result in this paper holds for general convex or nonconvex ML problems, thus is more general than the convergence results in \cite{UVeQFed,CongShen}.
\item[$\bullet$]\textbf{Optimization of Algorithm Parameters:} We optimize the algorithm parameters of GQFedWAvg to minimize the convergence error under the time cost, energy cost, step size sequence, and weight constraints.
    The corresponding optimization problem is a challenging non-convex problem.
    We equivalently transform the original problem to a more tractable one using multiple delicate tricks and propose an iterative algorithm to obtain a KKT point using general inner approximation (GIA) \cite{GIA} and several delicate tricks.
\item[$\bullet$]\textbf{Numerical Results:} We numerically demonstrate how the algorithm parameters of GQFedWAvg optimally adapt to the computing and communication resources at the server and workers and how the optimized GQFedWAvg achieves a trade-off among the time cost, energy cost, and convergence error.
    We also numerically and experimentally show remarkable gains of the optimized GQFedWAvg over existing FL algorithms \cite{PMSGD,YuHao,CongShen,GenQSGD}.
\end{itemize}

{\bf{Notation}}:
We use boldface letters (e.g., $\mathbf x$), non-boldface letters (e.g., $x$ or $X$), and calligraphic letters (e.g., $\mathcal X$) to represent vectors, scalar constants, and sets, respectively.
$\|\cdot\|_2$, $\mathbb E[\cdot]$, and $\mathbb I[\cdot]$ represent the $l_2$-norm, expectation, and indicator function, respectively.
\clrtmp{$\vert\cdot\vert$ represents the cardinality of a set.
$\langle\cdot,\cdot\rangle$ represents the inner product of two vectors.}
The set of real numbers, positive real numbers, and positive integers are denoted by $\mathbb R$, \clrtmp{$\mathbb R_{++}$, and $\mathbb Z_{++}$}, respectively.
The zero vector and all-ones vector are represented by $\mathbf0$ and $\mathbf 1$, respectively.
The key notation used in this paper is listed in Table \ref{tab:notation}.

\begin{table}[h]
\center
\fontsize{7pt}{8pt}\selectfont{
\begin{tabular}{|m{0.055\textwidth}<{\centering}|m{0.38\textwidth}<{\centering}|}
\hline
      Notation & Description \\ \hline
      $\setNbar$ & set of the server and $N$ workers \\ \hline
      $\mathcal{I}_n$ & set of $I_n$ samples held by worker $n$ \\ \hline
      $\mathbf{x}$ & global model parameters \\ \hline
      $F\left(\mathbf{x};\zeta_n\right)$ & loss incurred by the random variable $\zeta_n$ \\ \hline
      $f_n(\mathbf{x})$ & expected loss for worker $n$ \\ \hline
      $F_{n'}$ & CPU frequency \\ \hline
      $p_{n'}$ &  transmission power \\ \hline
      $r_{n'}$ & average transmission rate \\ \hline
      $\tilde{s}_{n'}$ & quantization level for magnitude \\ \hline
      $s_{n'}$ & quantization level for normalized vector \\ \hline
      $\Delta_{n'}$ & quantization input range \\ \hline
      $K_0$, $K_n$ & numbers of global iterations and local iterations of worker $n$ \\ \hline
      $B$ & local mini-batch size \\ \hline
      $\gamma^{(k_0)}$ & step size used within the $k_0$-th global iteration \\ \hline
      $W_n$ & weight for worker $n$ in the weighted average of local model updates \\ \hline
      $\Delta\hat{\mathbf{x}}^{(k_0)}$ & weighted average of the quantized version of normalized accumulated local model updates \\ \hline
      $\hat{\mathbf{x}}^{(k_0)}$ & global model within the $k_0$-th global iteration \\ \hline
      $\hat{\mathbf{x}}_n^{(k_0,k_n)}$ & the $k_n$-th local model of worker $n$ within the $k_0$-th global iteration \\
      \hline
    \end{tabular}
    }\\
  \caption{\small{Key notations. $n'\in\{0,1,\cdots,N\}$ represents the server ($n'=0$) or
  worker $n'$ ($n'=1,2,\cdots,N$). $n\in\{1,2,\cdots,N\}$ represents worker $n$.}}
  \label{tab:notation}
\end{table}\setlength{\textfloatsep}{10pt}

\section{System Model}
\subsection{Federated Edge Learning}
\begin{figure}[t]
\begin{center}
{\includegraphics[width=250pt]{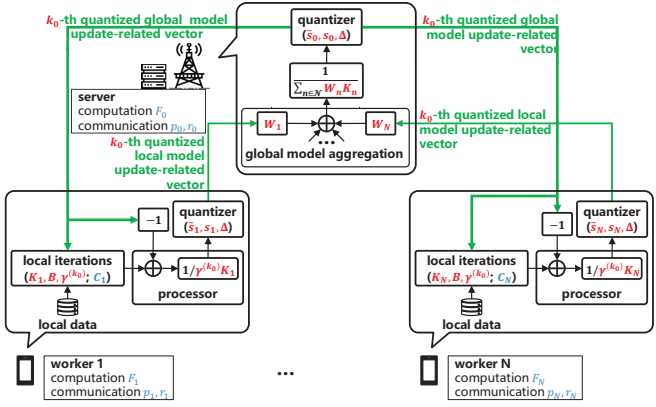}}
\caption{\small{Federated learning in a general edge computing system.}}
\label{Fig:SystemFig}
\end{center}
\end{figure}
\setlength{\textfloatsep}{10pt}
We consider an edge computing system consisting of one server and $N$ workers connected via wireless links, as shown in Fig.~\ref{Fig:SystemFig}.
Let $0$ and $\mathcal N\triangleq\{1,2,\cdots,N\}$ denote the server index and the set of worker indices, respectively.
For ease of exposition, we also denote $\setNbar\triangleq\{0\}\cup\mathcal N$.
We assume that each worker $n\in\mathcal N$ holds $I_n$ samples,
denoted by $\xi_i,i\in\mathcal I_n$ with $\left|\mathcal I_n\right|=I_n$.
Note that $\xi_i,i\in\mathcal I_n$ can be viewed as the realizations of a random variable, denoted by $\zeta_n$.
The server and $N$ workers aim to train a global model collaboratively by solving an ML problem based on the local data stored on the $N$ workers.
The global model is parameterized by a $D$-dimensional vector ${\mathbf x}\in \mathbb R^{D}$.
Specifically, for a given $\mathbf x\in\mathbb R^D$, define the loss incurred by $\zeta_n$ as $F({\mathbf x};\zeta_n)$ and define the expected loss as $f_n({\mathbf x})\triangleq\mathbb E\left[F({\mathbf x};\zeta_n)\right]$, with the expectation taken with respect to the distribution of $\zeta_n$, for all $n\in\mathcal N$.
Then, the expected risk function $f:\mathbb R^{D}\rightarrow\mathbb R$ of the model parameters ${\mathbf x}\in\mathbb R^{D}$ is defined as:
\begin{align}\label{eq:expected_loss}
f({\mathbf x})\triangleq\frac{1}{N}\sum_{n\in\mathcal N}f_n({\mathbf x}).
\end{align}
To be general, we do not assume $f({\mathbf x})$ to be convex.
Our goal is to minimize the expected risk function with respect to the model parameters ${\mathbf x}$ in the edge computing system.
\begin{align}\label{Obj_ML}
\text{(ML Problem)}\hspace{20pt}\min_{{\mathbf x}}f({\mathbf x})
\end{align}
where $f({\mathbf x})$ is given by \eqref{eq:expected_loss}.
The problem in \eqref{Obj_ML} is an unconstrained problem that may be convex or non-convex.
The goal of solving an unconstrained problem is generally to design an iterative algorithm to obtain a stationary point.
The server and $N$ workers all have computing and communication capabilities.
Let $F_0$ and $F_n$ denote the CPU frequencies (cycles/s) of the server and worker $n\in\mathcal N$, respectively.
Let $p_0$ and $p_n$ denote the transmission powers of the server and worker $n\in\mathcal N$, respectively.
In the process of collaborative training,
\clrtmp{the server and all $N$ workers employ quantization before transmitting messages}.
\clrtmp{Specifically}, the server multicasts messages to the $N$ workers at an average rate $r_0$ (b/s) over the whole frequency band,
and the $N$ workers transmit their messages to the server at average transmission rates $r_n,n\in\mathcal N$ (b/s) using frequency division multiple access (FDMA).

\begin{Rmk}[General Edge Computing System]
The edge computing system considered here is general (can be heterogeneous or homogeneous) in the sense that the system parameters, e.g., $F_n,p_n,r_n,n\in\setNbar$, can be different.
\end{Rmk}

\section{Quantization Scheme Design}\label{Sec:Quantization}
In this section, we first present a new random quantization scheme for quantizing vectors and analyze the quantization properties.
The proposed quantization scheme will be used for compressing wisely chosen model update-related vectors in GQFedWAvg.

First, we present an $\left(\mathfrak{s}+1\right)$-level random scalar quantizer which has an input set $[0,\Delta]$ and an output set $\mathcal S_{\mathfrak{s}}\triangleq\left\{0,\frac{1}{\mathfrak{s}},\frac{2}{\mathfrak{s}},\cdots,1\right\}$ and is specified by the mapping $\mathfrak{Q}(\cdot;\mathfrak{s},\Delta):[0,\Delta]\rightarrow\mathcal{S}_{\mathfrak{s}}$ for all $\mathfrak{s}\in\mathbb Z_{++}$ and $\Delta\in\mathbb R_{++}$.
For all $\mathfrak{s}\in\mathbb Z_{++}$, let $\mathcal C_{\mathfrak{s},0}\triangleq\left\{0\right\}$ and $\mathcal C_{\mathfrak{s},j}\triangleq\left(\frac{(j-1)\Delta}{\mathfrak{s}},\frac{j\Delta}{\mathfrak{s}}\right]\subset[0,\Delta]$, $j=1,2,\cdots,\mathfrak{s}$.
Clearly, $C_{\mathfrak{s},j}$, $j\in\left\{0,1,\cdots,\mathfrak{s}\right\}$ form a partition of the input set $[0,\Delta]$.
Then, for all $\mathfrak{s}\in\mathbb{Z}_{++}$ and $\Delta\in\mathbb{R}_{++}$, the mapping $\mathfrak{Q}(\cdot;\mathfrak{s},\Delta)$ is given by:
\begin{align}\label{eq:RandVar}
&\mathfrak{Q}(u;\mathfrak{s},\Delta)
=\nonumber\\&\left\{
    \begin{array}{lll}
    \!\!0,
    &\!\!\text{w.p. }1,
    &u\in\mathcal C_{\mathfrak{s},0},\\
    \!\!\left(j-1\right)\frac{\Delta}{\mathfrak{s}},
    &\!\!\text{w.p. }j-\frac{\mathfrak{s}u}{\Delta},
    &u\in\mathcal C_{\mathfrak{s},j}, j\in\left\{1,2,\cdots,\mathfrak{s}\right\},\\
    \!\!j\frac{\Delta}{\mathfrak{s}},
    &\!\!\text{w.p. }\frac{\mathfrak{s}u}{\Delta}-j+1,
    &u\in\mathcal C_{\mathfrak{s},j}, j\in\left\{1,2,\cdots,\mathfrak{s}\right\}.
    \end{array}\right.
\end{align}
Note that $\mathfrak{Q}(\cdot;\mathfrak{s},\Delta)$ is a random variable.

Next, we present a new random quantization scheme, parameterized by the numbers of quantization levels $\tilde s\in\mathbb Z_{++}$, $s\in\mathbb Z_{++}$, and the input range $\Delta\in\mathbb R_{++}$, for quantizing $D$-dimension vectors with magnitudes no greater than $\Delta$. For example, consider a $D$-dimension vector $\mathbf y\triangleq\left(y_d\right)_{d\in\mathcal D}\in\mathbb R^D$ with $\|\mathbf y\|_2\in[0,\Delta]$, where $\mathcal D\triangleq\{1,2,\cdots,D\}$.
By noting that for all $d\in\mathcal{D}$, $y_d$ can be expressed as $y_d=\|\mathbf y\|_2\cdot\text{sgn}\left(y_d\right)\cdot\frac{\left|y_d\right|}{\|\mathbf y\|_2}$,
we quantize $\|\mathbf y\|_2$, $\frac{|y_d|}{\|\mathbf y\|_2}$, $d\in\mathcal D$ separately and use $1$ bit to represent $\text{sgn}\left(y_d\right)$ for all $d\in\mathcal{D}$.
In particular, $\|\mathbf y\|_2\in[0,\Delta]$ is quantized by using $\mathfrak{Q}\left(\cdot;{\tilde s},\Delta\right)$
given by \eqref{eq:RandVar} and represented by $\log_2\left(\tilde s+1\right)$ bits,
and $\frac{|y_d|}{\|\mathbf y\|_2}\in[0,1]$, $d\in\mathcal D$ are quantized by using $\mathfrak{Q}\left(\cdot;s,1\right)$ given by \eqref{eq:RandVar} and represented by $\log_2\left(s+1\right)$ bits.
Thus, the total number of bits used for expressing the quantized version of $\mathbf{y}$ is given by:
\begin{align}\label{eq:Ms}
M_{\tilde s,s}\triangleq\log_2(\tilde s+1)+D\left(\log_2(s+1)+1\right),\
\end{align}
which increases both with the dimension of the global model $D$ and the quantization parameters $\tilde s$ and $s$.
Finally, the quantized version of $\mathbf y$, denoted by $\mathbf Q(\mathbf y;\tilde s,s,\Delta)\triangleq\left(Q_d(\mathbf y;\tilde s,s,\Delta)\right)_{d\in\mathcal D}:\mathbb R^D\rightarrow\mathbb R^D$, can be obtained according to:
\begin{align}\label{eq:Q_d}
Q_d(\mathbf y;\tilde s,s,\Delta)\triangleq
&\mathfrak{Q}\left(\|\mathbf y\|_2;{\tilde s},\Delta\right)\cdot\text{sgn}(y_d)\cdot\mathfrak{Q}\left(\frac{|y_d|}{\|\mathbf y\|_2};s,1\right),\nonumber\\
&\hspace{100pt}\ d\in\mathcal D.
\end{align}

Finally, we show that ${\mathbf Q}(\cdot;\tilde s,s,\Delta)$ given by \eqref{eq:Q_d} satisfies the following properties.
\begin{Lem}[Properties of ${\mathbf Q}(\cdot;\tilde s,s,\Delta)$ given by \eqref{eq:Q_d}]\label{Lem:Quantization}
For all $\Delta\in\mathbb R_{++}$, $\tilde s\in\mathbb Z_{++}$, $s\in\mathbb Z_{++}$, and $\mathbf y\in\mathbb R^D$ with $\|\mathbf y\|_2\in[0,\Delta]$, ${\mathbf Q}(\cdot;\tilde s,s,\Delta)$ given by \eqref{eq:Q_d} satisfies:
\begin{itemize}
\item[(i)] $\mathbb E\left[{\mathbf Q}(\mathbf y;\tilde s,s,\Delta)\right]=\mathbf y$;
\item[(ii)] $\mathbb E\left[\left\|{\mathbf Q}(\mathbf y;\tilde s,s,\Delta)-\mathbf y\right\|_2^2\right]
    \leq q_{\tilde s,s}\Delta^2
    +q_s\|\mathbf y\|_2^2$;
\item[(iii)] $\left(\left\|\mathbf y\right\|_2-\frac{1}{\tilde s}\right)^2\left(1-\frac{\sqrt{D}}{s}\right)^2
    \leq\left\|\mathbf Q(\mathbf y;\tilde s,s,\Delta)\right\|_2^2
    \leq\left(\left\|\mathbf y\right\|_2+\frac{1}{\tilde s}\right)^2\left(1+\frac{\sqrt{D}}{s}\right)^2$.
\end{itemize}
Here, the constants $q_{\tilde s,s}$ and $q_s$ are given by:\footnote{\clrtmp{Note that $q_{\tilde s,s}$ decreases with $\tilde s$ and $s$, and $q_s$ decreases with $s$}.}
\begin{align}
\!\!\!q_{\tilde s,s}
\!=\!\frac{1}{4\tilde s^2}\!\!\left(\!1+\min\!\left(\frac{D}{s^2}, \frac{\sqrt{D}}{s}\right)\!\!\right),\
\!q_s\!=\!\min\left(\frac{D}{s^2},\!\frac{\sqrt{D}}s\right)\!.\label{eq:qs}
\end{align}
\end{Lem}
\begin{IEEEproof}
See Appendix~A.
\end{IEEEproof}

\clrtmp{
Lemma~\ref{Lem:Quantization} (i) indicates that ${\mathbf Q}(\mathbf{y};\tilde s,s,\Delta)$ is an unbiased quantizer.
Lemma~\ref{Lem:Quantization} (ii) suggests that the variance of ${\mathbf Q}(\mathbf y;\tilde s,s,\Delta)$ increases with $\Delta$ and $\left\|\mathbf y\right\|_2$, decreases with $\tilde s$ and $s$, and vanishes as $\tilde s\rightarrow\infty$ and $s\rightarrow\infty$.
Lemma~\ref{Lem:Quantization} (iii) indicates that the difference between $\left\|\mathbf Q(\mathbf{y};\tilde s,s,\Delta)\right\|_2$ and $\left\|\mathbf{y}\right\|_2$ decreases with $\tilde s$ and $s$, and $\left\|\mathbf Q(\mathbf y;\tilde s,s,\Delta)\right\|_2\rightarrow\left\|\mathbf y\right\|_2^2$ as $\tilde s\rightarrow\infty$ and $s\rightarrow\infty$ almost surely, although $\left\vert{y}_d\right\vert$, $d\in\mathcal{D}$ are quantized separately.\footnote{\clrtmp{We do not restrain the quantized version of $\left(\frac{\vert y_d\vert}{\left\|\mathbf{y}\right\|_2}\right)_{d\in\mathcal{D}}$ to be a unit vector in order not to destroy the unbiased property, which is essential for the convergence of GQFedWAvg as illustrated in Sec.~\ref{SubSec:Conv}.}}
}

\section{Algorithm \clrtmp{with Quantized Message Passing} and Convergence Analysis}\label{Sec:Alg}
In this section, we first present a general quantized FL algorithm, named GQFedWAvg.
Then, we analyze the convergence of GQFedWAvg.

\subsection{GQFedWAvg}
The proposed GQFedWAvg applies the proposed quantization scheme to quantize wisely chosen model update-related vectors and adopts a generalized mini-batch SGD method with the weighted average local model updates in global model aggregation.
GQFedWAvg has several adjustable parameters including the quantization parameters, i.e., $\tilde{\mathbf s}\triangleq\left(\tilde s_n\right)_{n\in\setNbar}\in\mathbb Z_{++}^{N+1}$, $\mathbf s\triangleq\left(s_n\right)_{n\in\setNbar}\in\mathbb Z_{++}^{N+1}$, and $\mathbf\Delta\triangleq\left(\Delta_n\right)_{n\in\setNbar}\in\mathbb{R}_{++}^{N+1}$,
and the parameters of the underlying generalized mini-batch SGD algorithm, i.e., $\mathbf{K}\triangleq\left(K_n\right)_{n\in\setNbar}\in\mathbb Z_{++}^{N+1}$,
$B\in\mathbb Z_{++}$, $\mathbf\Gamma\triangleq\left(\gamma^{(k_0)}\right)_{k_0\in\mathcal K_0}\in\mathbb R_{++}^{K_0}$ with $\mathcal K_0\triangleq\{1,2,\cdots,K_0\}$, and $\mathbf W\triangleq\left(W_n\right)_{n\in\mathcal N}\in(0,1)^{N}$ with $\mathbf 1^T\mathbf W=1$.
These parameters are collectively referred to as the algorithm parameters of GQFedWAvg in the rest of this paper.
Specifically, $K_0$ represents the number of global iterations;
$K_n$ represents the number of local iterations of worker $n\in\mathcal N$;
$B$ represents the local mini-batch size;
$\gamma^{(k_0)}$ represents the step size used within the $k_0$-th global iteration;
\clrtmp{$W_n$ represents the weight for worker $n\in\mathcal N$ in the weighted average of local model updates; $\tilde s_0,s_0\in\mathbb Z_{++}$, $\Delta_0\in\mathbb R_{++}$ and $\tilde s_n,s_n\in\mathbb Z_{++}$, $\Delta_n\in\mathbb R_{++}$
represent the quantization parameters for the server and worker $n\in\mathcal N$, respectively}.
\clrtmp{Besides, the other notations in Algorithm~\ref{Alg:Alg1} are defined below}.
$\mathcal K_n\triangleq\{1,2,\cdots,K_n\}$ denotes the local iteration index set for worker $n\in\mathcal N$.
For all $k_0\in\mathcal K_0$, $\hat{\mathbf x}^{(k_0)}\in\mathbb R^D$ denotes the global model recovered by all $N$ workers at the beginning of the $k_0$-th global iteration,
${\mathbf x}_n^{(k_0,0)}\in\mathbb R^D$ denotes the initial local model of worker $n\in\mathcal N$ at the beginning of the $k_0$-th global iteration,
and $\Delta\hat{\mathbf x}^{(k_0)}\in\mathbb R^D$ represents the weighted average of the quantized version of normalized accumulated local model updates, \clrtmp{which is also} termed the global model update, at the $k_0$-th global iteration.
For all $k_0\in\mathcal K_0,k_n\in\mathcal K_n,n\in\mathcal N$, $\mathbf x_n^{(k_0,k_n)}\in\mathbb R^D$
and $\mathcal B_n^{(k_0,k_n)}\subseteq\left\{\xi_i:i\in\mathcal I_n\right\}$ denote the local model of worker $n$
and the mini-batch used by worker $n$, respectively, at the $k_n$-th local iteration within the $k_0$-th global iteration.
\clrtmp{We present GQFedWAvg in Algorithm~\ref{Alg:Alg1}.}

\begin{algorithm}[t]
\caption{\clrtmp{GQFedWAvg}}
\label{Alg:Alg1}
{\bf{Input:}} $\mathbf K\in\mathbb Z_{++}^{N+1}$, $B\in\mathbb Z_{++}$, $\mathbf \Gamma\in\mathbb R_{++}^{K_0}$, \clrtmp{$\mathbf W\in(0,1)^{N}$, $\tilde{\mathbf s}$, $\mathbf s$, and $\mathbf\Delta\in\mathbb{R}_{++}^{N+1}$.}\\
{\bf{Output:}} $\mathbf x^*\left(\mathbf K,B,\mathbf\Gamma,\clrtmp{\mathbf W,\tilde{\mathbf s},\mathbf s,\mathbf\Delta}\right)$.

\begin{algorithmic}[1]
{\small
\STATE {\bf{Initialize:}} The server generates the global model $\mathbf x_0^{(0)}$, sets $\Delta\hat{\mathbf x}^{(0)}=\mathbf x_0^{(0)}$, and sends the quantized result \clrtmp{$\mathbf Q(\clrtmp{\frac{\Delta\hat{\mathbf x}^{(0)}}{\sum_{n\in\mathcal N}W_nK_n}};\tilde s_0,s_0,\Delta_0)$} to all $N$ workers.\label{Step:initialization}
The $N$ workers set the recovered global model $\hat{\mathbf x}^{(0)}=0$.
\FOR {$k_0=1,2,\cdots,K_0$}
    \FOR {worker $n\in{\mathcal N}$}\label{Step:begin_global_iteration}
        \STATE Compute $\hat{\mathbf x}^{(k_0)}$ according to:
        \begin{align}\label{eq:RecoveredLocalModel}
        &\hat{\mathbf x}^{(k_0)}:=\hat{\mathbf x}^{(k_0-1)}\nonumber\\
        &+\gamma^{(k_0-1)}\mathbf Q\!\left(\clrtmp{\frac{\Delta\hat{\mathbf x}^{(k_0-1)}}{\sum\limits_{n\in\mathcal N}\!\!W_nK_n}};\clrtmp{\tilde s_0,s_0,\Delta_0}\!\!\right)\clrtmp{\!\!\sum_{n\in\mathcal N}\!W_nK_n},
        \end{align}
        and set its local model ${\mathbf x}_n^{(k_0,0)}=\hat{\mathbf x}^{(k_0)}$.\label{Step:initialize_local_model}
        \FOR {$k_n=1,2,\cdots,K_n$}\label{Step:begin_local_iteration}
            \STATE Randomly select a mini-batch $\mathcal B_n^{(k_0,k_n)}\subseteq\left\{\xi_i:i\in\mathcal I_n\right\}$ and update its local model ${\bf x}_n^{(k_0,k_n)}$ according to:
            \begin{align}\label{eq:local_update}
            &{\mathbf x}_n^{(k_0,k_n)}:={\mathbf x}_n^{(k_0,k_n-1)}\nonumber\\
            &-\frac{\gamma^{(k_0)}}{B}\sum_{\xi\in\mathcal B_n^{(k_0,k_n)}}\nabla{F\left({\mathbf x}_n^{(k_0,k_n-1)};\xi\right)}.
            \end{align}
        \ENDFOR\label{Step:end_local_iteration}
        \STATE Compute \clrtmp{$\frac{{\mathbf x}_n^{(k_0,K_n)}-\hat{\mathbf x}^{(k_0)}}{\gamma^{(k_0)}K_n}$}, and send the quantized result $\mathbf Q\left(\clrtmp{\frac{{\mathbf x}_n^{(k_0,K_n)}-\hat{\mathbf x}^{(k_0)}}{\gamma^{(k_0)}K_n}};\tilde s_n,s_n,\Delta_n\right)$ to the server.\label{Step:upload}
    \ENDFOR\label{Step:end_global_iteration}
    \STATE The server computes the \clrtmp{global model update} $\Delta\hat{\mathbf x}^{(k_0)}$ according to:
    \begin{align}\label{eq:GlobalUpdate}
    \Delta\hat{\mathbf x}^{(k_0)}:=&\clrtmp{\sum_{n\in\mathcal N}W_nK_n}\mathbf Q\left(\frac{{\mathbf x}_n^{(k_0,K_n)}-\hat{\mathbf x}^{(k_0)}}{\gamma^{(k_0)}K_n};\tilde s_n,s_n,\Delta_n\right),
    \end{align}
    compute $\frac{\Delta\hat{\mathbf x}^{(k_0)}}{\sum_{n\in\mathcal N}W_nK_n}$, and sends the quantized result $\mathbf Q(\clrtmp{\frac{\Delta\hat{\mathbf x}^{(k_0)}}{\sum_{n\in\mathcal N}W_nK_n}};\clrtmp{\tilde s_0,s_0,\Delta_0})$ to all $N$ workers.\label{Step:broadcast}
\ENDFOR\label{Step:end_global_iteration}
\STATE The server and all $N$ workers compute $\hat{\mathbf x}^{(K_0+1)}$ according to \eqref{eq:RecoveredLocalModel}, and set $\mathbf x^*\left(\mathbf K,B,\mathbf\Gamma,\clrtmp{\mathbf W,\tilde{\mathbf s},\mathbf s,\mathbf\Delta}\right)=\hat{\mathbf x}^{(K_0+1)}$.\label{step:global_average}}
\end{algorithmic}
\end{algorithm}

The properties of GQFedWAvg and its distinctions from GenQSGD in our previous work \cite{GenQSGD} are summarized as follows.
\begin{itemize}
  \item GQFedWAvg applies the quantization scheme in Sec.~\ref{Sec:Quantization} to wisely chosen model update-related vectors, aiming to effectively utilize the communication resources and facilitate the choices of the algorithm parameters.
      Specifically, in Algorithm~\ref{Alg:Alg1}, we quantize the scaled model update-related vectors
      $\frac{{\mathbf x}_n^{(k_0,K_n)}-\hat{\mathbf x}^{(k_0)}}{\gamma^{(k_0)}K_n}$
      (which will be shown to be bounded in Lemma~\ref{Lem:LocalUpdateBound} and no longer scale with $K_n$ and $\gamma^{(k_0)}$),
      $n\in\mathcal{N}$ and $\frac{\Delta\hat{\mathbf x}^{(k_0)}}{\sum_{n\in\mathcal N}W_nK_n}$
      (which will be shown to be bounded in Lemma~\ref{Lem:LocalUpdateBound} and no longer scale with $K_n$ and $W_n$),
      rather than the original unscaled model update-related vectors ${\mathbf x}_n^{(k_0,K_n)}-\hat{\mathbf x}^{(k_0)}$, $n\in\mathcal{N}$ and $\Delta\hat{\mathbf x}^{(k_0)}$, for all $k_0\in\mathcal{K}_0$.
      Thus, we can first choose the input ranges $\mathbf\Delta$ according to the upper bounds and then choose the parameters of the underlying mini-batch SGD including $\mathbf{K}$, $B$, $\mathbf\Gamma$, and $\mathbf{W}$.
      In contrast, GenQSGD only \clrtmp{quantizes} the normalized vectors $\frac{{\mathbf x}_n^{(k_0,K_n)}-\hat{\mathbf x}^{(k_0)}}{\left\|{\mathbf x}_n^{(k_0,K_n)}-\hat{\mathbf x}^{(k_0)}\right\|_2}$ and $\frac{\Delta\hat{\mathbf x}^{(k_0)}}{\left\|\Delta\hat{\mathbf x}^{(k_0)}\right\|_2}$ \clrtmp{and assumes} the exact values of the magnitudes $\left\|{\mathbf x}_n^{(k_0,K_n)}-\hat{\mathbf x}^{(k_0)}\right\|_2$ and $\left\|\Delta\hat{\mathbf x}^{(k_0)}\right\|_2$ are transmitted with finite bits for simplicity.
      The advantage of quantizing the scaled model update-related vectors over quantizing the original unscaled ones will be experimentally demonstrated in Sec.~\ref{SubSec:ExperimentalResults}.
  \item GQFedWAvg adopts the general weighted average of local model updates in global model aggregation to weigh the local model updates of the workers differently if they utilize distinct computing and communication resources.
      The goal is to improve the convergence performance of GQFedWAvg.
      Differently, GenQSGD adopts the average of local model updates in global model aggregation, which may not effectively utilize the computing and communication resources at the workers to improve the convergence performance.
      The advantage of allowing the general weighted average over restricting to the average will be numerically shown in Sec.~\ref{SubSec:ExperimentalResults}.
  \item GQFedWAvg allows flexible choices for algorithm parameters $\mathbf{K}$, $B$, $\mathbf\Gamma$, $\mathbf{W}$, $\tilde{\mathbf{s}}$, and $\mathbf{s}$ to adequately adapt to the uniform or nonuniform computing and communication resources at the workers.
      Specifically, the numbers of global and local iterations $\mathbf{K}$
      can be chosen to sufficiently adapt to the computing \clrtmp{resources at} the server and all workers;
      the quantization parameters $\tilde{\mathbf{s}}$ and $\mathbf{s}$ can be selected to adequately adapt to the communication \clrtmp{resources at} the server and all workers; \clrtmp{and $\mathbf{W}$ can be chosen to adapt to both the computing and communication resources at the workers}.
      \clrtmp{In addition, the step size sequence $\mathbf\Gamma$ and the mini-batch size $B$ can be chosen to further improve the convergence performance of GQFedWAvg.}
      The advantage of flexibly adapting the algorithm parameters to the computing and communication resources at the server and workers will be numerically and experimentally illustrated in Sec.~\ref{SubSec:ExperimentalResults}.
\end{itemize}

In the following, we discuss how to choose $\mathbf\Delta$ for all the random quantizers used in GQFedWAvg.
Before that, we derive upper bounds on the norms of the chosen model update-related vectors in Step~\ref{Step:upload} and Step~\ref{Step:broadcast} under the following assumption.
\begin{Asump}[Bounded Gradient of Loss Function]\label{Asump:BoundedNorm}
There exists a constant $R\in\mathbb{R}_{++}$ such that $\left\|\nabla{F\left({\mathbf x};\xi_i\right)}\right\|_2\leq R$ for all $ \mathbf x\in\mathbb R^D$, $i\in\mathcal{I}_n$, $n\in\mathcal N$.\footnote{Assumption~\ref{Asump:BoundedNorm}, which is stronger than the assumptions of bounded gradient variance and bounded
gradient moment, is needed to achieve the convergence results for general scenarios as illustrated in Sec.~\ref{sec:Introduction}.
Note that the boundedness condition is
easily satisfied in numerical experiments \cite{SSCA}.}
\end{Asump}
Under Assumption~\ref{Asump:BoundedNorm}, we have the following result.
\begin{Lem}[Bounded Quantization Inputs]\label{Lem:LocalUpdateBound}
Under Assumption~\ref{Asump:BoundedNorm}, the inputs of $\mathbf Q(\cdot;\tilde s_n,s_n,\Delta_n)$, $n\in\setNbar$ satisfy:
\begin{align}
\left\|\frac{{\mathbf x}_n^{(k_0,K_n)}-\hat{\mathbf x}^{(k_0)}}{\gamma^{(k_0)}K_n}\right\|_2
&\leq R,\ k_0\in\mathcal K_0,\ n\in\mathcal N,\label{NormBound1}\\
\left\|\frac{\Delta\hat{\mathbf x}^{(k_0)}}{\sum_{n\in\mathcal N}W_nK_n}\right\|_2
&\leq\left(R+1\right)\left(1+\sqrt{D}\right),\ k_0\in\mathcal K_0.\label{NormBound2}
\end{align}
\end{Lem}
\begin{IEEEproof}
Note that $\left\|{\mathbf x}_n^{(k_0,K_n)}-\hat{\mathbf x}^{(k_0)}\right\|_2
\overset{(a)}{=}\left\|\gamma^{(k_0)}\sum_{k_n\in\mathcal K_n}\frac{1}{B}\sum_{\xi\in\mathcal B_n^{(k_0,k_n)}}\nabla{F({\mathbf x}_n^{(k_0,k_n-1)};\xi)}\right\|_2
\overset{(b)}{\leq}\gamma^{(k_0)}\sqrt{\frac{K_n}{B}\sum_{k_n\in\mathcal K_n}\sum_{\xi\in\mathcal B_n^{(k_0,k_n)}}\left\|\nabla{F({\mathbf x}_n^{(k_0,k_n-1)};\xi)}\right\|_2^2}
\overset{(c)}\leq\gamma^{(k_0)}K_nR$ for all $k_0\in\mathcal K_0$ and $n\in\mathcal N$,
where (a) follows from \eqref{eq:local_update}, (b) follows from the inequality $\|\sum_{i=1}^n{\bf z}_i\|^2\leq n\sum_{i=1}^n{\|{\bf z}_i\|^2}$, and (c) follows from Assumption~\ref{Asump:BoundedNorm}.
Thus, we readily show \eqref{NormBound1}.
Moreover, we have
$\left\|\Delta\hat{\mathbf x}^{(k_0)}\right\|_2
\overset{(d)}{=}\left\|\sum\limits_{n\in\mathcal N}W_nK_n\mathbf Q\left(\frac{{\mathbf x}_n^{(k_0,K_n)}-\hat{\mathbf x}^{(k_0)}}{\gamma^{(k_0)}K_n};\tilde s_n,s_n,\Delta_n\right)\right\|_2
\overset{(e)}{\leq}\sum\limits_{n\in\mathcal N}W_nK_n\left\|\mathbf Q\left(\frac{{\mathbf x}_n^{(k_0,K_n)}-\hat{\mathbf x}^{(k_0)}}{\gamma^{(k_0)}K_n};\tilde s_n,s_n,\Delta_n\right)\right\|_2
\overset{(f)}{\leq}\sum_{n\in\mathcal{N}}W_nK_n\left(R+\frac{1}{\tilde{s}_n}\right)\left(1+\frac{\sqrt{D}}{s_n}\right)
\overset{(g)}{\leq}\left(R+1\right)\left(1+\sqrt{D}\right){\sum_{n\in\mathcal N}W_nK_n}$,
where (d) follows from \eqref{eq:GlobalUpdate}, (e) follows from the triangle inequality, (f) follows from Lemma~\ref{Lem:Quantization} (iii), and (g) follows from $\tilde{s}_n,s_n\in\mathbb{Z}_{++}$, $n\in\mathcal{N}$.
Thus, we readily show \eqref{NormBound2}.\end{IEEEproof}
\clrtmp{Based on Lemma~\ref{Lem:LocalUpdateBound}, we can choose $\mathbf\Delta$ according to:
\begin{align}\label{eq:Delta_n}
\Delta_n&=R,\ n\in\mathcal{N},\\
\Delta_0&=\left(R+1\right)\left(1+\sqrt{D}\right).
\end{align}
\clrtmp{The remaining algorithm parameters can be flexibly chosen.
Later in Sec.~\ref{sec:Optimization}, we will show how to select those parameters from an optimization point of view.}}
\begin{Rmk}[Generality of \clrtmp{GQFedWAvg}]
\clrtmp{GQFedWAvg} is general in the sense that it includes some existing algorithms as special cases.
For the sake of discussion,
we let \clrtmp{$\tilde s=s=\infty$} \clrtmp{represent} the case without quantization.
In particular, \clrtmp{GQFedWAvg} with $K_n=1,n\in\mathcal N$ \clrtmp{and $W_n=\frac{1}{N},n\in\mathcal N$ for $\tilde s_n=s_n=\infty,n\in\setNbar$} reduces to PM-SGD \cite{PMSGD};
\clrtmp{GQFedWAvg} with $K_n=l\frac{I_n}{B},l\in\mathbb Z_{++},n\in\mathcal N$ \clrtmp{and $W_n=\frac{1}{N},n\in\mathcal N$ for $\tilde s_n=s_n=\infty,n\in\setNbar$} reduces to FedAvg \cite{FedAvg};
\clrtmp{and GQFedWAvg with $W_n=\frac{1}{N},n\in\mathcal N$ for $\tilde s_n=\infty$, $n\in\setNbar$ reduces to GenQSGD in our previous work \cite{GenQSGD}}.
\end{Rmk}

\subsection{Convergence Analysis}\label{SubSec:Conv}
In this part, we analyze the convergence of GQFedWAvg.
Note that the convergence analysis of GQFedWAvg is more complicated than that of GenQSGD in \cite{GenQSGD} for the following reasons.
(i) The quantization method proposed in Sec.~\ref{Sec:Quantization} quantizes $\left\|\frac{{\mathbf x}_n^{(k_0,K_n)}-\hat{\mathbf x}^{(k_0)}}{\gamma^{(k_0)}K_n}\right\|_2$ and $\left\|\frac{\Delta\hat{\mathbf x}^{(k_0)}}{\sum_{n\in\mathcal N}W_nK_n}\right\|_2$ in addition to $\frac{{\mathbf x}_n^{(k_0,K_n)}-\hat{\mathbf x}^{(k_0)}}{\left\|{\mathbf x}_n^{(k_0,K_n)}-\hat{\mathbf x}^{(k_0)}\right\|_2}$ and $\frac{\Delta\hat{\mathbf x}^{(k_0)}}{\left\|\Delta\hat{\mathbf x}^{(k_0)}\right\|_2}$ for all $n\in\mathcal{N}$ and $k_0\in\mathcal{K}_0$, which introduces additional quantization errors and thus complicates the convergence analysis.
(ii) It is necessary to characterize the influence of the weights in global model aggregation, in addition to the other algorithm parameters $\mathbf{K}$, $B$, $\mathbf\Gamma$, $\tilde{\mathbf{s}}$, $\mathbf{s}$, and $\mathbf\Delta$, on the convergence error.
In the rest of this paper, we assume that the following typical assumptions are satisfied \cite{YuHao,GenQSGD}.
\begin{Asump}[I.I.D. Samples]\label{Asump:IID}
$\zeta_n,n\in\mathcal N$ are I.I.D..\footnote{The convergence analysis of FL algorithms for ML problems with non-i.i.d. data is beyond the scope of this paper and will be studied in our future work.}
\end{Asump}
\begin{Asump}[Smoothness]\label{Asump:Smoothness}
For all $n\in\mathcal N$, $f_n({\mathbf x})$ is continuously differentiable, and its gradient is Lipschitz continuous, i.e., there exists a constant $L>0$ such that $\left\|\nabla{f_n({\mathbf x})}-\nabla{f_n({\mathbf y})}\right\|_2 \leq L\left\|{\mathbf x}-{\mathbf y}\right\|_2$, for all $\mathbf x,\mathbf y\in\mathbb R^D$.
\end{Asump}
\begin{Asump}[Bounded Variances]\label{Asump:BoundedVariances}
For all $n\in\mathcal N$, there exists a constant $\sigma>0$ such that $\mathbb E\left[\left\|\nabla{F\left({\mathbf x};\zeta_n\right)}-\nabla{f_n({\mathbf x})}\right\|_2^2\right]\leq{\sigma^2}$, for all $\mathbf x\in\mathbb R^D$.
\end{Asump}
The convergence of GQFedWAvg is summarized below.
\begin{Thm}[Convergence of GQFedWAvg]\label{Thm:Convergence}
Suppose that Assumptions~\ref{Asump:BoundedNorm}, \ref{Asump:IID}, \ref{Asump:Smoothness}, and \ref{Asump:BoundedVariances} are satisfied, and the step size $\mathbf\Gamma$ and weights $\mathbf{W}$ satisfy:
\begin{align}
&1-L^2\left(\gamma^{(k_0)}\right)^2K_n-L\gamma^{(k_0)}\left(1+q_{s_0}\right)\left(N+q_{s_n}\right)W_nK_n\geq0,\nonumber\\
&\hspace{130pt}\ k_0\in\mathcal K_0,\ n\in\mathcal N,\label{eq:Cons_gamma}\\
&\mathbf 1^T\mathbf W=1.\label{eq:Cons_W}
\end{align}
Then, for all $\mathbf K\in\mathbb Z_{++}^{N+1}$, $B\in\mathbb Z_{++}$,
$\mathbf\Gamma$ satisfying \eqref{eq:Cons_gamma},
$\mathbf W$ satisfying \eqref{eq:Cons_W},
$\tilde{\mathbf s}\in\mathbb Z_{++}^{N+1}$,
$\mathbf s\in\mathbb Z_{++}^{N+1}$,
and $\mathbf\Delta\in\mathbb{R}_{++}^{N+1}$,
$\left\{\hat{\mathbf x}^{(k_0)}:k_0\in\mathcal K_0\right\}$
generated by GQFedWAvg satisfies:
\begin{align}
&\frac{\sum_{k_0\in\mathcal K_0}\gamma^{(k_0)}\mathbb E\left[\left\|\nabla f\left(\hat{\mathbf x}^{(k_0)}\right)\right\|^2\right]}{\sum_{k_0\in\mathcal K_0}\gamma^{(k_0)}}
\leq C\left(\mathbf K,B,\mathbf\Gamma,\mathbf W,\tilde{\mathbf s},\mathbf s\right),
\end{align}
where
\begin{align}\label{eq:Convergence_RHS}
&C\left(\mathbf K,B,\mathbf\Gamma,\mathbf W,\tilde{\mathbf s},\mathbf s\right)
\triangleq\frac{2\left(f\left(\hat{\mathbf x}^{(1)}\right)-f^*\right)}{\sum_{n\in\mathcal N}W_nK_n\sum_{k_0\in\mathcal K_0}\gamma^{(k_0)}}\nonumber\\
&+\frac{L^2\sigma^2\sum\limits_{n\in\mathcal N}W_nK_n(K_n+1)\sum\limits_{k_0\in\mathcal K_0}\left(\gamma^{(k_0)}\right)^3}{2B\sum\limits_{n\in\mathcal N}W_nK_n\sum\limits_{k_0\in\mathcal K_0}\gamma^{(k_0)}}\nonumber\\
&+\frac{L\sigma^2N\sum\limits_{n\in\mathcal N}W_n^2 K_n\sum\limits_{k_0\in\mathcal K_0}\left(\gamma^{(k_0)}\right)^2}{B\sum\limits_{n\in\mathcal N}W_nK_n\sum\limits_{k_0\in\mathcal K_0}\gamma^{(k_0)}}\nonumber\\
&+\frac{L\sigma^2Nq_{s_0}\sum\limits_{n\in\mathcal{N}}W_n^2K_n
\sum\limits_{k_0\in\mathcal{K}_0}\left(\gamma^{(k_0)}\right)^2}
{B\sum\limits_{n\in\mathcal{N}}W_nK_n\sum\limits_{k_0\in\mathcal{K}_0}\gamma^{(k_0)}}\nonumber\\
&+\frac{L\sigma^2\left(1+q_{s_0}\right)\sum\limits_{n\in\mathcal N}q_{s_n}W_n^2 K_n\sum\limits_{k_0\in\mathcal K_0}\left(\gamma^{(k_0)}\right)^2}{B\sum\limits_{n\in\mathcal N}W_nK_n\sum\limits_{k_0\in\mathcal K_0}\gamma^{(k_0)}}\nonumber\\
&+\frac{Lq_{\tilde s_0,s_0}\Delta_0^2\sum\limits_{n\in\mathcal N}W_nK_n\sum\limits_{k_0\in\mathcal K_0}\left(\gamma^{(k_0)}\right)^2}{\sum_{k_0\in\mathcal K_0}\gamma^{(k_0)}}\nonumber\\
&+\frac{L\left(1+ q_{s_0}\right)\sum_{n\in\mathcal N}q_{\tilde s_n,s_n}W_n^2K_n^2\Delta_n^2\sum\limits_{k_0\in\mathcal K_0}\left(\gamma^{(k_0)}\right)^2}{\sum_{n\in\mathcal N}W_nK_n\sum_{k_0\in\mathcal K_0}\gamma^{(k_0)}}.
\end{align}
Here,
$q_{\tilde s_n,s_n}\!,q_{s_n}\!,n\!\in\!\setNbar$ are given by \eqref{eq:qs},
and $f^*$ denotes the optimal value of the problem in \eqref{Obj_ML}.\footnote{The parameters $L$ and $\sigma$ and a lower bound of $f^*$ can be estimated from the training data.
Specifically, each worker $n\in\mathcal{N}$ estimates an upper bound of $\frac{\left\|\nabla{f_n({\mathbf x})}-\nabla{f_n({\mathbf y})}\right\|_2}{\left\|{\mathbf x}-{\mathbf y}\right\|_2}$ and $\mathbb E\left[\left\|\nabla{F\left({\mathbf x};\zeta_n\right)}-\nabla{f_n({\mathbf x})}\right\|_2^2\right]$, denoted as $L_n$ and $\sigma_n$, respectively, and estimates a lower bound of $f_n(\mathbf{x})$, denoted as $\hat{f}_n$, based on its local data and sends them to the server.
The server chooses $L\triangleq\max_{n\in\mathcal{N}}L_n$, $\sigma\triangleq\max_{n\in\mathcal{N}}\sigma_n$, and $f^*\triangleq\max_{n\in\mathcal{N}}\hat{f}_n$.}
\end{Thm}
\begin{IEEEproof}See Appendix~B.
\end{IEEEproof}
Theorem~1 indicates that the algorithm parameters $\mathbf K$, $B$, $\mathbf\Gamma$, $\mathbf W$, $\tilde{\mathbf s}$, and $\mathbf s$ influence the convergence of GQFedWAvg.
In the following, we illustrate how the seven terms on the R.H.S of (17) change with $\mathbf K$, $B$, $\mathbf\Gamma$, $\mathbf W$, $\tilde{\mathbf s}$, and $\mathbf s$.
The first term decreases with $K_n$, $n\in\mathcal N$, $W_n$, $n\in\mathcal N$, and $\gamma^{(k_0)}$, $k_0\in\mathcal K_0$.
The second to fifth terms decrease with $B$ due to the decrease of the variance of a stochastic gradient and vanishes as $B\rightarrow\infty$, implying that these terms are caused by the approximations of the expected gradients with the sample average gradients.
The third to seventh terms decrease with $\tilde{s}_n,n\in\setNbar$ and $s_n,n\in\setNbar$ (i.e., the quantization error) and vanishes as $\tilde{s}_n\rightarrow\infty,n\in\setNbar$ and $s_n\rightarrow\infty,n\in\setNbar$, implying that these terms are caused by the quantized message passing.
For any given $K_n,n\in\mathcal{N}$, $B$, $\mathbf{W}$, $\mathbf{\tilde{s}}$, and $\mathbf{s}$ and $\boldsymbol{\Gamma}$ satisfying $\sum_{k_0=1}^{\infty}\gamma^{(k_0)}=\infty$ and $\sum_{k_0=1}^{\infty}\left(\gamma^{(k_0)}\right)^2<\infty$ (i.e., non-sumable and square-sumable), all the terms vanishes as $K_0\rightarrow\infty$, implying that with non-sumable and square-sumable step size sequence, the influences of the gradient variances and quantization errors gradually disappear as GQFedWAvg proceeds. In other words, $\left\{\hat{\mathbf x}^{(k_0)}:k_0\in\mathcal K_0\right\}$ generated by GQFedWAvg is guaranteed to converge to a stationary point of the problem in (2), which consist with the convergence results of existing FL algorithms \cite{GenQSGD,LargeScaleML}.

\section{\clrtmp{Optimization of Algorithm Parameters}}\label{sec:Optimization}
In this section, we first introduce performance metrics for implementing \clrtmp{GQFedWAvg} in the edge computing system.
Then, \clrtmp{we optimize the algorithm parameters of GQFedWAvg to maximally adapt to the computing and communication resources at the server and workers}.

\subsection{Performance Metrics}\label{SubSec:PerformanceMetrics}
\clrtmp{We consider the same performance metrics as in \cite{GenQSGD}, i.e., time cost, energy cost, and convergence error, which are briefly illustrated below for completeness}.
Let $C_n$ denote the number of CPU cycles required for worker $n\in\mathcal N$ to compute $\nabla F({\mathbf x};\xi_i)$ for all $\mathbf x\in\mathbb R^{D}$ and $i\in\mathcal I_n$,
$C_0$ denote the number of CPU cycles needed for the server to compute one global model update,
and $\alpha_0$ and $\alpha_n$ denote the constant factors determined by the switched capacitances of the server and worker $n\in\mathcal N$, respectively \cite{capacitance}.
The communication time $T_{\text{comm}}(K_0,\tilde{\mathbf s},\mathbf s)$, communication energy $E_{\text{comm}}(K_0,\tilde{\mathbf s},\mathbf s)$, computing time $T_{\text{comp}}(\mathbf K,B)$, and computing energy $E_{\text{comp}}(\mathbf K,B)$ are given by \cite[Sec.IV-A]{GenQSGD}:
\begin{align*}
&T_{\text{comm}}(K_0,\tilde{\mathbf s},\mathbf s)
\!=\!K_0\!\!\left(\!\max_{n\in\mathcal N}\frac{\log_2(\tilde s_n\!+\!1)\!+\!D\!\left(\log_2(s_n\!+\!1)\!+\!1\right)}{r_n}\right.\\
&\hspace{70pt}\left.+\frac{\log_2(\tilde s_0+1)+D\left(\log_2(s_0+1)+1\right)}{r_0}\right),\\
&E_{\text{comm}}(K_0,\tilde{\mathbf s},\mathbf s)\nonumber\\
&\hspace{20pt}=K_0\sum_{n\in\setNbar}\frac{p_n\left(\log_2(\tilde s_n+1)+D\left(\log_2(s_n+1)+1\right)\right)}{r_n},\\
&T_{\text{comp}}(\mathbf K,B)
=K_0\left(B\max_{n\in\mathcal N}\frac{C_n}{F_n}K_n+\frac{C_0}{F_0}\right),\\
&E_{\text{comp}}(\mathbf K,B)
=K_0\left(B\sum_{n\in\mathcal N}\alpha_nC_nF_n^2K_n+\alpha_0C_0F_0^2\right).
\end{align*}
We use the sum of the communication and computing times $T(\mathbf K,B,\tilde{\mathbf s},\mathbf s)$ and the sum of the communication and computing energies $E(\mathbf K,B,\tilde{\mathbf s},\mathbf s)$ to measure the time cost and energy cost, respectively, for implementing GQFedWAvg, which are given by \cite[(17), (18)]{GenQSGD}:\footnote{The proposed framework can be easily extended to separate communication and computing costs.}
\begin{align}
&T(\mathbf K,B,\tilde{\mathbf s},\mathbf s)
=T_{\text{comm}}(K_0,\tilde{\mathbf s},\mathbf s)
+T_{\text{comp}}(\mathbf K,B),\label{eq:Time}\\
&E(\mathbf K,\!B,\!\tilde{\mathbf s},\!\mathbf s)
=E_{\text{comm}}(K_0,\tilde{\mathbf s},\mathbf{s})
+E_{\text{comp}}(\mathbf K,B).\label{eq:Energy}
\end{align}
We use $C\!\left(\mathbf K,\!B,\!\mathbf\Gamma\!,\!\mathbf W,\!\tilde{\mathbf s},\!\mathbf s\right)$ given by \eqref{eq:Convergence_RHS} to measure the convergence error of GQFedWAvg \cite{GenQSGD}.

\subsection{Problem Formulation}\label{SubSec:Formulation}
We optimize the algorithm parameters $\mathbf K\in\mathbb Z_{++}^{N+1}$, $B\in\mathbb Z_{++}$, $\mathbf\Gamma$ satisfying \eqref{eq:Cons_gamma},
\clrtmp{$\mathbf W\in(0,1)^N$ satisfying \eqref{eq:Cons_W}, $\tilde{\mathbf s}\in\mathbb Z_{++}^{N+1}$ and $\mathbf s\in\mathbb Z_{++}^{N+1}$} to minimize the \clrtmp{convergence error $C\left(\mathbf K,B,\mathbf\Gamma,\mathbf W,\tilde{\mathbf s},\mathbf s\right)$ subject to time cost, energy cost, step size sequence, and weight constraints}.\footnote{The three metrics introduced in Sec.~\ref{SubSec:PerformanceMetrics} can be considered in the optimization objective or constraints, depending on the specific requirements.
Specifically, performance metrics with firm requirements (if they exist) can be formulated into constraints, and performance metrics expected to be freely optimized can be incorporated into the objective function.
The problem is assumed to be feasible as in \cite{ResAlls2-1,ResAlls3-2}, otherwise it is not practically interesting or meaningful.
The proposed optimization framework is applicable for various formulations and can be used to optimize the parameters of existing FL algorithms, such as PM-SGD, FedAvg, and PR-SGD.}
For tractability, \clrtmp{as in \cite{FastFL,GenQSGD}}, we relax the integer constraints, $\mathbf K\in\mathbb Z_{++}^{N+1}$, $B\in\mathbb Z_{++}$, \clrtmp{$\tilde{\mathbf s}\in\mathbb Z_{++}^{N+1}$, and $\mathbf s\in\mathbb Z_{++}^{N+1}$}, to their continuous counterparts, $\mathbf K\succ\mathbf 0$, $B>0$, \clrtmp{$\tilde{\mathbf s}\succ\mathbf 0$, and $\mathbf s\succ\mathbf 0$}, respectively.
Note that a nearly optimal point satisfying the original integer constraints can be easily constructed based on an optimal point of a relaxed problem.
\clrtmp{\begin{Prob}[Optimization of Algorithm Parameters]\label{Prob:Optimization}
\begin{align}
\min_{\substack{\mathbf K,\mathbf\Gamma,\mathbf W,\tilde{\mathbf s},\mathbf s\succ\mathbf 0,B>0}}
&{\quad}C\left(\mathbf K,B,\mathbf\Gamma,\mathbf W,\tilde{\mathbf s},\mathbf s\right)\nonumber\\
\mathrm{s.t.}&{\quad}\eqref{eq:Cons_gamma},\ \eqref{eq:Cons_W},\nonumber\\
&{\quad}T(\mathbf K,B,\tilde{\mathbf s},\mathbf s)\leq T_{\max},\label{eq:Cons_time}\\
&{\quad}E(\mathbf K,B,\tilde{\mathbf s},\mathbf s)\leq E_{\max},\label{eq:Cons_energy}
\end{align}
where $T_{\max}$ and $E_{\max}$ denote the limits on the time cost and energy cost, respectively.
\end{Prob}}
Problem~\ref{Prob:Optimization} can be solved by the server (in an offline manner) before the implementation of \clrtmp{GQFedWAvg}.
\clrtmp{Problem~\ref{Prob:Optimization} is a challenging non-convex problem}.
This is because: (i) the dimension of $\mathbf\Gamma$, i.e., $K_0$, is an optimization variable;
(ii) the constraint in \eqref{eq:Cons_time} contains non-differentiable max functions;
\clrtmp{(iii) \eqref{eq:Cons_W} is a posynomial equality constraint, and \eqref{eq:Cons_time} involves the product of an optimization variable and additive log terms, making Problem~\ref{Prob:Optimization} more complicated than \cite[Problem~9]{GenQSGD};
(iv) the objective function to be minimized involves min functions, which are difficult to tackle compared to max functions.}
It is noteworthy that Challenges (i) and (ii) are similar to those for solving \cite[Problem~9]{GenQSGD}, whereas \clrtmp{Challenges (iii) and (iv) are additional and have to be addressed using more delicate tricks.}

\subsection{Solution}\label{SubSec:Solution}
In this part,
\clrtmp{we transform Problem~\ref{Prob:Optimization} to an equivalent problem to address Challenges (i)-(iv) mentioned in Sec.~\ref{SubSec:Formulation} and adopt GIA \cite{GIA} to obtain a KKT point of the equivalent problem.}
\clrtmp{First, to address Challenge (i), we consider the following problem \cite{GenQSGD}}.
\begin{Prob}[Convergence Error Minimization]\label{Prob:step_size_sequence}
For any given $\mathbf K\in\mathbb Z_{++}^{N+1}$, $B\in\mathbb Z_{++}$, \clrtmp{$\mathbf W\in(0,1)^N$ satisfying $\mathbf 1^T\mathbf W=1$, $\tilde{\mathbf s}\in\mathbb Z_{++}^{N+1}$, $\mathbf s\in\mathbb Z_{++}^{N+1}$}, and $S\in\mathbb{R}_{++}$,
\begin{align}
\min_{\mathbf\Gamma}&{\quad}\clrtmp{C\left(\mathbf K,B,\mathbf\Gamma,\mathbf W,\tilde{\mathbf s},\mathbf s\right)}\nonumber\\
\mathrm{s.t.}&{\quad}\eqref{eq:Cons_gamma},\nonumber\\
&{\quad}\sum_{k_0\in\mathcal K_0}\gamma^{(k_0)}=S.\label{eq:sum_constant}
\end{align}
\end{Prob}
\begin{Lem}[Optimal Step Size Sequence $\mathbf\Gamma$]\label{Lem:optimal_step_size}
\clrtmp{For any $S\in\mathbb{R}_{++}$ such that Problem~\ref{Prob:step_size_sequence} is feasible}, then an optimal solution of Problem~\ref{Prob:step_size_sequence} is $\frac{S}{K_0}\mathbf1$.
\end{Lem}
\begin{IEEEproof}
\clrtmp{First, we show that $\frac{S}{K_0}\mathbf1$ is feasible for Problem~\ref{Prob:step_size_sequence}.
Note that $\frac{S}{K_0}\mathbf1$ satisfies \eqref{eq:sum_constant} obviously.
It remains to show that $\frac{S}{K_0}\mathbf1$ satisfies \eqref{eq:Cons_gamma}.
For any given $\mathbf K\in\mathbb Z_{++}^{N+1}$, $\mathbf W\in(0,1)^N$ satisfying $\mathbf 1^T\mathbf W=1$, and $\mathbf s\in\mathbb Z_{++}^{N+1}$, denote $\mathcal{G}_{\mathbf K,\mathbf W,\mathbf s}\triangleq\{\gamma~\vert~1-L^2K_n\gamma^2-L(1+q_{s_0})(N+q_{s_n})W_nK_n\gamma\geq0,\ n\in\mathcal N\}$, which is a convex set.
Let $\hat{\mathbf\Gamma}\triangleq\left(\hat{\gamma}^{(k_0)}\right)_{k_0\in\mathcal{K}_0}$ denote an arbitrary feasible point, i.e., $\hat{\gamma}^{(k_0)}\in\mathcal{G}_{\mathbf K,\mathbf W,\mathbf s}$ for all $k_0\in\mathcal{K}_0$ and $\sum_{k_0\in\mathcal{K}_0}\hat{\gamma}^{(k_0)}=S$.
Then, $\frac{S}{K_0}=\frac{\sum_{k_0\in\mathcal{K}_0}\hat{\gamma}^{(k_0)}}{K_0}\in\mathcal{G}_{\mathbf K,\mathbf W,\mathbf s}$, implying that $\frac{S}{K_0}\mathbf1$ satisfies \eqref{eq:Cons_gamma}.
Thus, $\frac{S}{K_0}\mathbf1$ is feasible for Problem~\ref{Prob:step_size_sequence}.
Next, following the proof of \cite[Lemma~4]{GenQSGD}, we can show that $C\left(\mathbf K,B,\frac{S}{K_0}\mathbf1,\mathbf W,\tilde{\mathbf s},\mathbf s\right)\leq C\left(\mathbf K,B,\hat{\mathbf\Gamma},\mathbf W,\tilde{\mathbf s},\mathbf s\right)$.
Therefore, we can show Lemma~\ref{Lem:optimal_step_size}.}
\end{IEEEproof}
Lemma~\ref{Lem:optimal_step_size} indicates that the constant step size rule achieves the minimum convergence error with finite \clrtmp{global} iterations.
\clrtmp{Therefore, analogous to \cite{GenQSGD}}, based on Lemma~\ref{Lem:optimal_step_size}, we replace $\gamma^{(k_0)}$ in \clrtmp{$C\left(\mathbf K,B,\mathbf\Gamma,\mathbf W,\tilde{\mathbf s},\mathbf s\right)$} and \eqref{eq:Cons_gamma} with $\gamma$ for all $k_0\in\mathcal K_0$ and optimize $\gamma$ instead.
\clrtmp{Then, to address Challenge (ii), we introduce an auxiliary variable $T_1>0$ and impose the following constraint:
\begin{align}\label{eq:T1}
\frac{C_n}{F_n}K_nT_1^{-1}\leq1,\ n\in\mathcal N.
\end{align}
It remains to address Challenges (iii) and (iv), which do not exist for \cite[Problem~9]{GenQSGD}.}
Specifically, \clrtmp{to address Challenge (iii), we introduce auxiliary variables $T_2>0$ and $\mathbf S\succ\mathbf 0$ and impose the following constraints}:
\begin{align}
&\mathbf W^T\mathbf1\leq1,\label{eq:Eq_W1}\\
&\frac{1}{\mathbf W^T\mathbf1}\leq1,\label{eq:Eq_W2}\\
&\log_2\left(\tilde s_n+1\right)+D\log_2\left(s_n+1\right)+D\leq S_n,\ n\in\setNbar,\label{eq:Eq_Sn}\\
&\frac{1}{r_n}S_nT_2^{-1}\leq1,\ n\in\mathcal N;\label{eq:T2}
\end{align}
to address Challenge (iv), we introduce an auxiliary variable $\mathbf y\succ\mathbf 0$, treat $\mathbf q_{\mathbf s}\triangleq\left(q_{s_n}\right)_{n\in\setNbar}\in\mathbb R_{++}^{N+1}$ as optimization variables, and impose the following constraints:
\begin{align}
&\frac{q_{s_n}s_n^2}{D}\leq1,\
\frac{q_{s_n}s_n}{\sqrt{D}}\leq1,\ n\in\setNbar,\label{eq:q2}\\
&\frac{D}{s_n^2\left(q_{s_n}+y_n\right)}\leq1,\
\frac{\sqrt{D}}{s_n\left(q_{s_n}+y_n\right)}\leq1,\ n\in\setNbar,\label{eq:q4}\\
&\frac{\sqrt{D}s_n}{y_ns_n^2+D}\leq1,\
\frac{D}{y_ns_n^2+\sqrt{D}s_n}\leq1,\ n\in\setNbar;\label{eq:q6}
\end{align}
Consequently, we can equivalently convert Problem~\ref{Prob:Optimization} into the following problem.
\begin{Prob}[Equivalent Problem of Problem~\ref{Prob:Optimization}]\label{Prob:EqProb}
\begin{align}
&\min_{\substack{\mathbf K,\mathbf W,\tilde{\mathbf s},\mathbf s,\mathbf S,\mathbf q_{\mathbf s},\mathbf y\succ\mathbf 0,\\B,\gamma,T_1,T_2>0}}
\frac{2\left(f\left(\hat{\mathbf x}^{(1)}\right)-f^*\right)}{\gamma K_0\sum_{n\in\mathcal N}W_nK_n}\nonumber\\
&+\frac{L^2\sigma^2\gamma^2\!\!\sum\limits_{n\in\mathcal N}\!\!W_nK_n(K_n\!+\!1)}{2B\sum_{n\in\mathcal N}W_nK_n}
+L\gamma q_{\tilde s_0,s_0}\Delta_0^2\!\sum_{n\in\mathcal N}\!W_nK_n\nonumber\\
&+\frac{L\sigma^2\gamma\left(1+q_{s_0}\right)\sum\limits_{n\in\mathcal N}\left(N+q_{s_n}\right)W_n^2 K_n}{B\sum_{n\in\mathcal N}W_nK_n}\nonumber\\
&+\frac{L\left(q_{s_0}+1\right)\sum_{n\in\mathcal N}q_{\tilde s_n,s_n}W_n^2K_n^2\Delta_n^2}{\sum_{n\in\mathcal N}W_nK_n}
+\sum_{n\in\setNbar}y_n\nonumber\\
&\mathrm{s.t.}{\quad}\eqref{eq:T1},\ \eqref{eq:Eq_W1},\ \eqref{eq:Eq_W2},\ \eqref{eq:Eq_Sn},\ \eqref{eq:T2},\ \eqref{eq:q2},\ \eqref{eq:q4},\ \eqref{eq:q6},\nonumber\\
&L^2\gamma^2K_n
\!+\!L\gamma\left(1+q_{s_0}\right)\left(N+q_{s_n}\right)W_nK_n\!\leq\!1,n\in\mathcal N,\label{eq:Eq_gamma}\\
&\frac{K_0}{T_{\max}}\left(BT_1+\frac{C_0}{F_0}+T_2+\frac{S_0}{r_0}\right)\leq1,\label{eq:Eq_T}\\
&\frac{K_0}{E_{\max}}\!\left(\!\!B\!\!\sum_{n\in\mathcal N}\!\!\alpha_nC_nF_n^2K_n
\!+\!\alpha_0C_0F_0^2\!+\!\sum_{n\in\setNbar}\frac{p_nS_n}{r_n}\!\!\right)\!\leq\!1.\label{eq:Eq_E}
\end{align}
\end{Prob}
The equivalence between Problem~\ref{Prob:Optimization} and Problem~\ref{Prob:EqProb} is established in the following theorem.
\begin{Thm}[Equivalence between Problem~\ref{Prob:Optimization} and Problem~\ref{Prob:EqProb}]\label{Thm:Equivalence}
If $(\mathbf K^*,B^*,\gamma^*,\mathbf W^*,\tilde{\mathbf s}^*,\mathbf s^*,\mathbf S^*,\mathbf q_{\mathbf s}^*,\\T_1^*,T_2^*,\mathbf y^*)$ is an optimal point of Problem~\ref{Prob:EqProb} and $D\neq s_n^2$, $n\in\setNbar$,\footnote{\clrtmp{The condition $D\neq s_n^2$, $n\in\setNbar$ holds in most practical cases.}}
then $\left(\mathbf K^*,B^*,\gamma^*\mathbf1,\mathbf W^*,\tilde{\mathbf s}^*,\mathbf s^*\right)$ is an optimal point of Problem~\ref{Prob:Optimization}.
\end{Thm}
\begin{IEEEproof}
See Appendix~C.
\end{IEEEproof}
By Theorem~\ref{Thm:Equivalence}, we can solve Problem~\ref{Prob:EqProb} instead of Problem~\ref{Prob:Optimization}.
Note that Problem~\ref{Prob:EqProb} has the following structural properties:
the constraint functions in \eqref{eq:Eq_W1}, \eqref{eq:Eq_gamma}, \eqref{eq:Eq_T}, and \eqref{eq:Eq_E} are posynomials;
the constraint functions in \eqref{eq:Eq_W2}, \eqref{eq:q4}, and \eqref{eq:q6} are ratios between monomials and posynomials;
the constraint functions in \eqref{eq:T1}--\eqref{eq:q2} are monomials;
the constraint function in \eqref{eq:Eq_Sn} is concave and involves logarithmic functions;
and the objective involves ratios between posynomials.
Thus, Problem~\ref{Prob:EqProb} is more complicated than a CGP.\footnote{\clrtmp{CGP is to minimize the ratio between two posynomials subject to upper bound inequality constraints on ratios between posynomials (and equality constraints on monomials) \cite{CGP}.}}

In the following, using GIA \cite{GIA} and tricks for solving CGP \cite{CGP}, we propose an iterative algorithm to obtain a KKT point of Problem~\ref{Prob:EqProb}.
The idea is to construct and solve a sequence of successively refined approximate geometric programs (GPs).
Specifically, at iteration $t$, update \clrtmp{$\left(\mathbf K^{(t)},B^{(t)},\gamma^{(t)},\mathbf W^{(t)},\tilde{\mathbf s}^{(t)},\mathbf s^{(t)},\mathbf q_{\mathbf s}^{(t)},\mathbf y^{(t)}\right)$} by solving Problem~\ref{Prob:ApproximateGP},
which is parameterized by \clrtmp{$(\mathbf K^{(t-1)},\mathbf W^{(t-1)},\tilde{\mathbf s}^{(t-1)},\mathbf s^{(t-1)},\mathbf q_{\mathbf s}^{(t-1)},\mathbf y^{(t-1)})$} obtained at iteration $t-1$.
\begin{Prob}[Approximate GP of Problem~\ref{Prob:EqProb} at Iteration $t$]\label{Prob:ApproximateGP}
\begin{align}
&\min_{\substack{\mathbf K,\mathbf W,\tilde{\mathbf s},\mathbf s,\mathbf S,\mathbf q_{\mathbf s},\mathbf y\succ\mathbf 0,\\B,\gamma,T_1,T_2>0}}
\hat{C}\left(\mathbf K,B,\gamma
,\mathbf W,\tilde{\mathbf s},\mathbf q_{\mathbf s};\mathbf K^{(t-1)},\mathbf W^{(t-1)}\right)\nonumber\\
&\hspace{180pt}+\sum_{n\in\setNbar}y_n\nonumber\\
&\mathrm{s.t.}{\quad}\eqref{eq:T1},\ \eqref{eq:Eq_W1},\ \eqref{eq:T2},\ \eqref{eq:q2},\ \eqref{eq:Eq_gamma},\ \eqref{eq:Eq_T},\ \eqref{eq:Eq_E},\nonumber\\
&\frac{1}{\sum_{n\in\mathcal N}W_n^{(t-1)}\prod_{n\in\mathcal N}\left(\frac{W_n}{W_n^{(t-1)}}\right)^{\frac{W_n^{(t-1)}}{\sum_{n\in\mathcal N}W_n^{(t-1)}}}}\leq1,\label{eq:Approx_W2}\\
&\log_2\left(\tilde s_n^{(t-1)}+1\right)
+\frac{\tilde s_n-\tilde s_n^{(t-1)}}{\left(\tilde s_n^{(t-1)}+1\right)\ln2}+D\log_2\left(s_n^{(t-1)}+1\right)\nonumber\\
&+\frac{D\left(s_n-s_n^{(t-1)}\right)}{\left(s_n^{(t-1)}+1\right)\ln2}+D
\leq S_n,\ n\in\setNbar,\label{eq:Approx_Sn}\\
&\frac{D}{\left(\!q_{s_n}^{(t-1)}\!\!+\!y_n^{(t-1)}\!\right)\!s_n^2\!\left(\!\frac{q_{s_n}}{q_{s_n}^{(t-1)}}\!\right)^{\!\!\frac{q_{s_n}^{(t-1)}}{q_{s_n}^{(t-1)}\!+y_n^{(t-1)}}}\!\!\!\!\left(\!\frac{y_n}{y_n^{(t-1)}}\!\right)^{\!\!\frac{y_n^{(t-1)}}{q_{s_n}^{(t-1)}\!+y_n^{(t-1)}}}}\!\leq\!1,\nonumber\\
&\frac{\sqrt{D}}{\left(\!q_{s_n}^{(t-1)}\!+\!y_n^{(t-1)}\!\right)\!s_n\!\left(\!\!\frac{q_{s_n}}{q_{s_n}^{(t-1)}}\!\!\right)^{\!\!\frac{q_{s_n}^{(t-1)}}{q_{s_n}^{(t-1)}\!+\!y_n^{(t-1)}}}\!\!\left(\frac{y_n}{y_n^{(t-1)}}\!\right)^{\!\!\frac{y_n^{(t-1)}}{q_{s_n}^{(t-1)}\!+\!y_n^{(t-1)}}}}\!\leq\!1,\nonumber\\
&\hspace{180pt}\forall n\in\setNbar,\label{eq:Approx_q4}\\
&\frac{\sqrt{D}s_n}{\left(\!y_n^{(t-1)}\!\left(\!s_n^{(t-1)}\!\right)^2\!+D\!\right)\!\left(\!\frac{y_ns_n^2}{y_n^{(t-1)}\!\left(s_n^{(t-1)}\right)^2}\!\right)^{\!\frac{y_n^{(t-1)}\left(s_n^{(t-1)}\right)^2}{y_n^{(t-1)}\!\left(s_n^{(t-1)}\right)^2\!+\!D}}}\leq1,\nonumber\\
&{\quad}\frac{D}{\left(\!y_n^{(t-1)}s_n^{(t-1)}\!+\!\sqrt{D}\right)\!s_n\left(\frac{y_ns_n}{y_n^{(t-1)}s_n^{(t-1)}}\right)^{\frac{y_n^{(t-1)}s_n^{(t-1)}}{y_n^{(t-1)}s_n^{(t-1)}+\sqrt{D}}}}\leq1,\nonumber\\
&\hspace{180pt}\forall n\in\setNbar,\label{eq:Approx_q6}
\end{align}
where $\hat{C}\left(\mathbf K,B,\gamma,\mathbf W,\tilde{\mathbf s},\mathbf q_{\mathbf s};\mathbf K^{(t-1)},\mathbf W^{(t-1)}\right)$ is shown at the top of the next page,
and $(\mathbf K^{(t)}\!,B^{(t)}\!,\gamma^{(t)}\!,\mathbf W^{(t)}\!,\tilde{\mathbf s}^{(t)}\!,\mathbf s^{(t)}\!,\mathbf S^{(t)}\!,\mathbf q_{\mathbf s}^{(t)}\!,T_1^{(t)}\!,T_2^{(t)}\!,\mathbf y^{(t)})$ denotes an optimal solution of Problem~\ref{Prob:ApproximateGP}.
\end{Prob}

\begin{figure*}
\begin{align*}
&\hat{C}\left(\mathbf K,B,\gamma,\mathbf W,\tilde{\mathbf s},\mathbf q_{\mathbf s};\mathbf K^{(t-1)},\mathbf W^{(t-1)}\right)
=\frac{2\left(f\left(\hat{\mathbf x}^{(1)}\right)-f^*\right)}{\gamma K_0\!\sum\limits_{n\in\mathcal{N}}\!W_n^{(t-1)}K_n^{(t-1)}\prod\limits_{n\in\mathcal N}\left(\frac{W_nK_n}{W_n^{(t-1)}K_n^{(t-1)}}\right)^{\frac{W_n^{(t-1)}K_n^{(t-1)}}{\sum\limits_{n\in\mathcal N}W_n^{(t-1)}K_n^{(t-1)}}}}\nonumber\\
&+\frac{L^2\sigma^2\gamma^2\!\!\sum\limits_{n\in\mathcal N}\!\!W_nK_n(K_n+1)\!+\!2L\sigma^2\gamma\left(1+q_{s_0}\right)
\!\!\sum\limits_{n\in\mathcal{N}}\!\!\left(N+q_{s_n}\right)W_n^2K_n}{2B\sum\limits_{n\in\mathcal N}W_n^{(t-1)}K_n^{(t-1)}\prod\limits_{n\in\mathcal N}\left(\frac{W_nK_n}{W_n^{(t-1)}K_n^{(t-1)}}\right)^{\frac{W_n^{(t-1)}K_n^{(t-1)}}{\sum\limits_{n\in\mathcal N}W_n^{(t-1)}K_n^{(t-1)}}}}
+\frac{L\gamma\left(1+q_{s_0}\right)\Delta_0^2\!\sum\limits_{n\in\mathcal N}\!W_nK_n}{4\tilde s_0^2}\nonumber\\
&+\frac{L\gamma\left(1+q_{s_0}\right)}{4\sum\limits_{n\in\mathcal N}W_n^{(t-1)}K_n^{(t-1)}\prod\limits_{n\in\mathcal N}\left(\frac{W_nK_n}{W_n^{(t-1)}K_n^{(t-1)}}\right)^{\frac{W_n^{(t-1)}K_n^{(t-1)}}{\sum\limits_{n\in\mathcal N}W_n^{(t-1)}K_n^{(t-1)}}}}\sum\limits_{n\in\mathcal N}\frac{\left(1+q_{s_n}\right)W_n^2K_n^2\Delta_n^2}{\tilde s_n^2}
\end{align*}
\normalsize \hrulefill
\end{figure*}

\begin{algorithm}[t]
\caption{Algorithm for Obtaining a KKT Point of Problem~\ref{Prob:EqProb}}
\label{Alg:SolveEqProb}
\begin{algorithmic}[1]
\small{
\STATE {\bf{Initialize:}}
Choose any feasible solution $(\mathbf K^{(0)},B^{(0)},\gamma^{(0)},\mathbf W^{(0)},\tilde{\mathbf s}^{(0)},\mathbf s^{(0)},\mathbf S^{(0)},\mathbf q_{\mathbf s}^{(0)},T_1^{(0)},T_2^{(0)},\mathbf y^{(0)})$ of Problem~\ref{Prob:EqProb},
and set $t=1$.
\REPEAT
    \STATE
    Compute $(\mathbf K^{(t)}\!,B^{(t)}\!,\gamma^{(t)}\!,\mathbf W^{(t)}\!,\tilde{\mathbf s}^{(t)}\!,\mathbf s^{(t)}\!,\mathbf S^{(t)}\!,\mathbf q_{\mathbf s}^{(t)}\!,T_1^{(t)}\!,T_2^{(t)}\!,$ $\mathbf y^{(t)})$ by transforming Problem~\ref{Prob:EqProb} into a GP in convex form
    and solving it with standard convex optimization techniques.
    \STATE Set $t:=t+1$.
\UNTIL{Some convergence criteria is met.}
}\normalsize
\end{algorithmic}
\end{algorithm}

\clrtmp{The objective of Problem~\ref{Prob:ApproximateGP} and the constraint functions in \eqref{eq:Approx_W2}, \eqref{eq:Approx_q4}, and \eqref{eq:Approx_q6}, which are approximations of the objective of Problem~\ref{Prob:EqProb} and the constraint functions in \eqref{eq:Eq_W2}, \eqref{eq:q4}, and \eqref{eq:q6}, respectively, are constructed by adopting a commonly used trick in CGP \cite[Lemma~1]{CGP} that is based on the arithmetic-geometric mean inequality.
The constraint functions in \eqref{eq:Approx_Sn}, which are approximations of the constraint functions in \eqref{eq:Eq_Sn}, are constructed by linearizing the logarithmic functions in \eqref{eq:Eq_Sn} at $\left(\tilde{\mathbf s}_0,\mathbf s_0\right)=\left(\tilde{\mathbf s}_0^{(t-1)},\mathbf s_0^{(t-1)}\right)$.
The constraint functions in \eqref{eq:Approx_W2}--\eqref{eq:Approx_q6} are posynomials.}
As a result, Problem~\ref{Prob:ApproximateGP} is a standard GP and can be readily transformed into a convex problem and solved by using traditional convex optimization techniques such as interior-point methods. In particular, if an interior-point method is applied, the computational complexity for solving Problem~\ref{Prob:ApproximateGP} is $\mathcal O(N^{3.5})$ \cite{CVX}.
The details are summarized in Algorithm~\ref{Alg:SolveEqProb}.
Following \cite[Proposition 3]{CGP}, we have the following result.
\clrtmp{\begin{Thm}[Convergence of Algorithm~\ref{Alg:SolveEqProb}]\label{Thm:SolveEqProb_convergence}
$(\mathbf K^{(t)},B^{(t)},\gamma^{(t)},\mathbf W^{(t)},\tilde{\mathbf s}^{(t)},\mathbf s^{(t)},\mathbf S^{(t)},\mathbf q_{\mathbf s}^{(t)},T_1^{(t)},T_2^{(t)},\mathbf y^{(t)})$ obtained by Algorithm~\ref{Alg:SolveEqProb} converges to a KKT point of Problem~\ref{Prob:EqProb}, as $t\rightarrow\infty$.
\end{Thm}}
\begin{IEEEproof}
See Appendix~D.
\end{IEEEproof}

\section{Numerical and Experimental Results}\label{Sec:Results}
In this section, we  evaluate the convergence error of the proposed GQFedWAvg (GQFWA).\footnote{Source code for the experiments is available at \cite{GitHub}.}
We consider edge computing systems with $N$ workers and equally divide the set of workers, i.e., $\mathcal N$, into two classes, $\mathcal N_1$ and $\mathcal N_2$.
We set $F_n=F^{(1)}\in\mathbb Z_{++}$, \clrtmp{$r_n=r^{(1)}\in\mathbb R_{++}$}, $n\in\mathcal N_1$ and $F_n=F^{(2)}\in\mathbb Z_{++}$, \clrtmp{$r_n=r^{(2)}\in\mathbb R_{++}$}, $n\in\mathcal N_2$
such that $\frac{F^{(1)}+F^{(2)}}{2}=1\times10^9$ (cycles/s) and \clrtmp{$\frac{r^{(1)}+r^{(2)}}{2}=2.8\times10^6$ (b/s)}, respectively.
We use $\frac{F^{(1)}}{F^{(2)}}$ and \clrtmp{$\frac{r^{(1)}}{r^{(2)}}$} to characterize the differences in computing \clrtmp{and communication} capabilities between the two classes of workers.
\clrtmp{We consider three systems, including a homogeneous system (Homo) with $\frac{F^{(1)}}{F^{(2)}}=1$ and $\frac{r^{(1)}}{r^{(2)}}=1$,
a communication-heterogeneous system (CommH) with $\frac{F^{(1)}}{F^{(2)}}=1$ and $\frac{r^{(1)}}{r^{(2)}}=2.5$,
and a computing-heterogeneous system (CompH) with $\frac{F^{(1)}}{F^{(2)}}=10$ and $\frac{r^{(1)}}{r^{(2)}}=1$, respectively}.
Besides, we choose
$\alpha_n=2\times10^{-28},n\in\setNbar$,
$F_0=3\times\!10^9$ (cycles/s),
$C_0=100$ (cycles),
$p_0=20$ (W),
$r_0=7.5\times10^7$ (b/s),
$C_n=1\times10^6$ (cycles), $n\in\mathcal N$,
and $p_n=1.5$ (W), $n\in\mathcal N$.

\subsection{Numerical Results}\label{SubSec:NumericalResults}

For the numerical simulations, we consider the following baseline FL algorithms:
PR-SGD with $B=1$, $W_n=\frac{1}{N}\mathbf1$, $n\in\mathcal N$, $\tilde{s}_n=s_n=2^{32}$, $n\in\setNbar$ \cite{YuHao} (PR);
FedHQ with $W_n=\frac{\frac{1}{1+q_{s_n}}}{\sum_{n'\in\mathcal{N}}\frac{1}{1+q_{s_{n'}}}}$, $n\in\mathcal{N}$, $\tilde{s}_n=2^{8}$, $n\in\setNbar$ \cite{CongShen} (FHQ);
GenQSGD with $W_n=\frac{1}{N}\mathbf1$, $n\in\mathcal N$, $\tilde{s}_n=2^{8}$, $n\in\setNbar$ \cite{GenQSGD} (GQ);
GQFedWAvg with $K_1=\cdots=K_N=K$ for some $K\in\mathbb{Z}_{++}$ (sameK);
GQFedWAvg with $W_1=\cdots=W_N=1/N$ (sameW);
GQFedWAvg with $s_1=\cdots=s_N=s$ for some $s\in\mathbb{Z}_{++}$ (sameS);
GQFedWAvg with $\tilde{s}_1=\cdots=\tilde{s}_N=\tilde{s}$ for some $\tilde{s}\in\mathbb{Z}_{++}$ (sameTS);
GQFedWAvg with high-precision-quantization at the server (i.e., $\tilde{s}_0=s_0=2^{32}$) (HS);
and GQFedWAvg with accurate model exchange (without quantization) (AC).
The last six baseline algorithms are degenerated versions of the proposed GQFedWAvg and are used to show the advantage of flexible choices of GQFedWAvg's parameters.
All the remaining algorithm parameters of each baseline algorithm are optimized using the proposed method.

\begin{figure}[t]
\begin{center}
\subfigure[\small{Weights.}]
{\hspace{0pt}{\includegraphics[width=120pt]{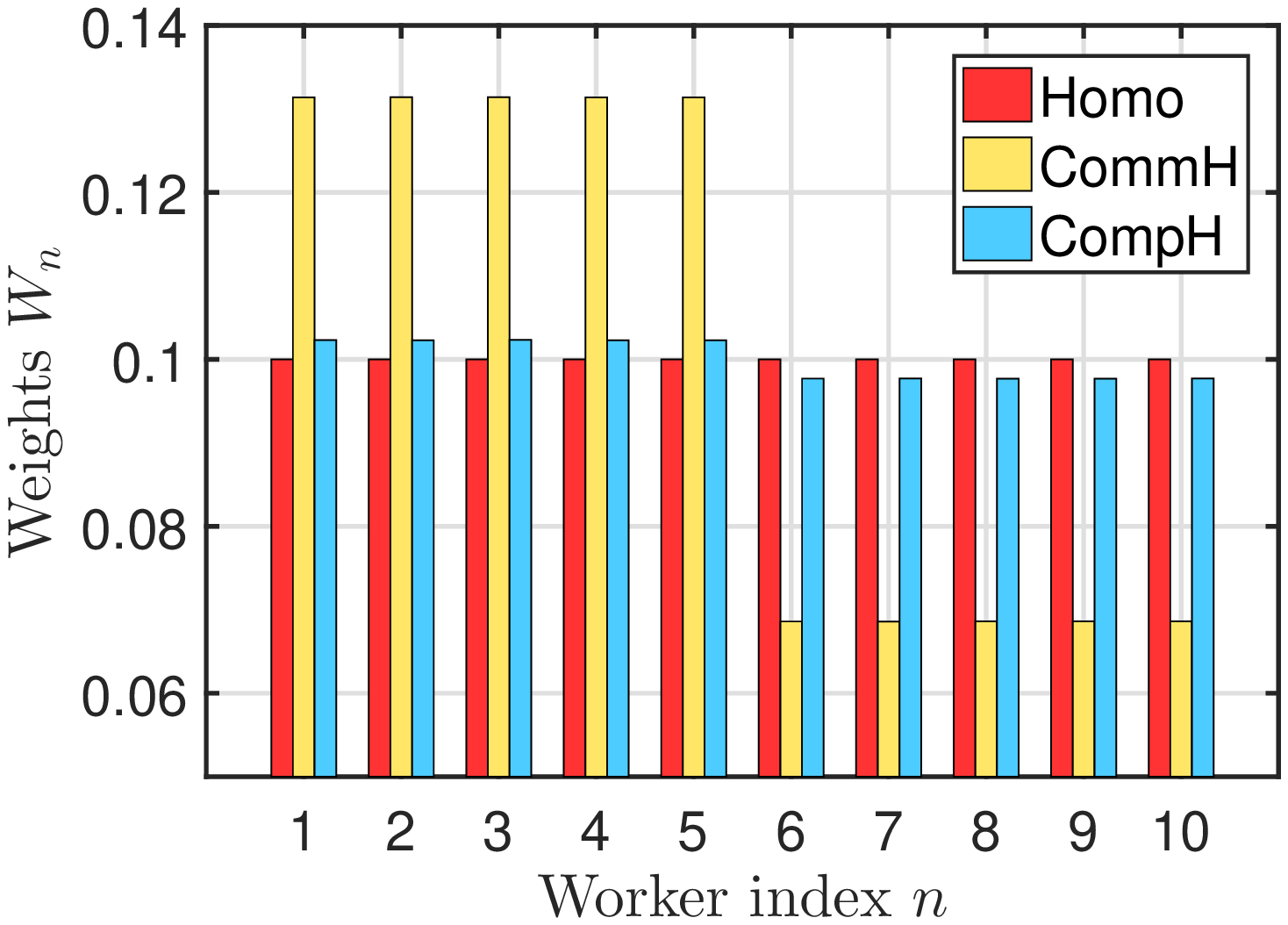}}\hspace{0pt}\label{Fig:Wn}}
\subfigure[\small{Numbers of local iterations.}]
{\hspace{0pt}{\includegraphics[width=120pt]{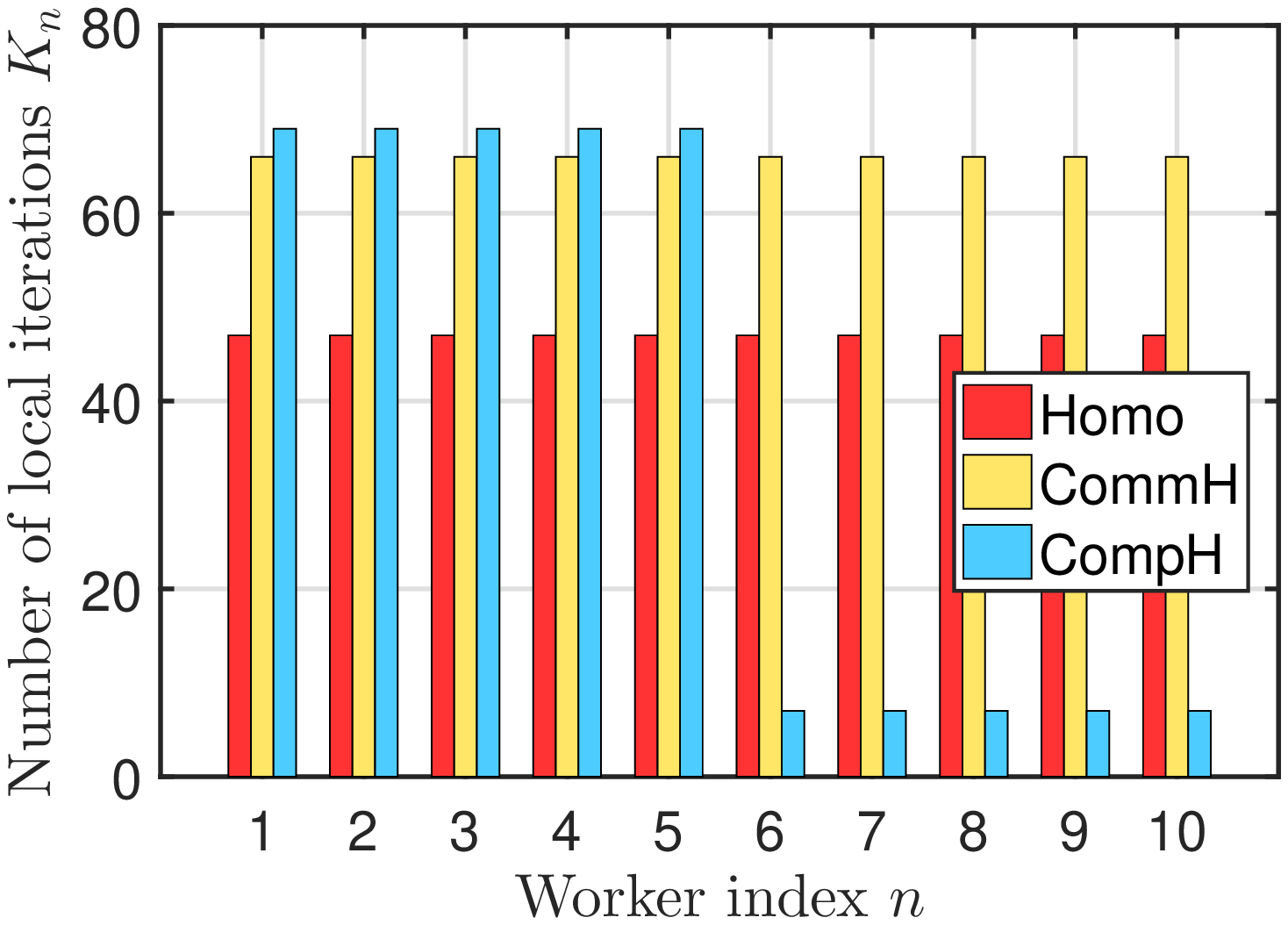}}\hspace{0pt}\label{Fig:Kn}}
\subfigure[\small{Numbers of quantization levels for norms of local model update directions.}]
{\hspace{0pt}{\includegraphics[width=120pt]{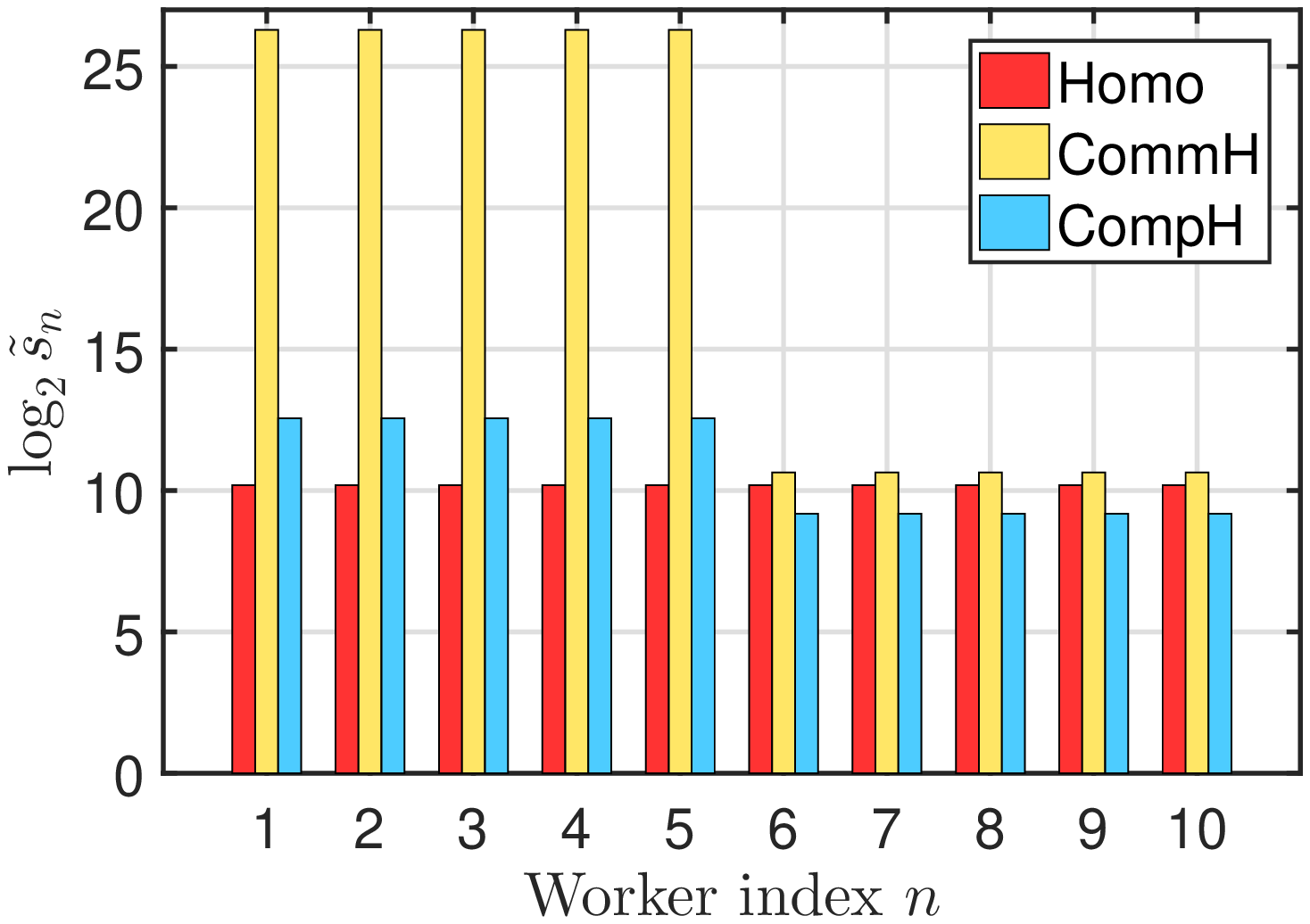}}\hspace{0pt}\label{Fig:tsn}}
\subfigure[\small{Numbers of quantization levels for the normalized local model update directions.}]
{\hspace{0pt}{\includegraphics[width=120pt]{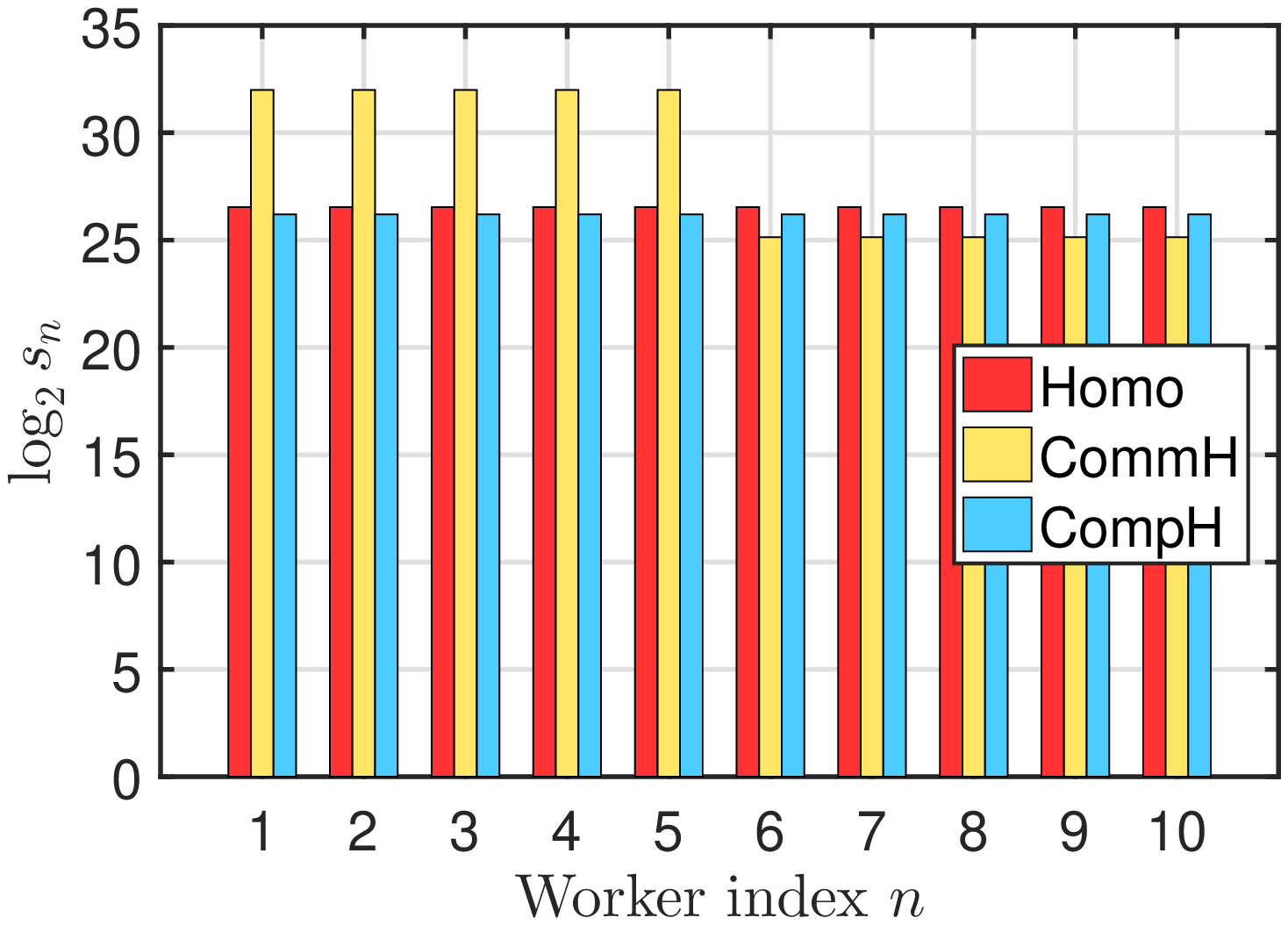}}\hspace{0pt}\label{Fig:sn}}
\end{center}
\caption{\small{Optimized algorithm parameters of GQFedWAvg in the homogeneous, communication-heterogeneous, and computing-heterogeneous systems.}}
\label{Fig:OptParamters}
\end{figure}

Fig.~\ref{Fig:OptParamters} demonstrates the optimized algorithm parameters of GQFedWAvg, including the weights $W_n$, $n\in\mathcal{N}$, numbers of local iterations $K_n$, $n\in\mathcal{N}$, and quantization parameters $\tilde{s}_n$, $s_n$, $n\in\mathcal{N}$, in the homogeneous, communication-heterogeneous, and computing-heterogeneous systems.
From Fig.~\ref{Fig:OptParamters}, we have the following observations.
In the homogeneous system, $W_n$ ($K_n$, $\tilde{s}_n$, and $s_n$, respectively), $n\in\mathcal{N}$ are the same, \clrtmp{indicating that all workers' computing and communication resources are uniformly utilized and all workers' local model updates are considered equally important to improve the convergence of GQFedWAvg}.
In the computing-heterogeneous (communication-heterogeneous) system, $K_n$ ($\tilde{s}_n$ and $s_n$) and $W_n$ for the workers with higher CPU frequencies $F_n$ (transmission rates $r_n$) are larger, indicating that for workers with stronger computing (communication) capabilities, more computing (communication) resources are utilized and the corresponding local model updates are given to more consideration to improve the convergence of GQFedWAvg.

\begin{figure}[t]
\begin{center}
\subfigure[\small{Convergence error versus $T_{\max}$ at $E_{\max}=500$.}]
{\hspace{0pt}{\includegraphics[width=120pt]{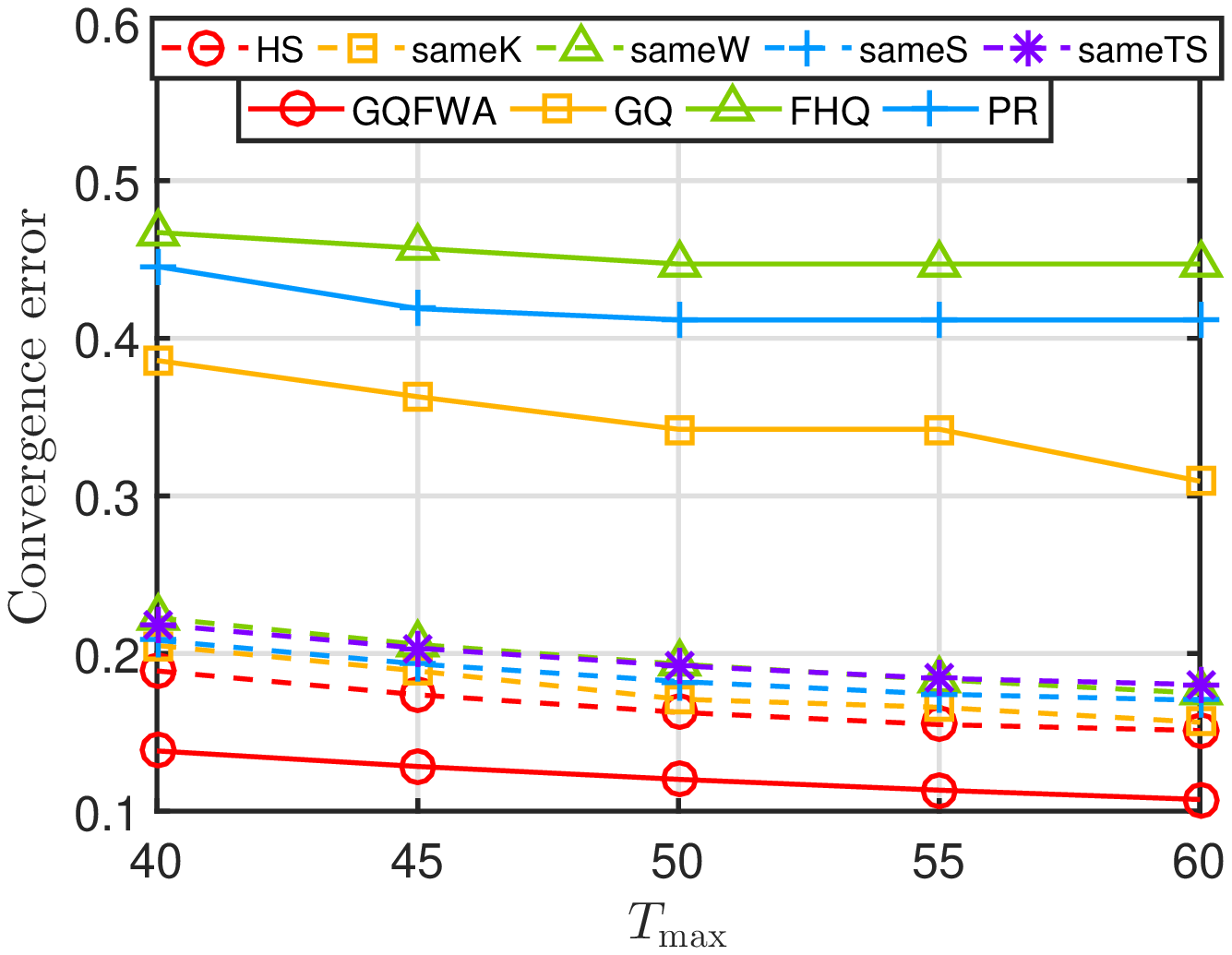}}\hspace{0pt}\label{Fig:C_Tmax_Homo}}
\subfigure[\small{Convergence error versus $E_{\max}$ at $T_{\max}=60$.}]
{\hspace{0pt}{\includegraphics[width=120pt]{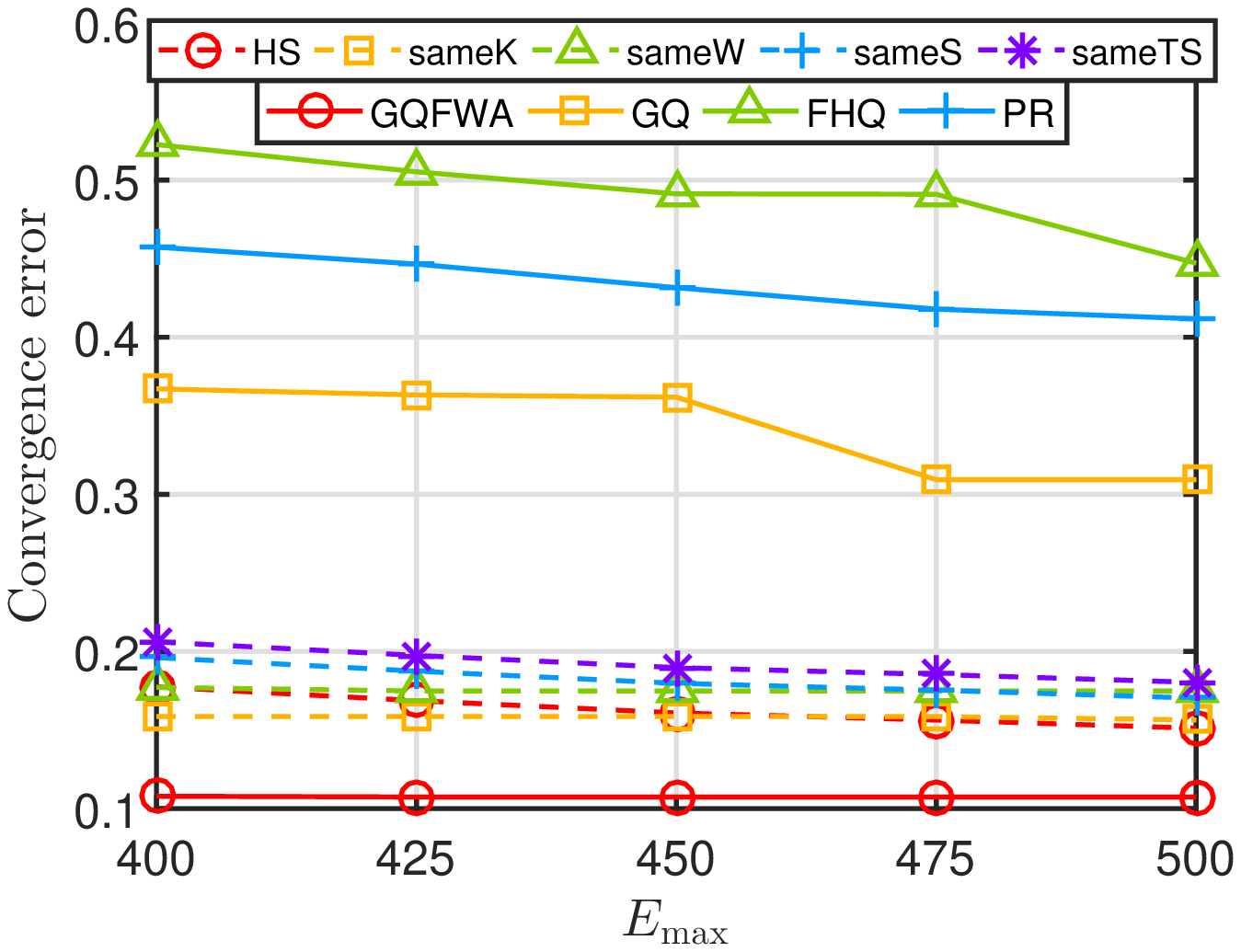}}\hspace{0pt}\label{Fig:C_Emax_Homo}}
\end{center}
\caption{\small{Convergence error for the homogeneous system.}}
\label{Fig:C_Homo}
\end{figure}
\begin{figure}[t]
\begin{center}
\subfigure[\small{Convergence error versus $T_{\max}$ at $E_{\max}=500$.}]
{\hspace{0pt}{\includegraphics[width=120pt]{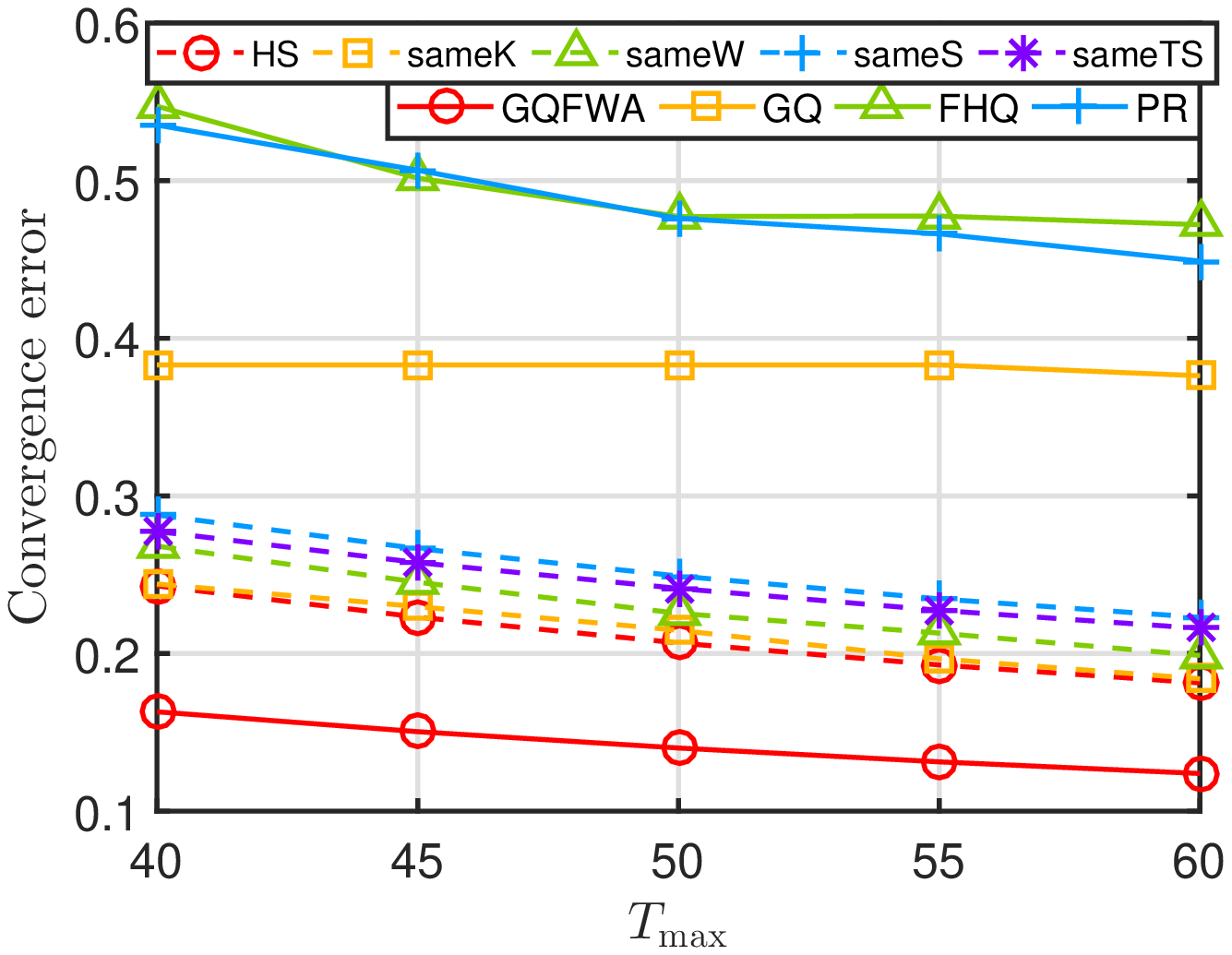}}\hspace{0pt}\label{Fig:C_Tmax_CommHetero}}
\subfigure[\small{Convergence error versus $E_{\max}$ at $T_{\max}=60$.}]
{\hspace{0pt}{\includegraphics[width=120pt]{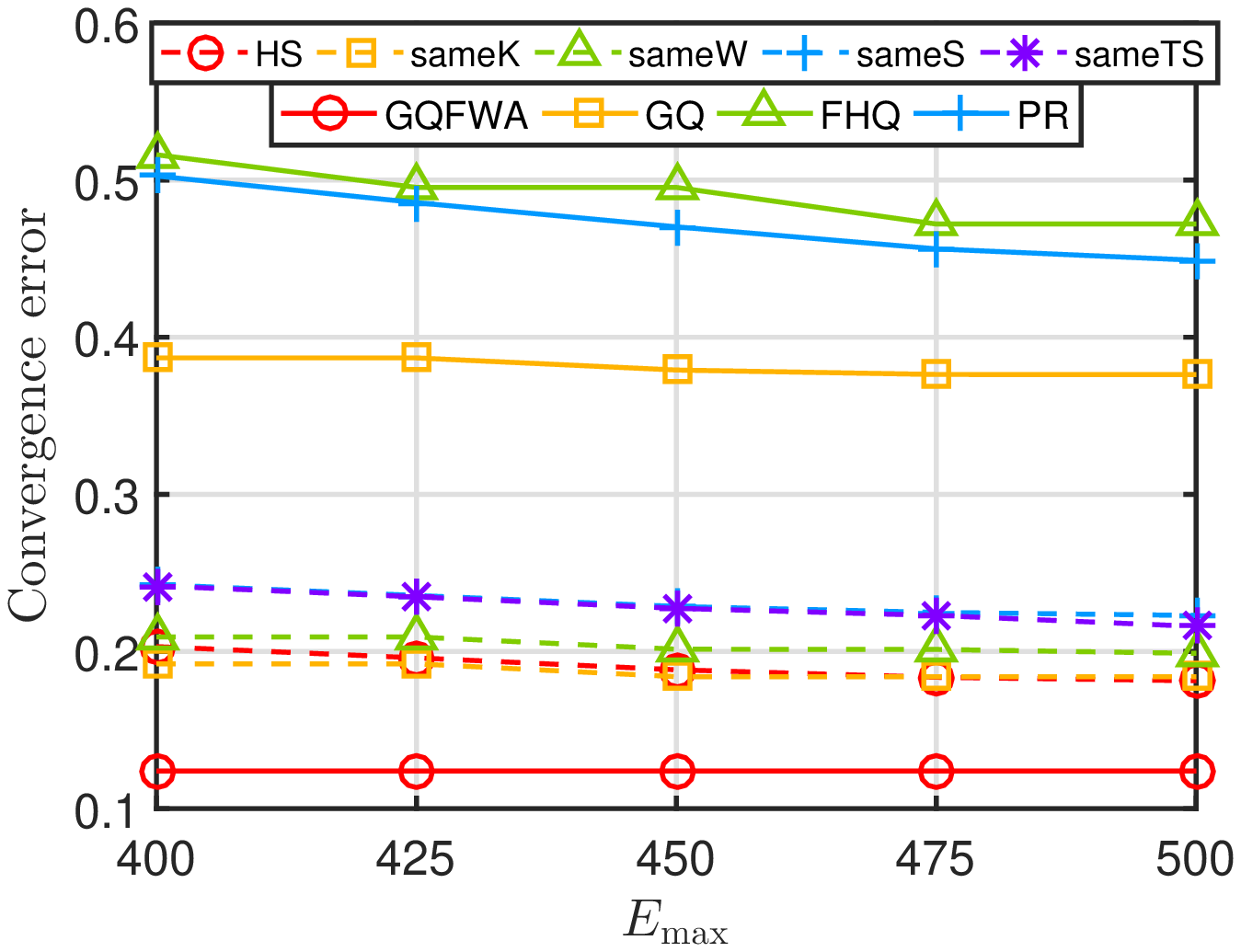}}\hspace{0pt}\label{Fig:C_Emax_CommHetero}}
\end{center}
\caption{\small{Convergence error for the communication-heterogeneous system.}}
\label{Fig:C_CommHetero}
\end{figure}
\begin{figure}[t]
\begin{center}
\subfigure[\small{Convergence error versus $T_{\max}$ at $E_{\max}=500$.}]
{\hspace{0pt}{\includegraphics[width=120pt]{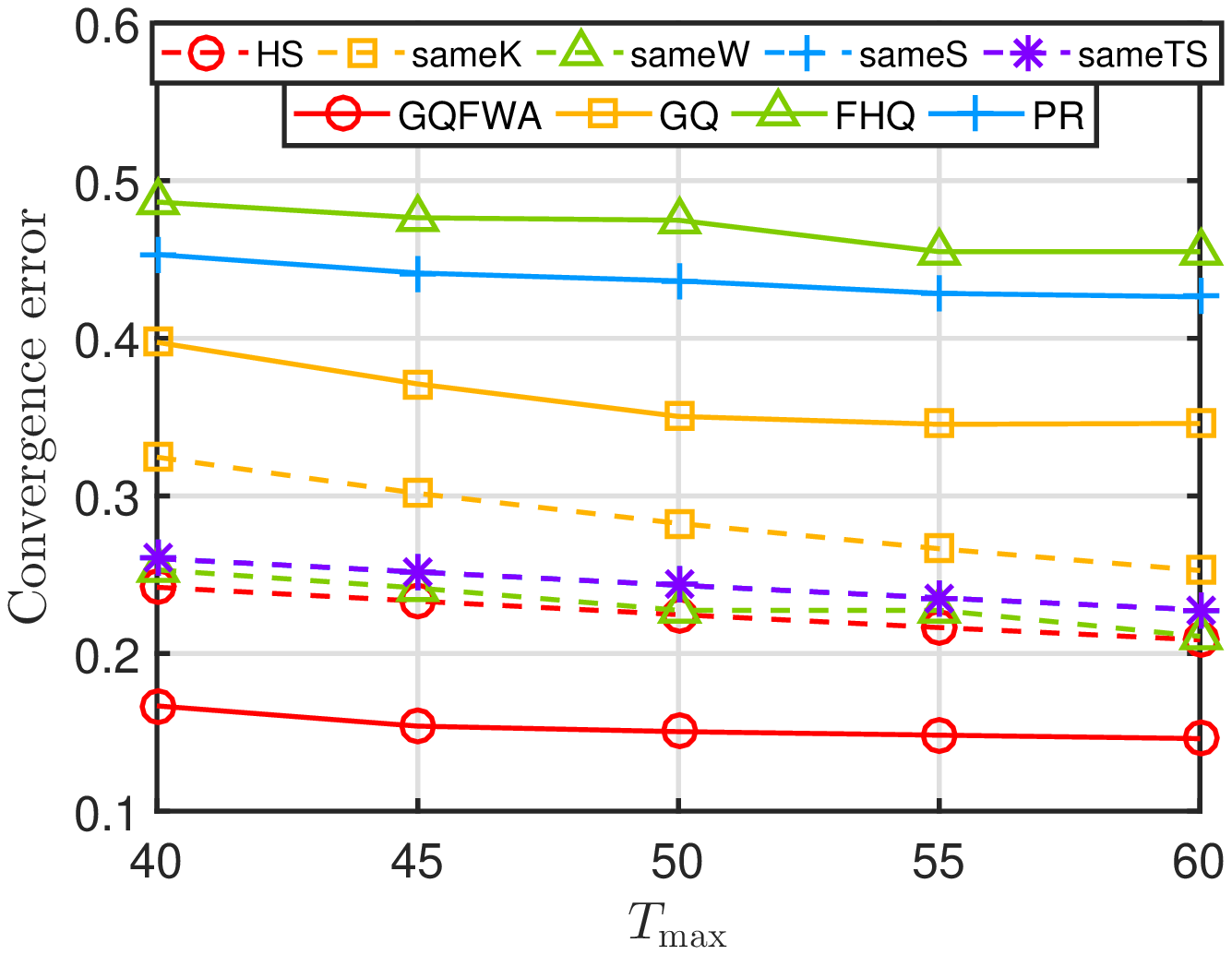}}\hspace{0pt}\label{Fig:C_Tmax_CompHetero}}
\subfigure[\small{Convergence error versus $E_{\max}$ at $T_{\max}=60$.}]
{\hspace{0pt}{\includegraphics[width=120pt]{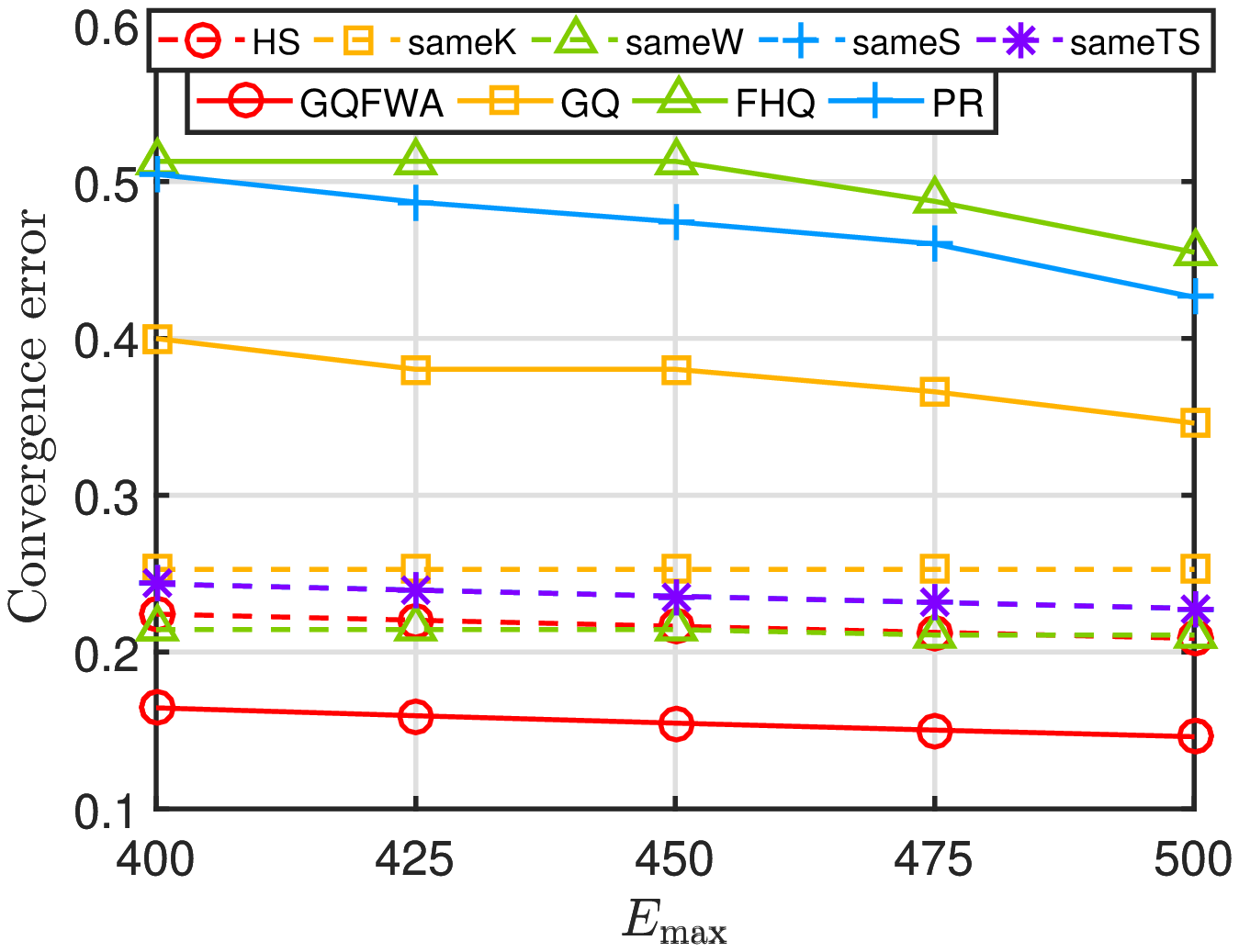}}\hspace{0pt}\label{Fig:C_Emax_CompHetero}}
\end{center}
\caption{\small{Convergence error for the computing-heterogeneous system.}}
\label{Fig:C_CompHetero}
\end{figure}

\begin{figure}[t]
\begin{minipage}[c]{120pt}
\begin{center}
{\resizebox{120pt}{!}{\includegraphics{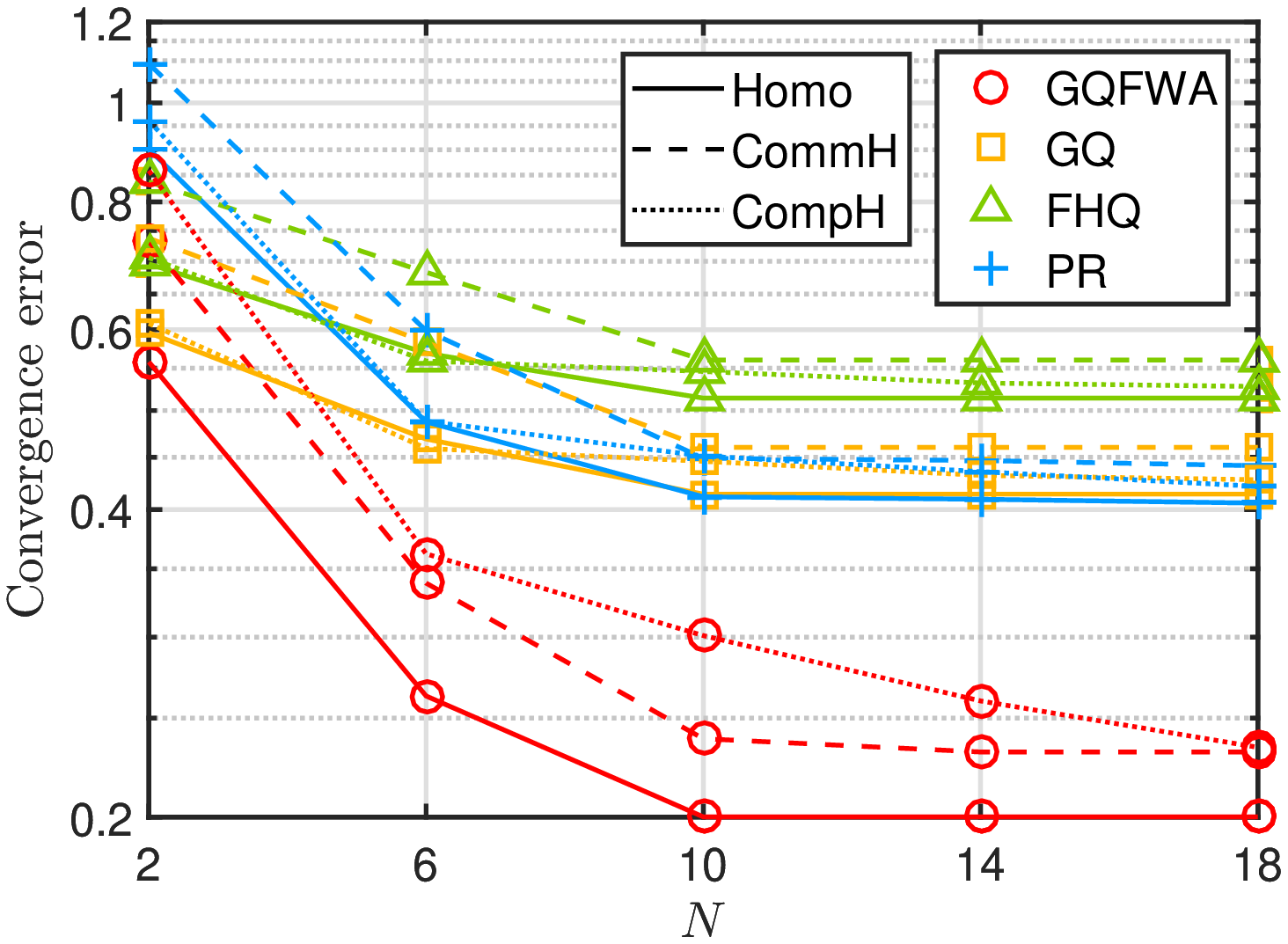}}}
\end{center}
\caption{\small{Convergence error versus $N$ at $T_{\max}=60$.}}
\label{Fig:diffN}
\end{minipage}
\begin{minipage}[c]{120pt}
\begin{center}
{\resizebox{120pt}{!}{\includegraphics{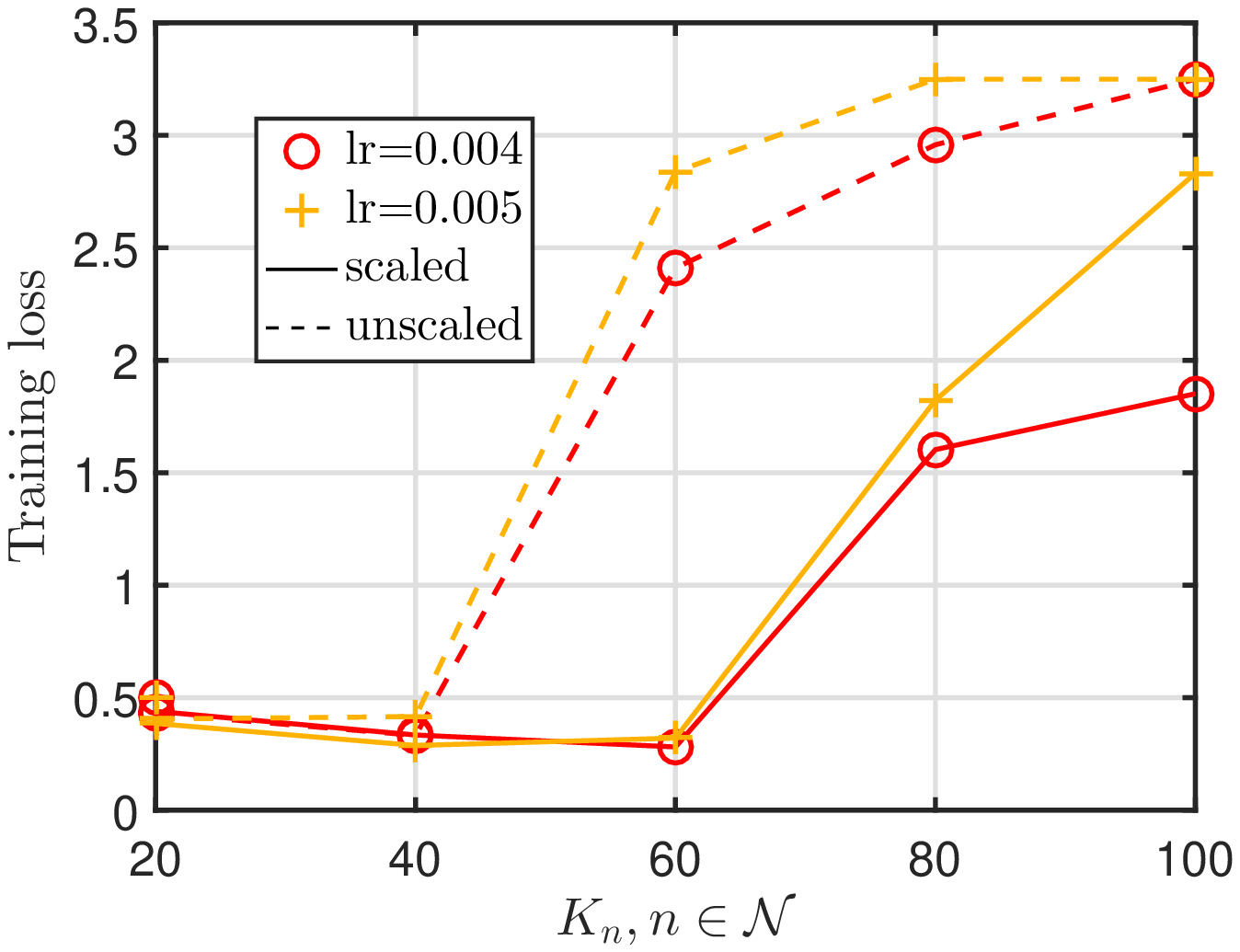}}}
\end{center}
\caption{\small{Training loss versus $K_n,n\in\mathcal{N}$ at $B=10$.}}
\label{Fig:Scale}
\end{minipage}
\end{figure}

Fig.~\ref{Fig:C_Homo}-Fig.~\ref{Fig:diffN} compare the performance of all FL algorithms for the homogeneous, communication-heterogeneous, and computing-heterogeneous systems.
Specifically, Fig.~\ref{Fig:C_Homo}-Fig.~\ref{Fig:C_CompHetero} illustrate the convergence errors of all FL algorithms versus $T_{\max}$ and $E_{\max}$,
and Fig.~\ref{Fig:diffN} illustrates the convergence errors of existing FL algorithms and the proposed GQFedWAvg versus $N$.
From Fig.~\ref{Fig:C_Homo}-Fig.~\ref{Fig:diffN}, we can draw the following conclusions.
Firstly, GQFedWAvg outperforms all baseline FL algorithms in each system, and the gain of GQFedWAvg over each baseline FL algorithm in the homogeneous system is smaller than those in the two heterogeneous systems in most cases.
The two facts indicate the importance of optimally and flexibly adapting the algorithm parameters  of all workers to the computing and communication resources at workers (rather than restricting all workers to adopt identical algorithm parameters) and the importance of properly choosing the quantization parameters of the server when the communication cost is considered.
Secondly, for each FL algorithm, the convergence error in the homogeneous system is lower than in the two heterogeneous systems, which indicates that the computing and communication resources of the homogeneous system can be more effectively utilized.
Thirdly, the convergence error of each FL algorithm decreases with $T_{\max}$ and $E_{\max}$, implying that the proposed optimization framework can be applied to all the baseline FL algorithms and can be widely used to trade-off the time cost, energy cost, and convergence error.
Finally, the convergence errors of existing FL algorithms and the proposed GQFedWAvg all decrease with $N$ (under $E_{\max}=50N$, i.e., keeping the energy limit per worker unchanged), which demonstrates the advantage of involving more workers in the collaborative training at the cost of increased overall time and energy consumptions.

\subsection{Experimental Results}\label{SubSec:ExperimentalResults}

\begin{figure}[t]
\begin{center}
\subfigure[\small{Training loss versus $T_{\max}$ at $E_{\max}=500$.}]
{\hspace{0pt}{\includegraphics[width=120pt]{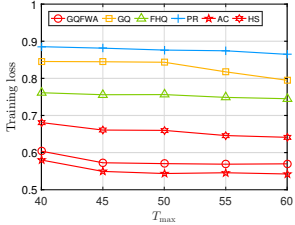}}\hspace{0pt}\label{Fig:L_Tmax_Homo_cost}}
\subfigure[\small{Training loss versus $E_{\max}$ at $T_{\max}=60$.}]
{\hspace{0pt}{\includegraphics[width=120pt]{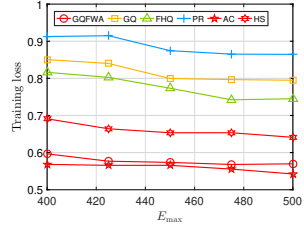}}\hspace{0pt}\label{Fig:L_Emax_Homo_cost}}
\subfigure[\small{Test accuracy versus $T_{\max}$ at $E_{\max}=500$.}]
{\hspace{0pt}{\includegraphics[width=120pt]{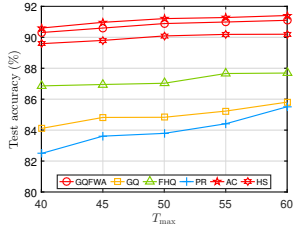}}\hspace{0pt}\label{Fig:L_Tmax_Homo_acc}}
\subfigure[\small{Test accuracy versus $E_{\max}$ at $T_{\max}=60$.}]
{\hspace{0pt}{\includegraphics[width=120pt]{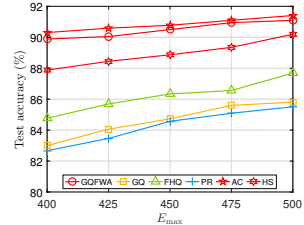}}\hspace{0pt}\label{Fig:L_Emax_Homo_acc}}
\end{center}
\caption{\small{Training loss and test accuracy for the homogeneous system.}}
\label{Fig:L_Homo}
\end{figure}
\begin{figure}[t]
\begin{center}
\subfigure[\small{Training loss versus $T_{\max}$ at $E_{\max}=500$.}]
{\hspace{0pt}{\includegraphics[width=120pt]{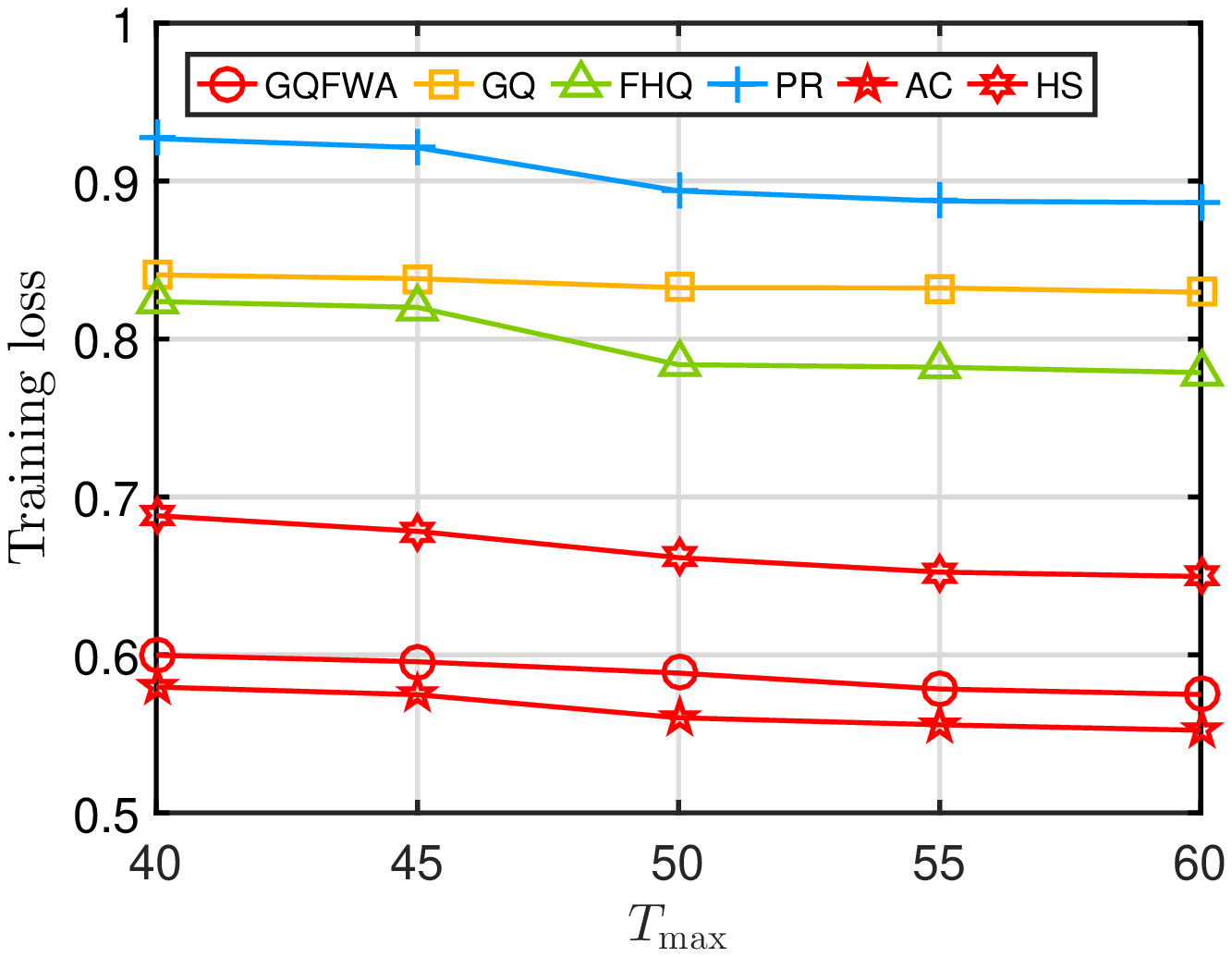}}\hspace{0pt}\label{Fig:L_Tmax_CommHetero_cost}}
\subfigure[\small{Training loss versus $E_{\max}$ at $T_{\max}=60$.}]
{\hspace{0pt}{\includegraphics[width=120pt]{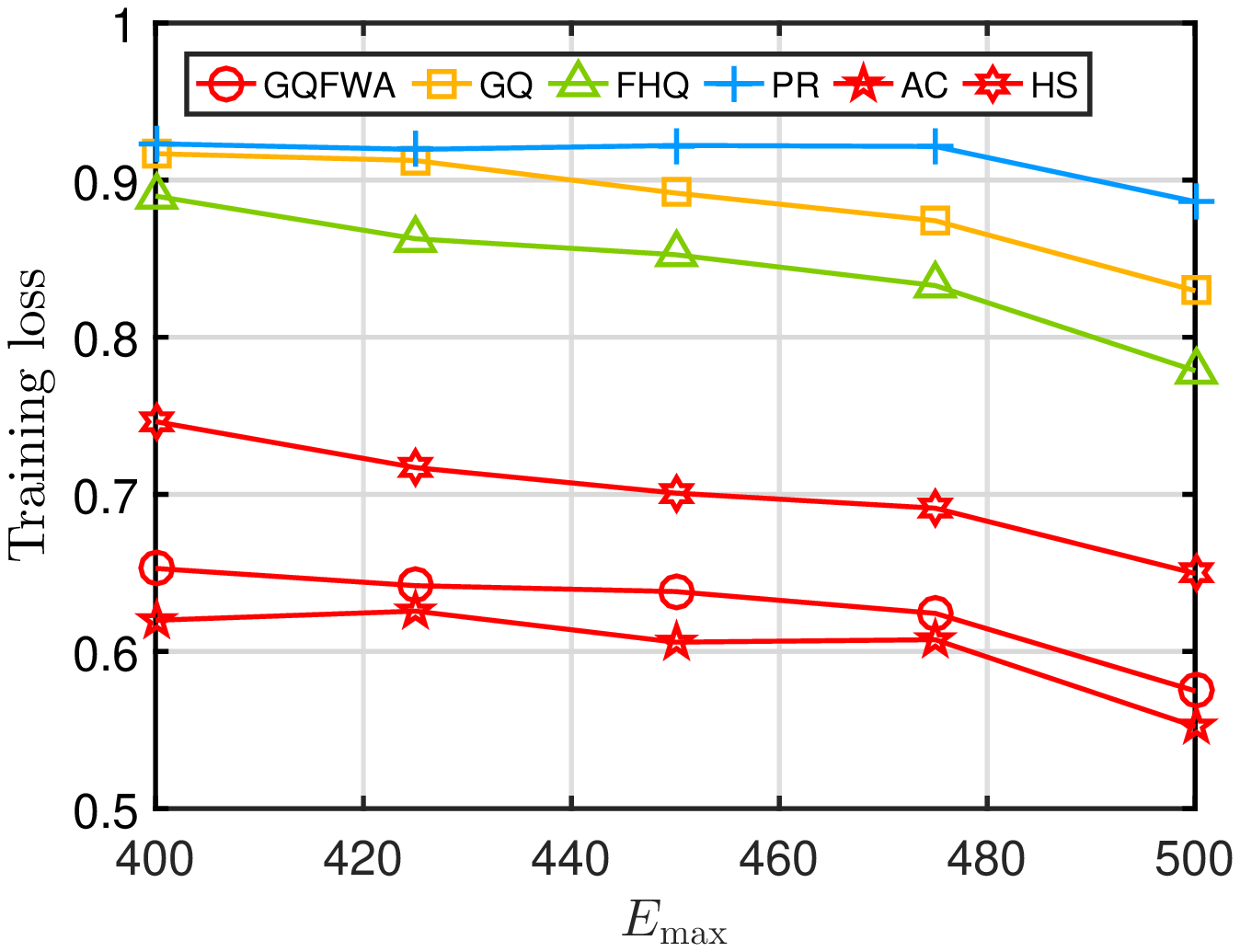}}\hspace{0pt}\label{Fig:L_Emax_CommHetero_cost}}
\subfigure[\small{Test accuracy versus $T_{\max}$ at $E_{\max}=500$.}]
{\hspace{0pt}{\includegraphics[width=120pt]{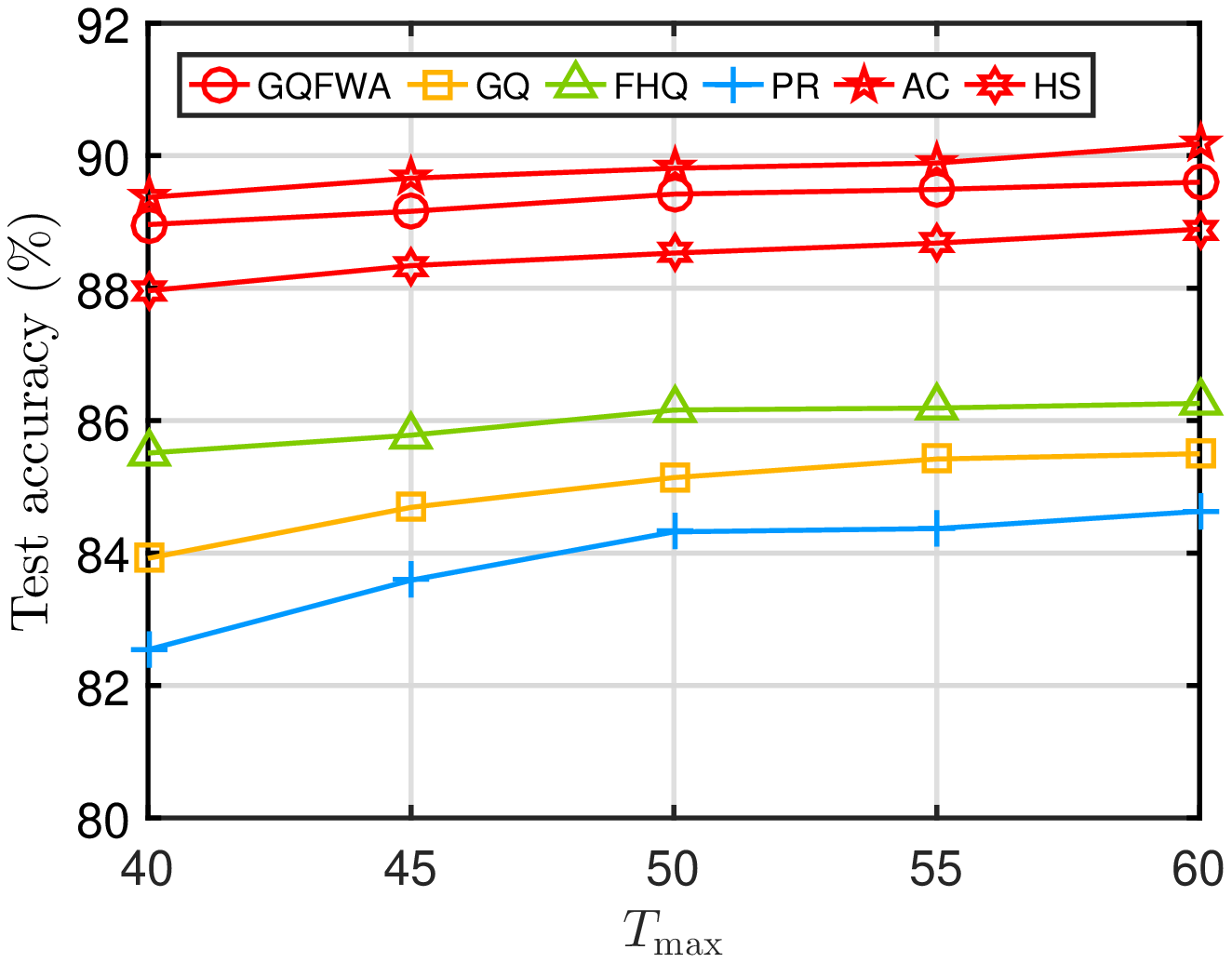}}\hspace{0pt}\label{Fig:L_Tmax_CommHetero_acc}}
\subfigure[\small{Test accuracy versus $E_{\max}$ at $T_{\max}=60$.}]
{\hspace{0pt}{\includegraphics[width=120pt]{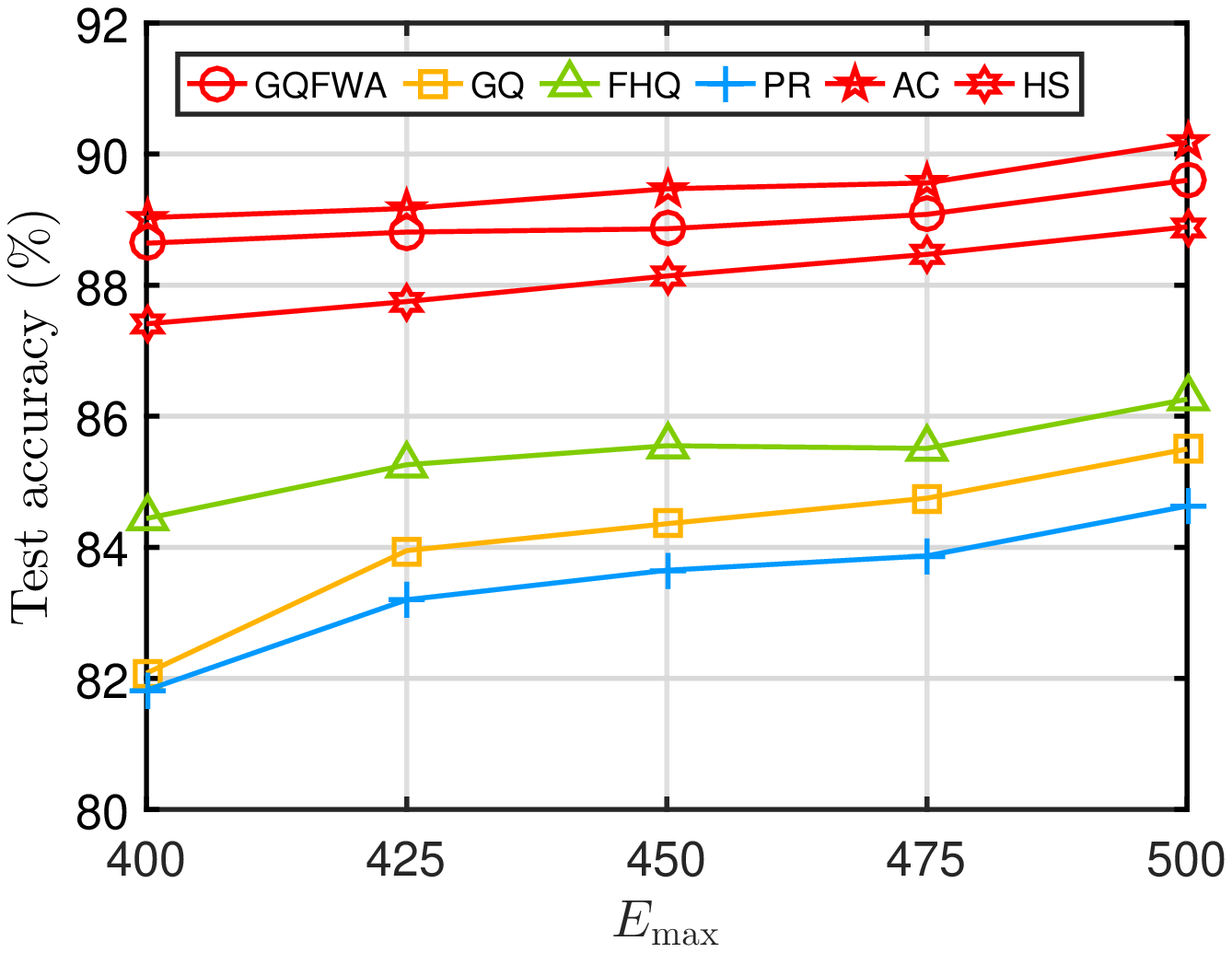}}\hspace{0pt}\label{Fig:L_Emax_CommHetero_acc}}
\end{center}
\caption{\small{Training loss and test accuracy for the communication-heterogeneous system.}}
\label{Fig:L_CommHetero}
\end{figure}
\begin{figure}[t]
\begin{center}
\subfigure[\small{Training loss versus $T_{\max}$ at $E_{\max}=500$.}]
{\hspace{0pt}{\includegraphics[width=120pt]{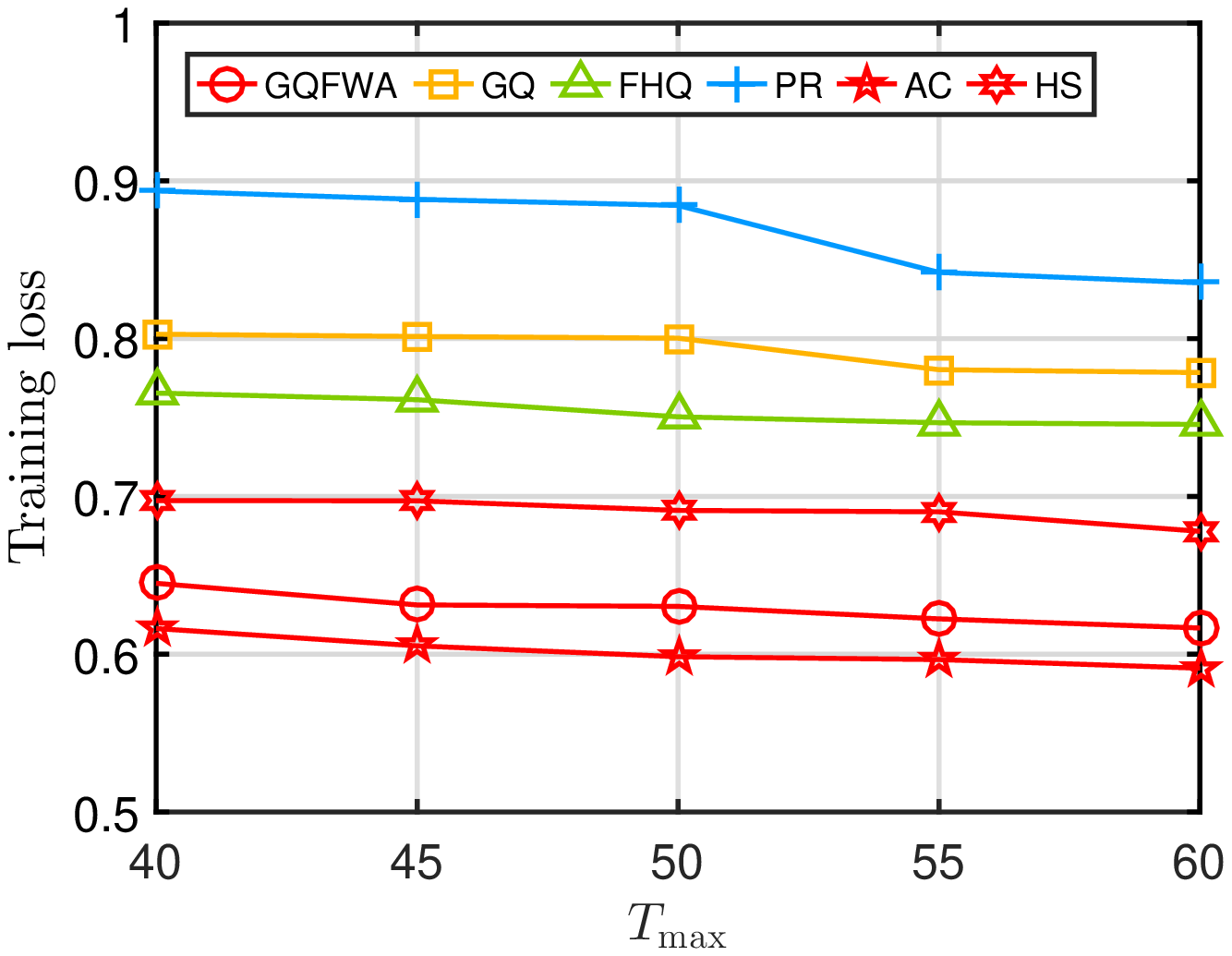}}\hspace{0pt}\label{Fig:L_Tmax_CompHetero_cost}}
\subfigure[\small{Training loss versus $E_{\max}$ at $T_{\max}=60$.}]
{\hspace{0pt}{\includegraphics[width=120pt]{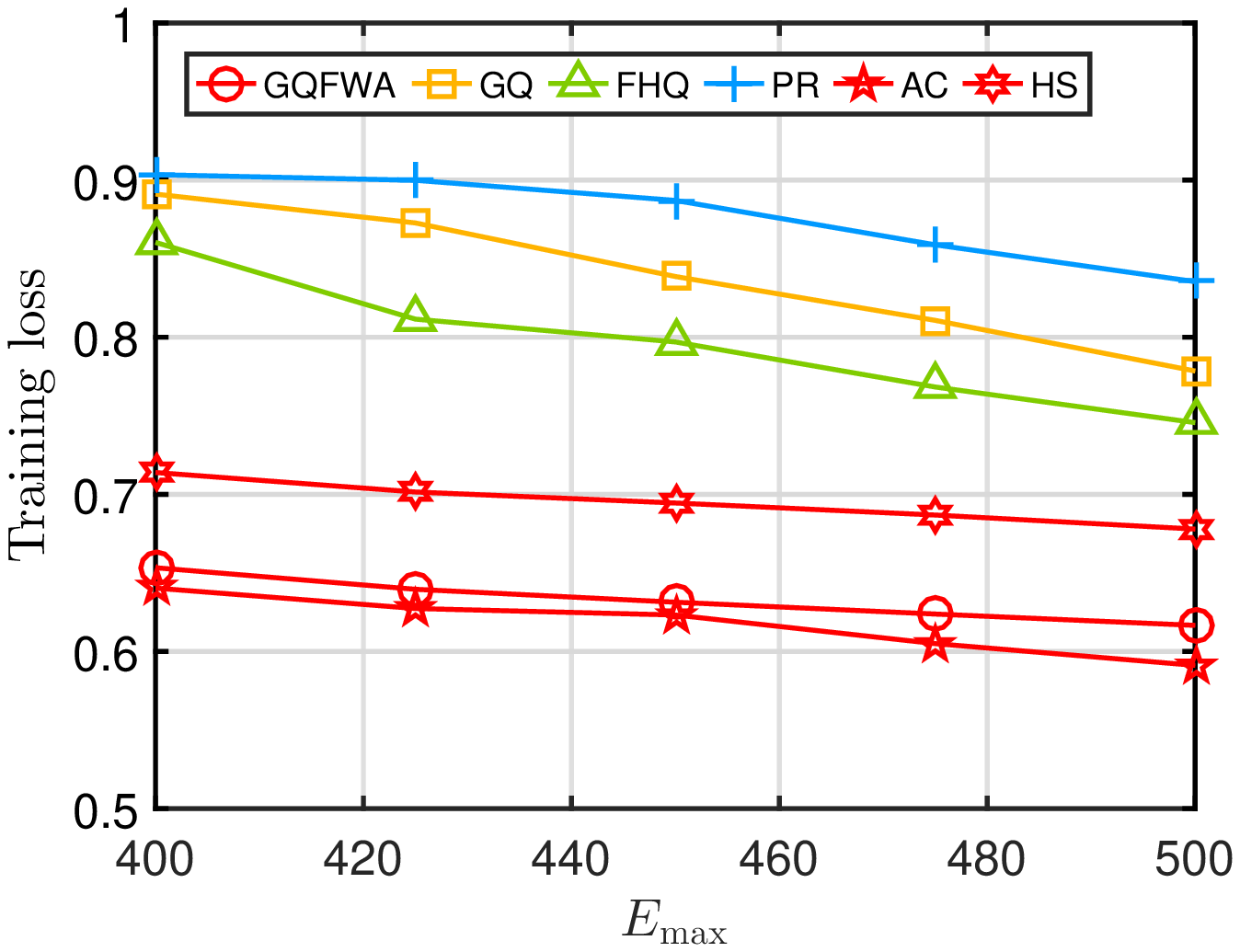}}\hspace{0pt}\label{Fig:L_Emax_CompHetero_cost}}
\subfigure[\small{Test accuracy versus $T_{\max}$ at $E_{\max}=500$.}]
{\hspace{0pt}{\includegraphics[width=120pt]{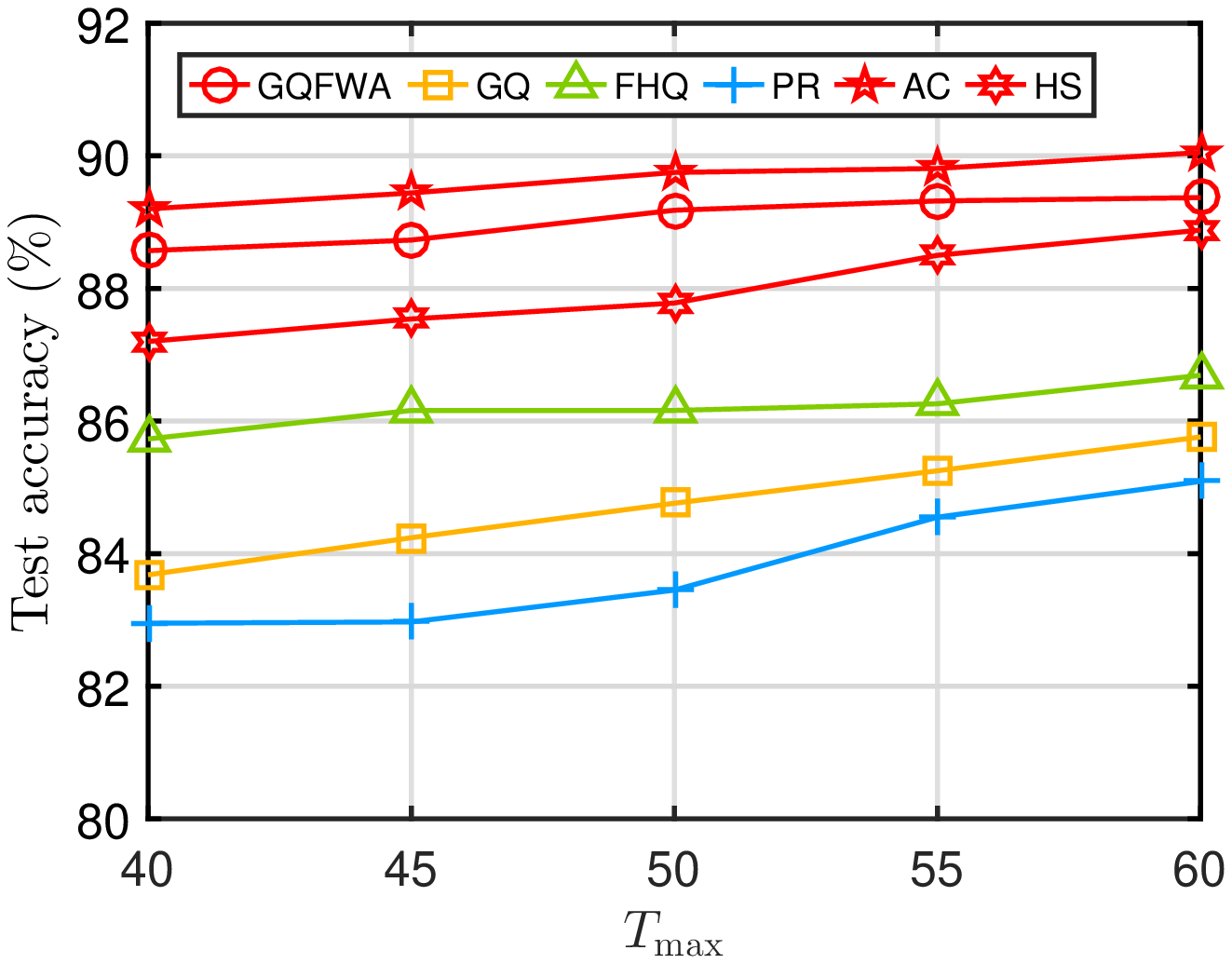}}\hspace{0pt}\label{Fig:L_Tmax_CompHetero_acc}}
\subfigure[\small{Test accuracy versus $E_{\max}$ at $T_{\max}=60$.}]
{\hspace{0pt}{\includegraphics[width=120pt]{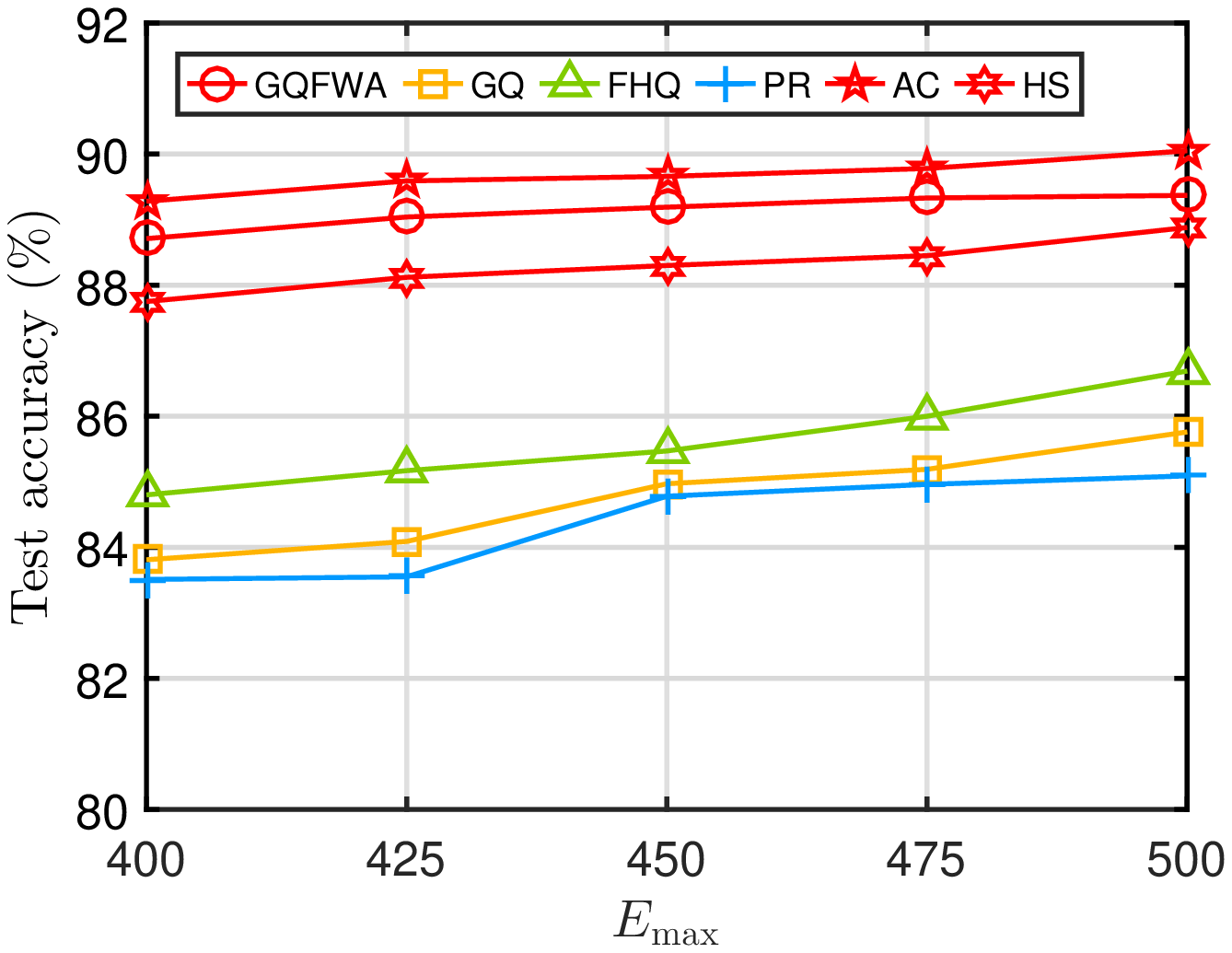}}\hspace{0pt}\label{Fig:L_Emax_CompHetero_acc}}
\end{center}
\caption{\small{Training loss and test accuracy for the computing-heterogeneous system.}}
\label{Fig:L_CompHetero}
\end{figure}

For the real-data experiments, we consider a ten-class classification problem with the MNIST dataset ($I=6\times10^4$) and partition it into $N=10$ subsets, each of which is stored on one worker.
We consider a neural network with three layers (i.e., two-layer neural network), including an input layer composed of $784$ cells, a hidden layer composed of $128$ cells, and an output layer composed of $10$ cells.
We use the sigmoid activation function for the hidden layer and the softmax activation function for the output layer.
We also use the cross entropy loss function to measure the classification performance of the considered neural network.
Consequently, we can specify the objective function of the ML problem, i.e., $f(\mathbf{x})$ in \eqref{eq:expected_loss}, and obtain the ML problem parameters $L$ and $\sigma$.
In each real-data experiment, we set system parameters (i.e., $C_n,F_n,p_n,r_n,n\in\setNbar$) and specify the optimization of algorithm parameters in Problem~\ref{Prob:Optimization}.
Then, by solving Problem~\ref{Prob:Optimization} using Algorithm~\ref{Alg:SolveEqProb}, we obtain the optimized FL algorithm parameters, (i.e., $\mathbf{K}$, $B$, $\mathbf\Gamma$, $\mathbf{W}$, $\tilde{\mathbf{s}}$, and $\mathbf{s}$).
Next, we deploy the optimized FL algorithm to solve the ML problem in the edge computing system and evaluate the training loss.

Fig.~\ref{Fig:Scale} plots the training losses of GQFedWAvg (with scaled model update-related vectors ${\mathbf x}_n^{(k_0,K_n)}-\hat{\mathbf x}^{(k_0)}$, $n\in\mathcal{N}$ and $\Delta\hat{\mathbf x}^{(k_0)}$) and a modified version of GQFedWAvg (with the original unscaled model update-related vectors $\frac{{\mathbf x}_n^{(k_0,K_n)}-\hat{\mathbf x}^{(k_0)}}{\gamma^{(k_0)}K_n}$, $n\in\mathcal{N}$ and $\frac{\Delta\hat{\mathbf x}^{(k_0)}}{\sum_{n\in\mathcal N}W_nK_n}$) versus $K_n,n\in\mathcal{N}$.
From Fig.~\ref{Fig:Scale}, we can see that the training loss of GQFedWAvg is lower than that of the unscaled version of it, which indicates the benefit of effectively utilizing the communication resources.
Fig.~\ref{Fig:L_Homo}-Fig.~\ref{Fig:L_CompHetero} show the training losses and test accuracies of existing FL algorithms and the proposed GQFedWAvg versus $T_{\max}$ and $E_{\max}$ for the homogeneous, communication-heterogeneous, and computing-heterogeneous systems, respectively.
From Fig.~\ref{Fig:L_Homo}-Fig.~\ref{Fig:L_CompHetero}, we can draw the following conclusions.
First, the trends of the training losses and the convergence errors are similar, which indicates that the convergence error given by \eqref{eq:Convergence_RHS} can effectively characterize the training loss and in practice \cite{GenQSGD}.
Second, the performance of GQFedWAvg is worse than that of AC. Their gap represents the performance loss due to quantization.

\section{Conclusion}
This paper attempted to optimize FL \clrtmp{implementation in} general edge computing systems.
We first \clrtmp{presented} a new quantization scheme and analyzed its properties.
Then, we \clrtmp{proposed} a general quantized FL algorithm, GQFedWAvg, which \clrtmp{applies the proposed quantization scheme to quantize wisely chosen model update-related vectors and adopts a generalized mini-batch SGD method with the weighted average local model updates in global model aggregation}.
GQFedWAvg has multiple adjustable algorithm parameters to \clrtmp{flexibly} adapt to the uniform or nonuniform computing and communication resources at the workers.
Moreover, we proposed an optimization framework for optimizing the algorithm parameters of GQFedWAvg to minimize the convergence error under the constraints on the time cost and energy cost.
Finally, numerical results reveal several important insights for optimally implementing FL.
Firstly, the wisely chosen scaled model update-related vectors in the proposed quantization scheme can effectively utilize the communication resources.
Secondly, the optimized numbers of local iterations and quantization parameters adapt to the computing and communication resources at workers, respectively.
Thirdly, the optimized weights in the weighted average of local model updates simultaneously fit the computing and communication resources at the workers.
Last but not least, the computing and communication resources in a homogeneous system can be more effectively utilized than those in a heterogeneous system.

\section*{Appendix A: Proof of Lemma~\ref{Lem:Quantization}}\label{Prf:Quantization}
\clrtmp{First, we show Lemma~\ref{Lem:Quantization}~(i).
We have:
\begin{small}\begin{align}
&\mathbb E\left[Q_d(\mathbf y;\tilde s,s,\Delta)\right]
{\overset{(a)}{=}}\mathbb E\left[\mathfrak{Q}\left(\|\mathbf y\|_2;\tilde s,\Delta\right)\text{sgn}(y_d)\mathfrak{Q}\left(\frac{|y_d|}{\|\mathbf y\|_2};s,1\right)\right]\nonumber\\ 
=&\text{sgn}(y_d)\mathbb E\left[\mathfrak{Q}\left(\|\mathbf y\|_2;\tilde s,\Delta\right)\right]\mathbb E\left[\mathfrak{Q}\left(\frac{|y_d|}{\|\mathbf y\|_2};s,1\right)\right]\nonumber\\
{\overset{(b)}{=}}&\text{sgn}(y_d)\|\mathbf y\|_2\frac{|y_d|}{\|\mathbf y\|_2}
=y_d,\ d\in\mathcal D,
\end{align}\end{small}
where (a) follows from \eqref{eq:Q_d}, and (b) follows because $\mathbb E\left[\mathfrak{Q}\left(\|\mathbf y\|_2;\tilde s,\Delta\right)\right]=\|\mathbf y\|_2$ and $\mathbb E[\mathfrak{Q}(\frac{|y_d|}{\|\mathbf y\|_2};s,1)]
\\=\frac{|y_d|}{\|\mathbf y\|_2}$, $d\in\mathcal{D}$ \cite[Lemma~3.1 (i)]{QSGD}.
Thus, we readily show Lemma~\ref{Lem:Quantization}~(i).}

Next, we show Lemma~\ref{Lem:Quantization}~(ii).
We have:
\begin{small}\begin{align}\label{eq:SquareStoQ}
&\mathbb E\!\left[\mathfrak{Q}\!\left(u;\mathfrak{s},\Delta\right)^2\right]
\!\!=\!\mathbb E\!\left[\mathfrak{Q}\!\left(u;\mathfrak{s},\Delta\right)\right]^2
\!\!+\!\mathbb E\!\!\left[\left(\mathfrak{Q}\!\left(u;\mathfrak{s},\Delta\right)\!-\!\mathbb E\!\left[\mathfrak{Q}\!\left(u;\mathfrak{s},\Delta\right)\right]\right)^2\right]\nonumber\\
&{\overset{(c)}{=}}u^2\!+\!\frac{\Delta^2}{\mathfrak{s}^2}\!\left(j-\frac{\mathfrak{s}u}{\Delta}\right)\!\!\left(1-j+\frac{\mathfrak{s}u}{\Delta}\right)
{\overset{(d)}{\leq}}u^2+\frac{\Delta^2}{4\mathfrak{s}^2},\ u\in[0,\Delta],
\end{align}\end{small}where (c) follows from \eqref{eq:RandVar} and Lemma~\ref{Lem:Quantization}~(i), and (d) is due to $\left(j-\frac{\mathfrak{s}u}{\Delta}\right)\in[0,1]$ and $\left(1-j+\frac{\mathfrak{s}u}{\Delta}\right)\in[0,1]$.
In addition, we have:
\begin{small}\begin{align}\label{eq:Lem1_tmp3}
&\mathbb E\left[\left\|{\mathbf Q}(\mathbf y;\tilde s,s,\Delta)\right\|_2^2\right]
{\overset{(e)}{=}}\!\sum_{d\in\mathcal D}\!\mathbb E\!\left[\!\left(\!\mathfrak{Q}\!\left(\|\mathbf y\|_2;\tilde s,\Delta\right)\mathfrak{Q}\!\left(\!\frac{|y_d|}{\|\mathbf y\|_2};s,1\right)\right)^2\right]\nonumber\\
=&\!\mathbb E\left[\mathfrak{Q}\left(\|\mathbf y\|_2;\tilde s,\Delta\right)^2\right]\!\sum_{d\in\mathcal D}\!\mathbb E\left[\mathfrak{Q}\left(\frac{|y_d|}{\|\mathbf y\|_2};s,1\right)^2\right]\nonumber\\    
{\overset{(f)}{\leq}}&\left(\|\mathbf y\|_2^2+\frac{\Delta^2}{4\tilde s^2}\right)\sum_{d\in\mathcal D}\mathbb E\left[\mathfrak{Q}\left(\frac{|y_d|}{\|\mathbf y\|_2};s,1\right)^2\right]\nonumber\\
{\overset{(g)}{\leq}}&\left(\|\mathbf y\|_2^2+\frac{\Delta^2}{4\tilde s^2}\right)\left(1+\min\left(\frac{D}{s^2}, \frac{\sqrt{D}}{s}\right)\right),
\end{align}\end{small}where (e) follows from \eqref{eq:Q_d}, (f) follows from \eqref{eq:SquareStoQ} and (g) can be shown following the proof of \cite[Lemma~3.1]{QSGD}.
Thus, by \eqref{eq:Lem1_tmp3} and $\mathbb E\left[\left\|{\mathbf Q}(\mathbf y;\tilde s,s,\Delta)-\mathbf y\right\|_2^2\right]
=\mathbb E\left[\left\|{\mathbf Q}(\mathbf y;\tilde s,s,\Delta)\right\|_2^2\right]-\|\mathbf y\|_2^2$, we readily show Lemma~\ref{Lem:Quantization}~(ii).

\clrtmp{
Finally, we show Lemma~\ref{Lem:Quantization}~(iii).
We have:
\begin{small}\begin{align}
&\left\|{\mathbf Q}(\mathbf y;\tilde s,s,\Delta)\right\|_2^2
{\overset{(h)}{=}}\sum_{d\in\mathcal D}\left(\mathfrak{Q}\left(\|\mathbf y\|_2;\tilde s,\Delta\right)\mathfrak{Q}\left(\frac{|y_d|}{\|\mathbf y\|_2};s,1\right)\right)^2\nonumber\\
=&\mathfrak{Q}\left(\|\mathbf y\|_2;\tilde s,\Delta\right)^2\sum_{d\in\mathcal D}\mathfrak{Q}\left(\frac{|y_d|}{\|\mathbf y\|_2};s,1\right)^2\nonumber\\
{\overset{(i)}{\leq}}&\left(\left\|\mathbf{y}\right\|_2+\frac{\Delta}{\tilde{s}}\right)^2\sum_{d\in\mathcal D}\left(\frac{\vert y_d\vert}{\left\|\mathbf{y}\right\|_2}+\frac{1}{s}\right)^2\nonumber\\
=&\left(\left\|\mathbf{y}\right\|_2+\frac{\Delta}{\tilde{s}}\right)^2\sum_{d\in\mathcal D}\left(\frac{\vert y_d\vert^2}{\left\|\mathbf{y}\right\|_2^2}+\frac{1}{s^2}+\frac{2\vert y_d\vert}{s\left\|\mathbf{y}\right\|_2}\right)\nonumber\\
{\overset{(j)}{=}}&\left(\left\|\mathbf{y}\right\|_2+\frac{\Delta}{\tilde{s}}\right)^2\left(1+\frac{D}{s^2}+\frac{2\sqrt{D}}{s}\right)\nonumber\\
=&\left(\left\|\mathbf{y}\right\|_2+\frac{\Delta}{\tilde{s}}\right)^2\left(1+\frac{\sqrt{D}}{s}\right)^2,\\
&\left\|{\mathbf Q}(\mathbf y;\tilde s,s,\Delta)\right\|_2^2
{\overset{(k)}{=}}\sum_{d\in\mathcal D}\left(\mathfrak{Q}\left(\|\mathbf y\|_2;\tilde s,\Delta\right)\mathfrak{Q}\left(\frac{|y_d|}{\|\mathbf y\|_2};s,1\right)\right)^2\nonumber\\
=&\mathfrak{Q}\left(\|\mathbf y\|_2;\tilde s,\Delta\right)^2\sum_{d\in\mathcal D}\mathfrak{Q}\left(\frac{|y_d|}{\|\mathbf y\|_2};s,1\right)^2\nonumber\\
{\overset{(l)}{\geq}}&\left(\left\|\mathbf{y}\right\|_2-\frac{\Delta}{\tilde{s}}\right)^2\sum_{d\in\mathcal D}\left(\frac{\vert y_d\vert}{\left\|\mathbf{y}\right\|_2}-\frac{1}{s}\right)^2\nonumber\\
=&\left(\left\|\mathbf{y}\right\|_2-\frac{\Delta}{\tilde{s}}\right)^2\sum_{d\in\mathcal D}\left(\frac{\vert y_d\vert^2}{\left\|\mathbf{y}\right\|_2^2}+\frac{1}{s^2}-\frac{2\vert y_d\vert}{s\left\|\mathbf{y}\right\|_2}\right)\nonumber\\
{\overset{(m)}{=}}&\left(\left\|\mathbf{y}\right\|_2-\frac{\Delta}{\tilde{s}}\right)^2\left(1+\frac{D}{s^2}-\frac{2\sqrt{D}}{s}\right)\nonumber\\
=&\left(\left\|\mathbf{y}\right\|_2-\frac{\Delta}{\tilde{s}}\right)^2\left(1-\frac{\sqrt{D}}{s}\right)^2,
\end{align}\end{small}where (h) and (k) follow from \eqref{eq:Q_d}, (i) and (l) follow from \eqref{eq:RandVar}, and (j) and (m) follow because $\sum_{d\in\mathcal{D}}\vert y_d\vert=\sqrt{\left(\sum_{d\in\mathcal{D}}\vert y_d\vert\right)^2}\leq\sqrt{D\sum_{d\in\mathcal{D}}y_d^2}=\sqrt{D}\left\|\mathbf{y}\right\|_2$.
Therefore, we complete the proof of Lemma~\ref{Lem:Quantization}.}

\section*{Appendix B: Proof of Theorem~\ref{Thm:Convergence}}\label{Prf:Convergence}
To prove Theorem~\ref{Thm:Convergence}, we first establish the following lemmas.
\begin{Lem}\label{Lem:xn-hatx}
Suppose that Assumptions~\ref{Asump:IID} and \ref{Asump:BoundedVariances} are satisfied.
Then, for all $\mathbf K\in\mathbb Z_{++}^{N+1}$, $B\in\mathbb Z_{++}$, and $\mathbf W\in(0,1)^N$ with $\mathbf 1^T\mathbf W=1$,
$\left\{\hat{\mathbf x}^{(k_0)}:k_0\in\mathcal K_0\right\}$ and
$\left\{\mathbf x_n^{(k_0,k_n)}:n\in\mathcal N,k_0\in\mathcal K_0,k_n\in\mathcal K_n\right\}$ satisfy:
\begin{small}\begin{align}\label{eq:xn-hatx}
&\mathbb E\left[\left\|\mathbf x_n^{(k_0,k_n)}-\hat{\mathbf x}^{(k_0)}\right\|_2^2\right]
\leq\left(\gamma^{(k_0)}\right)^2\frac{K_n\sigma^2}{B}\nonumber\\
&+\left(\gamma^{(k_0)}\right)^2K_n\sum_{k\in\mathcal K_n}\mathbb E\left[\left\|\nabla f\left(\mathbf x_n^{(k_0,k)}\right)\right\|_2^2\right],\ k_0\in\mathcal K_0.
\end{align}\end{small}
\end{Lem}
\begin{IEEEproof}
By ${\mathbf x}_n^{(k_0,0)}=\hat{\mathbf x}^{(k_0)}$ and \eqref{eq:local_update}, we have:
\begin{scriptsize}\begin{align}
&\mathbb E\left[\left\|\mathbf x_n^{(k_0,k_n)}-\hat{\mathbf x}^{(k_0)}\right\|_2^2\right]\nonumber\\
=&\mathbb E\left[\left\|\gamma^{(k_0)}\sum_k^{k_n}\frac{1}{B}\sum_{\xi\in\mathcal B_n^{(k_0,k)}}\nabla{F\left({\mathbf x}_n^{(k_0,k-1)};\xi\right)}\right\|_2^2\right]\nonumber\\
{\overset{(a)}{=}}&\!\left(\!\gamma^{(k_0)}\!\right)^2\!\mathbb E\left[\left\|\sum_k^{k_n}\!\left(\!\frac{1}{B}\!\!\!\!\sum_{\xi\in\mathcal B_n^{(k_0,k)}}\!\!\!\!\nabla{F\left({\mathbf x}_n^{(k_0,k-1)};\xi\right)}\!-\!\nabla f\left(\mathbf x_n^{(k_0,k-1)}\right)\right)\right\|_2^2\right]\nonumber\\
&\!+\!\left(\gamma^{(k_0)}\!\right)^2\!\mathbb E\left[\left\|\sum_k^{k_n}\!\nabla f\left(\mathbf x_n^{(k_0,k-1)}\right)\right\|_2^2\right]\nonumber\\
{\overset{(b)}{\leq}}&\left(\gamma^{(k_0)}\right)^2\sum_k^{k_n}\mathbb E\left[\left\|\frac{1}{B}\sum_{\xi\in\mathcal B_n^{(k_0,k)}}\nabla{F\left({\mathbf x}_n^{(k_0,k-1)};\xi\right)}-\nabla f\left(\mathbf x_n^{(k_0,k)}\right)\right\|_2^2\right]\nonumber\\
&\!+\!\left(\gamma^{(k_0)}\right)^2k_n\sum_k^{k_n}\mathbb E\left[\left\|\nabla f\left(\mathbf x_n^{(k_0,k-1)}\right)\right\|_2^2\right]\nonumber\\
{\overset{(c)}{\leq}}&\left(\gamma^{(k_0)}\right)^2\frac{K_n\sigma^2}{B}
+\left(\gamma^{(k_0)}\right)^2K_n\sum_k^{k_n}\mathbb E\left[\left\|\nabla f\left(\mathbf x_n^{(k_0,k-1)}\right)\right\|_2^2\right],\nonumber\\
&\ n\in\mathcal N,\ k_0\in\mathcal K_0,\ k_n\in\mathcal K_n,\nonumber
\end{align}\end{scriptsize}where (a) is due to $\mathbb E\left[\frac{1}{B}\sum_{\xi\in\mathcal B_n^{(k_0,k)}}\nabla{F\left({\mathbf x}_n^{(k_0,k-1)};\xi\right)}\right]=\nabla f\left(\mathbf x_n^{(k_0,k-1)}\right)$ and $\mathbb E\left[\left\|\mathbf z-\mathbb E\left[\mathbf z\right]\right\|_2^2\right]=\mathbb E\left[\left\|\mathbf z\right\|_2^2\right]-\left\|\mathbb E\left[\mathbf z\right]\right\|_2^2$ for any random vector $\mathbf z$,
(b) is due to $\frac{1}{B}\sum_{\xi\in\mathcal B_n^{(k_0,k)}}\nabla{F\left({\mathbf x}_n^{(k_0,k-1)};\xi\right)}-\nabla f\left(\mathbf x_n^{(k_0,k-1)}\right)$, $n\in\mathcal N$ are independent and have zero mean,
and (c) follows from Assumption~\ref{Asump:BoundedVariances} and inequalities $k_n\leq K_n$, $k_n\in\mathcal{K}_n$, $n\in\mathcal{N}$ and $\|\sum_{i=1}^n{\mathbf z}_i\|^2\leq n\sum_{i=1}^n{\|{\mathbf z}_i\|^2}$.
Thus, we can show Lemma~\ref{Lem:xn-hatx}.
\end{IEEEproof}
\begin{Lem}\label{Lem:Term1Bound}
Suppose that Assumptions~\ref{Asump:IID}, \ref{Asump:Smoothness}, and \ref{Asump:BoundedVariances} are satisfied.
Then, for all $\mathbf K\in\mathbb Z_{++}^{N+1}$, $B\in\mathbb Z_{++}$, and $\mathbf W\in(0,1)^N$ with $\mathbf 1^T\mathbf W=1$,
$\left\{\hat{\mathbf x}^{(k_0)}:k_0\in\mathcal K_0\right\}$ satisfies:
\begin{small}\begin{align}\label{eq:Term1Bound}
&\mathbb E\left[\left\langle\nabla f\left(\hat{\mathbf x}^{(k_0)}\right),\hat{\mathbf x}^{(k_0+1)}\!-\!\hat{\mathbf x}^{(k_0)}\right\rangle\right]\nonumber\\
\leq&\!-\frac{\gamma^{(k_0)}}{2}\!\!\sum_{n\in\mathcal N}\!\!W_nK_n\mathbb E\left[\left\|\nabla f\left(\hat{\mathbf x}^{(k_0)}\right)\right\|_2^2\right]\nonumber\\
&\!-\!\frac{\gamma^{(k_0)}}{2}\!\!\sum_{n\in\mathcal N}\!\!W_n\!\!\left(\!1\!-\!L^2\!\!\left(\!\gamma^{(k_0)}\!\right)^2\!K_n\!\!\right)\!\!\sum_{k_n\in\mathcal K_n}\!\!\!\mathbb E\!\left[\left\|\nabla f\left(\mathbf x_n^{(k_0,k_n-1)}\right)\right\|_2^2\right]\nonumber\\
&+\frac{L^2\sigma^2\left(\gamma^{(k_0)}\right)^3}{4B}\sum_{n\in\mathcal N}W_nK_n\left(K_n+1\right),\ k_0\in\mathcal K_0.
\end{align}\end{small}
\end{Lem}
\begin{IEEEproof}
We have:
\begin{small}\begin{align}
&\mathbb E\left[\left\langle\nabla f\left(\hat{\mathbf x}^{(k_0)}\right),\hat{\mathbf x}^{(k_0+1)}-\hat{\mathbf x}^{(k_0)}\right\rangle\right]\nonumber\\
{\overset{(a)}{=}}&\mathbb E\left[\left\langle\nabla f\left(\hat{\mathbf x}^{(k_0)}\right),\sum_{n\in\mathcal N}W_n\left(\mathbf x_n^{(k_0,k_n)}-\hat{\mathbf x}^{(k_0)}\right)\right\rangle\right]\nonumber\\
{\overset{(b)}{=}}&-\mathbb E\left[\left\langle\nabla f\left(\hat{\mathbf x}^{(k_0)}\right),\gamma^{(k_0)}\sum_{n\in\mathcal N}W_n\sum_{k_n\in\mathcal K_n}\nabla f\left(\mathbf x_n^{(k_0,k_n-1)}\right)\right\rangle\right]\nonumber\\
{\overset{(c)}{=}}&-\frac{\gamma^{(k_0)}}{2}\sum_{n\in\mathcal N}W_n\sum_{k_n\in\mathcal K_n}\mathbb E\left[\left\|\nabla f\left(\hat{\mathbf x}^{(k_0)}\right)\right\|_2^2\right.\nonumber\\
&\left.+\left\|\nabla f\left(\mathbf x_n^{(k_0,k_n-1)}\right)\right\|_2^2\!-\!\left\|\nabla f\left(\hat{\mathbf x}^{(k_0)}\right)\!-\!\nabla f\left(\mathbf x_n^{(k_0,k_n-1)}\right)\right\|_2^2\right]\nonumber\\
{\overset{(d)}{\leq}}&-\frac{\gamma^{(k_0)}}{2}\sum_{n\in\mathcal N}W_n\sum_{k_n\in\mathcal K_n}\mathbb E\left[\left\|\nabla f\left(\hat{\mathbf x}^{(k_0)}\right)\right\|_2^2\right]\nonumber\\
&-\frac{\gamma^{(k_0)}}{2}\sum_{n\in\mathcal N}W_n\sum_{k_n\in\mathcal K_n}\mathbb E\left[\left\|\nabla f\left(\mathbf x_n^{(k_0,k_n-1)}\right)\right\|_2^2\right]\nonumber\\
&+\frac{L^2\gamma^{(k_0)}}{2}\sum_{n\in\mathcal N}W_n\sum_{k_n\in\mathcal K_n}\mathbb E\left[\left\|\mathbf x_n^{(k_0,k_n-1)}-\hat{\mathbf x}^{(k_0)}\right\|_2^2\right]\nonumber\\
{\overset{(e)}{\leq}}&-\frac{\gamma^{(k_0)}}{2}\sum_{n\in\mathcal N}W_nK_n\mathbb E\left[\left\|\nabla f\left(\hat{\mathbf x}^{(k_0)}\right)\right\|_2^2\right]
-\frac{\gamma^{(k_0)}}{2}\nonumber\\
&\times\!\!\sum_{n\in\mathcal N}\!W_n\!\!\left(\!1\!-\!L^2\!\left(\!\gamma^{(k_0)}\!\right)^2\!K_n\!\right)\!\!\sum_{k_n\in\mathcal K_n}\mathbb E\left[\left\|\nabla f\left(\mathbf x_n^{(k_0,k_n-1)}\right)\right\|_2^2\right]\nonumber\\
&+\frac{L^2\sigma^2\left(\gamma^{(k_0)}\right)^3}{4B}\sum_{n\in\mathcal N}W_nK_n\left(K_n+1\right),\ k_0\in\mathcal K_0,\nonumber
\end{align}\end{small}where (a) follows from \eqref{eq:RecoveredLocalModel}, \eqref{eq:GlobalUpdate}, and Lemma~\ref{Lem:Quantization} (i), (b) is due to \eqref{eq:local_update} and $\mathbb E\left[\frac{1}{B}\sum_{\xi\in\mathcal B_n^{(k_0,k_n)}}\right.\\\left.\nabla{F\left({\mathbf x}_n^{(k_0,k_n-1)};\xi\right)}\right]=\nabla f\left(\mathbf x_n^{(k_0,k_n-1)}\right)$,
(c) follows from the equality $\langle\mathbf z_1,\mathbf z_2\rangle=\frac{1}{2}(\|\mathbf z_1\|^2+\|\mathbf z_2\|^2-\|\mathbf z_1-\mathbf z_2\|^2)$ for any $\mathbf z_1$ and $\mathbf z_2$ with the same length,
(d) follows from ${\mathbf x}_n^{(k_0,0)}=\hat{\mathbf x}^{(k_0)}$ and Assumption~\ref{Asump:Smoothness},
and (e) follows from \eqref{eq:xn-hatx} and the equality $\sum_{i=1}^n i=\frac{n(n+1)}{2}$.
Thus, we can show Lemma~\ref{Lem:Term1Bound}.
\end{IEEEproof}
\begin{Lem}\label{Lem:Term2Bound}
Suppose that Assumptions~\ref{Asump:BoundedNorm}, \ref{Asump:IID}, \ref{Asump:Smoothness}, and \ref{Asump:BoundedVariances} are satisfied.
Then, for all $\mathbf K\in\mathbb Z_{++}^{N+1}$, $B\in\mathbb Z_{++}$, and $\mathbf W\in(0,1)^N$ with $\mathbf 1^T\mathbf W=1$,
$\left\{\hat{\mathbf x}^{(k_0)}:k_0\in\mathcal K_0\right\}$ satisfies:
\begin{small}\begin{align}\label{eq:Term2Bound}
&\mathbb E\left[\left\|\hat{\mathbf x}^{(k_0+1)}-\hat{\mathbf x}^{(k_0)}\right\|_2^2\right]\nonumber\\
\leq&\left(\gamma^{(k_0)}\right)^2\left(q_{\tilde s_0,s_0}\Delta_0^2+\left(q_{s_0}+1\right)\sum_{n\in\mathcal N}q_{\tilde s_n,s_n}W_n^2K_n^2\Delta_n^2\right)\nonumber\\
&+\frac{\sigma^2\left(\gamma^{(k_0)}\right)^2\left(1+q_{s_0}\right)}{B}\sum_{n\in\mathcal N}\!\left(N+q_{s_n}\right)\!W_n^2K_n\nonumber\\
&+\left(\gamma^{(k_0)}\right)^2\left(1+q_{s_0}\right)\sum_{n\in\mathcal N}\left(N+q_{s_n}\right)W_n^2K_n\nonumber\\
&\times\sum_{k_n\in\mathcal K_n}\mathbb E\!\left[\left\|\nabla f\left(\mathbf x_n^{(k_0,k_n-1)}\right)\right\|_2^2\right],k_0\in\mathcal K_0.
\end{align}\end{small}
\end{Lem}
\begin{IEEEproof}
We have:
\begin{scriptsize}\begin{align}
&\mathbb E\left[\left\|\hat{\mathbf x}^{(k_0+1)}\!-\!\hat{\mathbf x}^{(k_0)}\right\|_2^2\right]\nonumber\\
&\!{\overset{(a)}{=}}\mathbb E\!\left[\left\|\gamma^{(k_0)}\!\mathbf Q\!\left(\!\frac{\!\sum\limits_{n\in\mathcal N}\!W_n\!K_n\!\mathbf Q\!\left(\!\frac{{\mathbf x}_n^{(k_0,K_n)}-\hat{\mathbf x}^{(k_0)}}{\gamma^{(k_0)}K_n};\tilde s_n,s_n,\!\Delta_n\!\right)}{\sum_{n\in\mathcal N}W_nK_n};\tilde s_0,s_0,\Delta_0\!\!\right)\right.\right.\nonumber\\
&\left.\left.\times\sum_{n\in\mathcal N}\!\!W_nK_n\right\|_2^2\right]\nonumber\\
&{\overset{(b)}{=}}\left(\gamma^{(k_0)}\right)^2\left(\sum_{n\in\mathcal N}W_nK_n\right)^2\nonumber\\
&\times\mathbb E\left[\left\|\mathbf Q\left(\frac{\sum_{n\in\mathcal N}W_nK_n\mathbf Q\left(\frac{{\mathbf x}_n^{(k_0,K_n)}-\hat{\mathbf x}^{(k_0)}}{\gamma^{(k_0)}K_n};\tilde s_n,s_n,\Delta_n\right)}{\sum_{n\in\mathcal N}W_nK_n};\tilde s_0,s_0,\Delta_0\right)\right.\right.\nonumber\\
&-\left.\left.\frac{\sum_{n\in\mathcal N}W_nK_n\mathbf Q\left(\frac{{\mathbf x}_n^{(k_0,K_n)}-\hat{\mathbf x}^{(k_0)}}{\gamma^{(k_0)}K_n};\tilde s_n,s_n,\Delta_n\right)}{\sum_{n\in\mathcal N}W_nK_n}\right\|_2^2\right]\nonumber\\
&+\left(\gamma^{(k_0)}\right)^2\left(\sum_{n\in\mathcal N}W_nK_n\right)^2\nonumber\\
&\times\mathbb E\left[\left\|\frac{\sum_{n\in\mathcal N}W_nK_n\mathbf Q\left(\frac{{\mathbf x}_n^{(k_0,K_n)}-\hat{\mathbf x}^{(k_0)}}{\gamma^{(k_0)}K_n};\tilde s_n,s_n,\Delta_n\right)}{\sum_{n\in\mathcal N}W_nK_n}\right\|_2^2\right]\nonumber\\
&{\overset{(c)}{\leq}}\left(\gamma^{(k_0)}\right)^2\left(\sum_{n\in\mathcal N}W_nK_n\right)^2\left(q_{\tilde s_0,s_0}\Delta_0^2\right.\nonumber\\
&\left.+q_{s_0}\mathbb E\left[\left\|\frac{\sum_{n\in\mathcal N}W_nK_n\mathbf Q\left(\frac{{\mathbf x}_n^{(k_0,K_n)}-\hat{\mathbf x}^{(k_0)}}{\gamma^{(k_0)}K_n};\tilde s_n,s_n,\Delta_n\right)}{\sum_{n\in\mathcal N}W_nK_n}\right\|_2^2\right]\right)\nonumber\\
&+\left(\gamma^{(k_0)}\right)^2\left(\sum_{n\in\mathcal N}W_nK_n\right)^2\nonumber\\
&\times\mathbb E\left[\left\|\frac{\sum_{n\in\mathcal N}W_nK_n\mathbf Q\left(\frac{{\mathbf x}_n^{(k_0,K_n)}-\hat{\mathbf x}^{(k_0)}}{\gamma^{(k_0)}K_n};\tilde s_n,s_n,\Delta_n\right)}{\sum_{n\in\mathcal N}W_nK_n}\right\|_2^2\right]\nonumber\\
&=\left(\gamma^{(k_0)}\right)^2\!q_{\tilde s_0,s_0}\Delta_0^2\left(\sum_{n\in\mathcal N}W_nK_n\right)^2
+\!\left(\gamma^{(k_0)}\right)^2\!\left(1+q_{s_0}\right)\nonumber\\
&\times\mathbb E\left[\left\|\sum_{n\in\mathcal N}W_nK_n\mathbf Q\left(\frac{{\mathbf x}_n^{(k_0,K_n)}-\hat{\mathbf x}^{(k_0)}}{\gamma^{(k_0)}K_n};\tilde s_n,s_n,\Delta_n\right)\right\|_2^2\right]\nonumber\\
&{\overset{(d)}{=}}\left(\gamma^{(k_0)}\right)^2q_{\tilde s_0,s_0}\Delta_0^2\left(\sum_{n\in\mathcal N}W_nK_n\right)^2
+\left(\gamma^{(k_0)}\right)^2\left(1+q_{s_0}\right)\nonumber\\
&\times\left(\mathbb E\left[\left\|\sum_{n\in\mathcal N}W_nK_n\mathbf Q\left(\frac{{\mathbf x}_n^{(k_0,K_n)}-\hat{\mathbf x}^{(k_0)}}{\gamma^{(k_0)}K_n};\tilde s_n,s_n,\Delta_n\right)\right.\right.\right.\nonumber\\
&\left.\left.\left.-\sum_{n\in\mathcal N}W_nK_n\frac{{\mathbf x}_n^{(k_0,K_n)}-\hat{\mathbf x}^{(k_0)}}{\gamma^{(k_0)}K_n}\right\|_2^2\right]\right.\nonumber\\
&\left.+\mathbb E\left[\left\|\sum_{n\in\mathcal N}W_nK_n\frac{{\mathbf x}_n^{(k_0,K_n)}-\hat{\mathbf x}^{(k_0)}}{\gamma^{(k_0)}K_n}\right\|_2^2\right]\right)\nonumber\\
&{\overset{(e)}{\leq}}\left(\gamma^{(k_0)}\right)^2q_{\tilde s_0,s_0}\Delta_0^2\left(\sum_{n\in\mathcal N}W_nK_n\right)^2
+\left(\gamma^{(k_0)}\right)^2\left(1+q_{s_0}\right)\nonumber\\
&\times\sum_{n\in\mathcal N}W_n^2K_n^2\mathbb E\left[\left\|\mathbf Q\left(\frac{{\mathbf x}_n^{(k_0,K_n)}-\hat{\mathbf x}^{(k_0)}}{\gamma^{(k_0)}K_n};\tilde s_n,s_n,\Delta_n\right)\right.\right.\nonumber\\
&-\left.\left.\frac{{\mathbf x}_n^{(k_0,K_n)}-\hat{\mathbf x}^{(k_0)}}{\gamma^{(k_0)}K_n}\right\|_2^2\right]\nonumber\\
&+\left(\gamma^{(k_0)}\right)^2\left(1+q_{s_0}\right)N\sum_{n\in\mathcal N}W_n^2K_n^2\mathbb E\left[\left\|\frac{{\mathbf x}_n^{(k_0,K_n)}-\hat{\mathbf x}^{(k_0)}}{\gamma^{(k_0)}K_n}\right\|_2^2\right]\nonumber\\
&{\overset{(f)}{\leq}}\left(\gamma^{(k_0)}\right)^2\!q_{\tilde s_0,s_0}\Delta_0^2\!\left(\sum_{n\in\mathcal N}\!W_nK_n\!\right)^2
\!\!+\!\left(\gamma^{(k_0)}\right)^2\!\left(1+q_{s_0}\right)\nonumber\\
&\times\sum_{n\in\mathcal N}\!W_n^2K_n^2\!\left(\!q_{\tilde s_n,s_n}\Delta_n^2+q_{s_n}\mathbb E\left[\left\|\frac{{\mathbf x}_n^{(k_0,K_n)}-\hat{\mathbf x}^{(k_0)}}{\gamma^{(k_0)}K_n}\right\|_2^2\right]\right)\nonumber\\
&+\left(\gamma^{(k_0)}\right)^2\left(1+q_{s_0}\right)N\sum_{n\in\mathcal N}W_n^2K_n^2\mathbb E\left[\left\|\frac{{\mathbf x}_n^{(k_0,K_n)}-\hat{\mathbf x}^{(k_0)}}{\gamma^{(k_0)}K_n}\right\|_2^2\right]\nonumber\\
&=\!\left(\!\gamma^{(k_0)}\!\right)^2\!\!\left(\!q_{\tilde s_0,s_0}\Delta_0^2\!\left(\!\sum_{n\in\mathcal N}\!W_nK_n\!\right)^2\!+\!\left(1+q_{s_0}\right)\!\sum_{n\in\mathcal N}\!W_n^2K_n^2q_{\tilde s_n,s_n}\Delta_n^2\right)\nonumber\\
&+\left(1+q_{s_0}\right)\sum_{n\in\mathcal N}\left(N+q_{s_n}\right)W_n^2\mathbb E\left[\left\|{\mathbf x}_n^{(k_0,K_n)}-\hat{\mathbf x}^{(k_0)}\right\|_2^2\right]\nonumber\\
&{\overset{(g)}{\leq}}\!\left(\!\gamma^{(k_0)}\!\right)^2\!\!\left(\!q_{\tilde s_0,s_0}\Delta_0^2\!\left(\!\sum_{n\in\mathcal N}\!W_nK_n\!\right)^2\!+\!\left(1+q_{s_0}\right)\!\sum_{n\in\mathcal N}\!W_n^2K_n^2q_{\tilde s_n,s_n}\Delta_n^2\right)\nonumber\\
&+\left(\gamma^{(k_0)}\right)^2\left(1+q_{s_0}\right)\sum_{n\in\mathcal N}\left(N+q_{s_n}\right)W_n^2K_n\nonumber\\
&\times\left(\frac{\sigma^2}{B}+\sum_{k_n\in\mathcal K_n}\mathbb E\left[\left\|\nabla f\left(\mathbf x_n^{(k_0,k_n-1)}\right)\right\|_2^2\right]\right),\ k_0\in\mathcal K_0,\nonumber
\end{align}\end{scriptsize}where (a) follows from \eqref{eq:RecoveredLocalModel}, (b) and (d) follow from the equality $\mathbb E\left[\left\|\mathbf z-\mathbb E\left[\mathbf z\right]\right\|_2^2\right]=\mathbb E\left[\left\|\mathbf z\right\|_2^2\right]-\left\|\mathbb E\left[\mathbf z\right]\right\|_2^2$ for any random vector $\mathbf{z}$,
(c) and (f) follow from Lemma~\ref{Lem:Quantization} (ii),
(e) follows from the inequality $\|\sum_{i=1}^n{\bf z}_i\|^2\leq n\sum_{i=1}^n{\|{\bf z}_i\|^2}$,
and (g) follows from Lemma~\ref{Lem:xn-hatx}.
Thus, we can show Lemma~\ref{Lem:Term2Bound}.
\end{IEEEproof}
Then, by \eqref{eq:Cons_gamma}, \eqref{eq:Term1Bound}, \eqref{eq:Term2Bound}, and Assumption~\ref{Asump:Smoothness}, we have:
\begin{scriptsize}\begin{align}\label{eq:Final1}
&\mathbb E\left[f\left(\hat{\mathbf x}^{(k_0+1)}\right)-f\left(\hat{\mathbf x}^{(k_0)}\right)\right]\nonumber\\
&\leq\mathbb E\left[\!\left\langle\nabla f\!\left(\hat{\mathbf x}^{(k_0)}\right),\hat{\mathbf x}^{(k_0+1)}-\hat{\mathbf x}^{(k_0)}\right\rangle\right]
+\frac{L}{2}\mathbb E\left[\left\|\hat{\mathbf x}^{(k_0+1)}-\hat{\mathbf x}^{(k_0)}\right\|_2^2\right]\nonumber\\
&\leq-\frac{\gamma^{(k_0)}}{2}\!\!\sum_{n\in\mathcal N}\!W_nK_n\mathbb E\left[\left\|\nabla f\left(\hat{\mathbf x}^{(k_0)}\right)\right\|_2^2\right]\nonumber\\
&-\frac{\gamma^{(k_0)}}{2}\!\!\sum_{n\in\mathcal N}W_n\left(1-L^2\left(\gamma^{(k_0)}\right)^2\!\!K_n\!\right)\!\!\sum_{k_n\in\mathcal K_n}\!\!\mathbb E\!\left[\left\|\nabla f\!\left(\mathbf x_n^{(k_0,k_n-1)}\right)\right\|_2^2\right]\nonumber\\
&+\frac{L^2\sigma^2\left(\gamma^{(k_0)}\right)^3}{4B}\sum_{n\in\mathcal N}W_nK_n\left(K_n+1\right)+\frac{L\left(\gamma^{(k_0)}\right)^2}{2}\nonumber\\
&\times\left(q_{\tilde s_0,s_0}\Delta_0^2\left(\sum_{n\in\mathcal N}W_nK_n\right)^2+\left(1+q_{s_0}\right)\sum_{n\in\mathcal N}W_n^2K_n^2q_{\tilde s_n,s_n}\Delta_n^2\right)\nonumber\\
&+\frac{L\left(\gamma^{(k_0)}\right)^2\left(1+q_{s_0}\right)}{2}\sum_{n\in\mathcal N}\left(N+q_{s_n}\right)W_n^2K_n\nonumber\\
&\times\left(\frac{\sigma^2}{B}+\sum_{k_n\in\mathcal K_n}\mathbb E\left[\left\|\nabla f\left(\mathbf x_n^{(k_0,k_n-1)}\right)\right\|_2^2\right]\right)\nonumber\\
&=-\frac{\gamma^{(k_0)}}{2}\sum_{n\in\mathcal N}W_nK_n\mathbb E\left[\left\|\nabla f\left(\hat{\mathbf x}^{(k_0)}\right)\right\|_2^2\right]\nonumber\\
&+\frac{L^2\sigma^2\left(\gamma^{(k_0)}\right)^3}{4B}\sum_{n\in\mathcal N}W_nK_n\left(K_n+1\right)
-\frac{\gamma^{(k_0)}}{2}\sum_{n\in\mathcal N}W_n\nonumber\\
&\times\left(1-L^2\left(\gamma^{(k_0)}\right)^2K_n-L\gamma^{(k_0)}\left(1+q_{s_0}\right)\left(N+q_{s_n}\right)W_nK_n\right)\nonumber\\
&\times\sum_{k_n\in\mathcal K_n}\mathbb E\left[\left\|\nabla f\left(\mathbf x_n^{(k_0,k_n-1)}\right)\right\|_2^2\right]\nonumber\\
&+\!\frac{L\left(\gamma^{(k_0)}\right)^2}{2}\!\!\left(\!q_{\tilde s_0,s_0}\Delta_0^2\!\left(\!\sum_{n\in\mathcal N}\!W_nK_n\!\right)^2\!\!+\!\left(1\!+\!q_{s_0}\right)\!\!\sum_{n\in\mathcal N}\!W_n^2K_n^2q_{\tilde s_n,s_n}\Delta_n^2\!\right)\nonumber\\
&+\!\frac{L\sigma^2\left(\gamma^{(k_0)}\!\right)^2\!\left(1+q_{s_0}\right)}{2B}\!\!\sum_{n\in\mathcal N}\!\left(N\!+\!q_{s_n}\right)\!W_n^2K_n\nonumber\\
&\leq\!-\!\frac{\gamma^{(k_0)}}{2}\!\!\sum_{n\in\mathcal N}\!\!W_nK_n\mathbb E\left[\left\|\nabla f\left(\hat{\mathbf x}^{(k_0)}\right)\right\|_2^2\right]\nonumber\\
&+\frac{L^2\sigma^2\left(\gamma^{(k_0)}\right)^3}{4B}\sum_{n\in\mathcal N}W_nK_n\left(K_n+1\right)
+\frac{L\left(\gamma^{(k_0)}\right)^2}{2}\nonumber\\
&\times\left(\!q_{\tilde s_0,s_0}\Delta_0^2\!\left(\sum_{n\in\mathcal N}\!\!W_nK_n\!\!\right)^2
+\left(1+q_{s_0}\right)\sum_{n\in\mathcal N}W_n^2K_n^2q_{\tilde s_n,s_n}\Delta_n^2\right)\nonumber\\
&+\frac{L\sigma^2\left(\gamma^{(k_0)}\right)^2\left(1+q_{s_0}\right)}{2B}\sum_{n\in\mathcal N}\left(N+q_{s_n}\right)W_n^2K_n,\ k_0\in\mathcal K_0.
\end{align}\end{scriptsize}
Thus, we have:
\begin{scriptsize}\begin{align}\label{eq:Final2}
&\gamma^{(k_0)}\sum_{n\in\mathcal N}W_nK_n\mathbb E\left[\left\|\nabla f\left(\hat{\mathbf x}^{(k_0)}\right)\right\|_2^2\right]
{\overset{(a)}{\leq}}2\mathbb E\left[f\left(\hat{\mathbf x}^{(k_0+1)}\right)-f\left(\hat{\mathbf x}^{(k_0)}\right)\right]\nonumber\\
&+\frac{L^2\sigma^2\left(\gamma^{(k_0)}\right)^3}{2B}\sum_{n\in\mathcal N}W_nK_n\left(K_n+1\right)+L\left(\gamma^{(k_0)}\right)^2\nonumber\\
&\times\left(q_{\tilde s_0,s_0}\Delta_0^2\left(\sum_{n\in\mathcal N}W_nK_n\right)^2
+\left(1+q_{s_0}\right)\sum_{n\in\mathcal N}W_n^2K_n^2q_{\tilde s_n,s_n}\Delta_n^2\right)\nonumber\\
&+\frac{L\sigma^2\left(\gamma^{(k_0)}\right)^2\left(1+q_{s_0}\right)}{B}\sum_{n\in\mathcal N}\left(N+q_{s_n}\right)W_n^2K_n,\ k_0\in\mathcal K_0,
\end{align}\end{scriptsize}where (a) follows from \eqref{eq:Final1}.
By summing both sides of \eqref{eq:Final2} over $k_0\in\mathcal K_0$, we readily show Theorem~\ref{Thm:Convergence}.

\section*{Appendix C: Proof of Theorem~\ref{Thm:Equivalence}}\label{Prf:Equivalence}
First, we construct an equivalent problem of Problem~\ref{Prob:Optimization} as follows.
\begin{Prob}[Equivalent Problem of Problem~\ref{Prob:Optimization}]\label{Prob:Eq1}
\begin{align}
\min_{\substack{\mathbf K,\mathbf W,\tilde{\mathbf s},\mathbf s\succ\mathbf 0,\\B,\gamma>0}}
&{\quad}C\left(\mathbf K,B,\gamma\mathbf1,\mathbf W,\tilde{\mathbf s},\mathbf s\right)\nonumber\\
\mathrm{s.t.}&{\quad}\eqref{eq:Cons_W},\ \eqref{eq:Cons_time},\ \eqref{eq:Cons_energy},\ \eqref{eq:Eq_gamma}.\nonumber
\end{align}
\end{Prob}
\begin{Lem}[Equivalence between Problem~\ref{Prob:Optimization} and Problem~\ref{Prob:Eq1}]\label{Lem:Eq1}
If $\left(\mathbf K^*,B^*,\gamma^*,\mathbf W^*,\tilde{\mathbf s}^*,\mathbf s^*\right)$ is an optimal point of Problem~\ref{Prob:Eq1}, then $\left(\mathbf K^*,B^*,\gamma^*\mathbf 1,\mathbf W^*,\tilde{\mathbf s}^*,\mathbf s^*\right)$ is an optimal point of Problem~\ref{Prob:Optimization}.
\end{Lem}
\begin{IEEEproof}
Following the proof of \cite[Theorem~8]{GenQSGD}, we can show Lemma~\ref{Lem:Eq1}.
\end{IEEEproof}
Then, we construct an equivalent problem of Problem~\ref{Prob:Eq1} as follows.
\begin{Prob}[Equivalent Problem of Problem~\ref{Prob:Eq1}]\label{Prob:Eq2}
\begin{align}
&\min_{\substack{\mathbf K,\mathbf W,\tilde{\mathbf s},\mathbf s,\mathbf S\succ\mathbf 0,\\B,\gamma,T_1,T_2>0}}
{\quad}C\left(\mathbf K,B,\gamma\mathbf1,\mathbf W,\tilde{\mathbf s},\mathbf s\right)\nonumber\\
&\mathrm{s.t.}{\quad}\eqref{eq:Cons_W},\ \eqref{eq:T1},\ \eqref{eq:Eq_Sn},\ \eqref{eq:T2},\ \eqref{eq:Eq_gamma},\ \eqref{eq:Eq_T},\ \eqref{eq:Eq_E}.\nonumber
\end{align}
\end{Prob}
\begin{Lem}[Equivalence between Problem~\ref{Prob:Eq1} and Problem~\ref{Prob:Eq2}]\label{Lem:Eq2}
If $(\mathbf K^*,B^*,\gamma^*,\mathbf W^*,\tilde{\mathbf s}^*,\mathbf s^*,\mathbf S^*,T_1^*,T_2^*)$ is an optimal point of Problem~\ref{Prob:Eq2}, then $\left(\mathbf K^*,B^*,\gamma^*,\mathbf W^*,\tilde{\mathbf s}^*,\mathbf s^*\right)$ is an optimal point of Problem~\ref{Prob:Eq1}.
\end{Lem}
\begin{IEEEproof}
Suppose that there exists $\left(\mathbf K^\dag,B^\dag,\gamma^\dag,\mathbf W^\dag,\tilde{\mathbf s}^\dag,\mathbf s^\dag\right)\neq\left(\mathbf K^*,B^*,\gamma^*,\mathbf W^*,\tilde{\mathbf s}^*,\mathbf s^*\right)$, where $\mathbf K^\dag\triangleq(K_n^\dag)_{n\in\setNbar}$,
$\tilde{\mathbf s}^\dag\triangleq(\tilde s_n^\dag)_{n\in\setNbar}$,
$\mathbf s^\dag\triangleq(s_n^\dag)_{n\in\setNbar}$,
$\mathbf W^\dag\triangleq(W_n^\dag)_{n\in\setNbar}$,
satisfying all constraints of Problem~\ref{Prob:Eq1} and $ C\left(\mathbf K^\dag,B^\dag,\gamma^\dag\mathbf1,\mathbf W^\dag,\tilde{\mathbf s}^\dag,\mathbf s^\dag\right)< C\left(\mathbf K^*,B^*,\gamma^*\mathbf1,\mathbf W^*,\tilde{\mathbf s}^*,\mathbf s^*\right)$.
Construct $\hat{T_1}=\max_{n\in\mathcal N}\frac{C_n}{F_n}K_n^\dag$,
$\hat T_2=\max_{n\in\mathcal N}\frac{S_n}{r_n}$,
and $\hat{\mathbf S}=\left(\hat S_n\right)_{n\in\setNbar}$ with $\hat{S_n}=\log_2\left(\tilde s_n^\dag+1\right)+D\log_2\left(s_n^\dag+1\right)+D,n\in\setNbar$.
Obviously, $(\mathbf K^\dag,B^\dag,\gamma^\dag,\mathbf W^\dag,\tilde{\mathbf s}^\dag,\mathbf s^\dag,\hat{\mathbf S},\hat{T_1},\hat{T_2})$ satisfies all constraints of Problem~\ref{Prob:Eq2} and $C(\mathbf K^\dag,B^\dag,\gamma^\dag\mathbf1,\mathbf W^\dag,\tilde{\mathbf s}^\dag,\mathbf s^\dag)<C(\mathbf K^*,B^*,\gamma^*\mathbf1,\mathbf W^*,\tilde{\mathbf s}^*,\mathbf s^*)$,
contradicting with the optimality of $\left(\mathbf K^*,B^*,\gamma^*,\mathbf W^*,\tilde{\mathbf s}^*,\mathbf s^*,\mathbf S^*,T_1^*,T_2^*\right)$ for Problem~\ref{Prob:Eq2}.
Thus, by contradiction, we can show that $\left(\mathbf K^*,B^*,\gamma^*,\mathbf W^*,\tilde{\mathbf s}^*,\mathbf s^*\right)$ is an optimal point of Problem~\ref{Prob:Eq1}.
\end{IEEEproof}
Next, we construct an equivalent problem of Problem~\ref{Prob:Eq2} as follows.
\begin{Prob}[Equivalent Problem of Problem~\ref{Prob:Eq2}]\label{Prob:Eq3}
\begin{align}
&\min_{\substack{\mathbf K,\mathbf W,\tilde{\mathbf s},\mathbf s,\mathbf S,\mathbf q_{\mathbf s}\succ\mathbf 0,\\B,\gamma,T_1,T_2>0}}
{\quad}C\left(\mathbf K,B,\gamma\mathbf1,\mathbf W,\tilde{\mathbf s},\mathbf s\right)\nonumber\\
&\mathrm{s.t.}{\quad}\eqref{eq:T1},\ \eqref{eq:Eq_W1},\ \eqref{eq:Eq_W2},\ \eqref{eq:Eq_Sn},\ \eqref{eq:T2},\ \eqref{eq:Eq_gamma},\ \eqref{eq:Eq_T},\ \eqref{eq:Eq_E},\nonumber\\
&\quad\quad q_{s_n}=\min\left(\frac{D}{s_n^2},\frac{\sqrt{D}}{s_n}\right),\ n\in\setNbar.\nonumber
\end{align}
\end{Prob}
\begin{Lem}[Equivalence between Problem~\ref{Prob:Eq2} and Problem~\ref{Prob:Eq3}]\label{Lem:Eq3}
If $(\mathbf K^*\!,B^*\!,\gamma^*\!,\mathbf W^*\!,\tilde{\mathbf s}^*\!,\mathbf s^*\!,\mathbf S^*\!,T_1^*\!,T_2^*\!,\mathbf q_{\mathbf s}^*)$ is an optimal point of Problem~\ref{Prob:Eq3}, then $\left(\mathbf K^*,B^*,\gamma^*,\mathbf W^*,\tilde{\mathbf s}^*,\mathbf s^*,\mathbf S^*,T_1^*,T_2^*\right)$ is an optimal point of Problem~\ref{Prob:Eq2}.
\end{Lem}
By Lemma~\ref{Lem:Eq1}, Lemma~\ref{Lem:Eq2}, and Lemma~\ref{Lem:Eq3}, we can show the equivalence between Problem~\ref{Prob:Eq3} and Problem~\ref{Prob:Optimization}.
It remains to show the equivalence between Problem~\ref{Prob:Eq3} and Problem~\ref{Prob:EqProb}.
Before that, we show the following lemma.
\begin{Lem}\label{Lem:min_eq}
The condition $A_{\min}=\min\left(A_1,A_2\right)$ holds if and only if the conditions
\begin{align}\label{eq:min_eq}
&A_{\min}\leq A_1,\
A_{\min}\leq A_2,\
A_{\min}\geq A_1-y,\
A_{\min}\geq A_2-y,\nonumber\\
&\left|A_1-A_2\right|=y,\
y\geq0
\end{align}
hold, where $A_1,A_2,A_{\min},y\in\mathbb R$.
\end{Lem}
\begin{IEEEproof}
It is obvious that if $A_{\min}=\min\left(A_1,A_2\right)$ holds, then \eqref{eq:min_eq} holds.
Next, we show that if the equations in \eqref{eq:min_eq} hold, then $A_{\min}=\min\left(A_1,A_2\right)$ holds by considering two cases.
Case (i) $A_1\leq A_2$, i.e., $y=A_2-A_1$: As $y=A_2-A_1$, we have $A_1-y=2A_1-A_2\leq A_1$ and $A_2-y=A_1$. Thus, $A_{\min}\geq A_1$. Combining $A_{\min}\geq A_1$ with $A_{\min}\leq A_1$, we have
$A_{\min}=A_1=\min\left(A_1,A_2\right)$.
Case (ii) $A_1\geq A_2$, i.e., $y=A_1-A_2$: By symmetry, we can show $A_{\min}=A_2=\min\left(A_1,A_2\right)$ following the proof for Case (i).
Thus, we complete the proof for Lemma~\ref{Lem:min_eq}.
\end{IEEEproof}
\clrtmp{Based on Lemma~\ref{Lem:min_eq}, Problem~\ref{Prob:Eq3} can be
equivalently transformed into the following problem:
\begin{Prob}[Equivalent Problem of Problem~\ref{Prob:Eq3}]\label{Prob:Eq4}
\begin{align}
&\min_{\substack{\mathbf K,\mathbf W,\tilde{\mathbf s},\mathbf s,\mathbf S,\mathbf q_{\mathbf s}\succ\mathbf 0,\mathbf{y}\succeq\mathbf0,\\B,\gamma,T_1,T_2>0}}
{\quad}C\left(\mathbf K,B,\gamma\mathbf1,\mathbf W,\tilde{\mathbf s},\mathbf s\right)\nonumber\\
\mathrm{s.t.}&{\quad}\eqref{eq:T1},\ \eqref{eq:Eq_W1},\ \eqref{eq:Eq_W2},\ \eqref{eq:Eq_Sn},\ \eqref{eq:T2},\ \eqref{eq:Eq_gamma},\ \eqref{eq:Eq_T},\ \eqref{eq:Eq_E},\nonumber\\
&q_{s_n}\!\leq\frac{D}{s_n^2},\
q_{s_n}\!\leq\frac{\sqrt{D}}{s_n},\
q_{s_n}\!\geq\frac{D}{s_n^2}-y_n,\
q_{s_n}\!\geq\frac{\sqrt{D}}{s_n}-y_n,\nonumber\\
&\left\vert\frac{D}{s_n^2}-\frac{\sqrt{D}}{s_n}\right\vert=y_n,\  n\in\setNbar.\nonumber
\end{align}
\end{Prob}
Furthermore, by relaxing $\left\vert\frac{D}{s_n^2}-\frac{\sqrt{D}}{s_n}\right\vert=y_n$ to $-y_n\leq\frac{D}{s_n^2}-\frac{\sqrt{D}}{s_n}\leq y_n$, $n\in\setNbar$  and adding the non-decreasing penalty term $\sum_{n\in\setNbar}y_n$ to the objective of Problem~\ref{Prob:Eq4}, we transform Problem~\ref{Prob:Eq4} into the following problem, which can be shown to be equivalent to Problem~\ref{Prob:Eq4}.
\begin{Prob}[Equivalent Problem of Problem~\ref{Prob:Eq4}]\label{Prob:Eq5}
\begin{align}
&\min_{\substack{\mathbf K,\mathbf W,\tilde{\mathbf s},\mathbf s,\mathbf S,\mathbf q_{\mathbf s}\succ\mathbf 0,\mathbf{y}\succeq\mathbf0,\\B,\gamma,T_1,T_2>0}}
{\quad}C\left(\mathbf K,B,\gamma\mathbf1,\mathbf W,\tilde{\mathbf s},\mathbf s\right)
+\sum_{n\in\setNbar}y_n\nonumber\\
\mathrm{s.t.}&{\quad}\eqref{eq:T1},\ \eqref{eq:Eq_W1},\ \eqref{eq:Eq_W2},\ \eqref{eq:Eq_Sn},\ \eqref{eq:T2},\ \eqref{eq:Eq_gamma},\ \eqref{eq:Eq_T},\ \eqref{eq:Eq_E},\nonumber\\
&q_{s_n}\!\leq\!\frac{D}{s_n^2},\
q_{s_n}\!\leq\!\frac{\sqrt{D}}{s_n},\
q_{s_n}\!\geq\!\frac{D}{s_n^2}-y_n,\
q_{s_n}\!\geq\!\frac{\sqrt{D}}{s_n}-y_n,\nonumber\\
&-y_n\!\leq\!\frac{D}{s_n^2}-\frac{\sqrt{D}}{s_n}\!\leq\! y_n,\  n\in\setNbar.\nonumber
\end{align}
\end{Prob}
By comparing Problem~\ref{Prob:Eq5} (where $\mathbf{y}\succeq\mathbf0$) and Problem~\ref{Prob:EqProb} (where $\mathbf{y}\succ\mathbf0$), which differ only in the range of $\mathbf{y}$, we can show Theorem~\ref{Thm:Equivalence}.}

\section*{Appendix D: Proof of Theorem~\ref{Thm:SolveEqProb_convergence}}\label{Prf:SolveEqProb_convergence}
\clrtmp{First, by the first-order Taylor series expansion \cite{MM},
we can show that the approximation of the function $\log_2\left(x+1\right)$, i.e., $\log_2\left(\hat{x}+1\right)+\frac{x-\hat{x}}{\left(\hat{x}+1\right)\ln2}$,
satisfies Properties (i), (ii), and (iii) in \cite{GIA}.
Letting $x=s_n,n\in\setNbar$ and $x=\tilde s_n,n\in\setNbar$, respectively, we can construct the constraint function in \eqref{eq:Approx_Sn} as the approximation of the constraint function in \eqref{eq:Eq_Sn}.
Furthermore, it is clear that the posynomials in the denominators of the objective and the constraint functions in \eqref{eq:Eq_W2}, \eqref{eq:q4}, and \eqref{eq:q6} have the same structural properties as those in \cite[(25)]{GenQSGD}.
Thus, by slightly extending the proof of \cite[Theorem~3]{GenQSGD}, we can show Theorem~\ref{Thm:SolveEqProb_convergence}.}

\bibliographystyle{IEEEtran}      
\bibliography{IEEEabrv,Journal2022}                        

\end{document}